\documentclass{CVM}

\usepackage{algorithmic}
\usepackage{multirow}
\usepackage{xspace}
\usepackage{stfloats}

\newcommand\self{LDTR\xspace}

\CVMsetup{
type      = {Research Article},
doi       = {s41095-0xx-xxxx-x},
title     = {\self: Transformer-based Lane Detection with Anchor-chain Representation},
author    = {Zhongyu Yang$^{1}$, Chen Shen$^{2}$, Wei Shao$^{2}$, Tengfei Xing$^{2}$, Runbo Hu$^{2}$, Pengfei Xu$^{2}$, Hua Chai$^{2}$, and {Ruini Xue}$^{1}$\cor\\
},
runauthor = {Zhongyu Yang et. al.},
abstract = {
   Despite recent advances in lane detection methods, scenarios with limited- or
   no-visual-clue of lanes due to factors such as lighting conditions and
   occlusion remain challenging and crucial for automated driving. 
   Moreover, current lane representations require complex post-processing and struggle with specific instances.
   Inspired by the DETR architecture, we propose \self, a transformer-based model
   to address these issues. Lanes are modeled with a novel anchor-chain, regarding a
   lane as a whole from the beginning, which enables \self to handle special
   lanes inherently. To enhance lane instance perception, \self incorporates a novel
   multi-referenced deformable attention module to distribute attention around
   the object. Additionally, \self incorporates two line IoU algorithms to improve
   convergence efficiency and employs a Gaussian heatmap auxiliary branch to
   enhance model representation capability during training. To evaluate lane 
   detection models, we rely on Fréchet distance, 
   parameterized F1-score, and additional synthetic metrics. Experimental
   results demonstrate that \self achieves state-of-the-art performance on well-known datasets.
},
keywords = {Transformer, Lane detection, Anchor-chain.},
copyright = {The Author(s)},
}

\begin{document}

\maketitle

    \begin{figure}[b] \vskip -4mm
    \small\renewcommand\arraystretch{1.3}
        \begin{tabular}{p{80.5mm}} \toprule\\ \end{tabular}
        \vskip -4.5mm \noindent \setlength{\tabcolsep}{1pt}
        \begin{tabular}{p{3.5mm}p{80mm}}
    $1\quad $ & University of Electronic Science and Technology of China, Chengdu, 611731, China. E-mail: Z. Yang, 202021080612@std.uestc.edu.cn; R. Xue, xueruini@uestc.edu.cn\cor.\\
    $2\quad $ & Didi Chuxing, Beijing, 100081, China. E-mail: C. Shen, jayshenchen@didiglobal.com; W. Shao, wayneshaowei@didiglobal.com; T. Xing, xingtengfei@didiglobal.com; R. Hu, hurunbo@didiglobal.com; P. Xu, pengfeixu@didiglobal.com; H. Chai, chaihua@didiglobal.com.\\
&\hspace{-5mm} Manuscript received: 2024-01-18; accepted: 2024-02-29\vspace{-2mm}
    \end{tabular} \vspace {-3mm}
    \end{figure}

\section{Introduction}
\label{sec:intro}
Autonomous driving is an important application of deep learning, in which the
ability to perceive road elements is particularly crucial, especially lane
markings, one of the most essential components of road traffic signs. However,
due to the complexity of road scenarios and lane deformation from varying
perspectives, accurate lane detection remains challenging, particularly
determining lanes with little- or no-visual-clue, and precise representations of special lanes.

Given good visibility and simple road conditions, traditional vision
research methods~\cite{kim2008robust, borkar2009robust} perform very
well. However, they lack robustness in complex real-world scenarios. Recently,
various DNN models~\cite{scnn, lanenet, laneaf, condlanenet} have been trained on
large-scale datasets, to infer lane positions via deep semantic features,
delivering significantly improved generalization and robustness compared to traditional
approaches.
\begin{figure*}[!t]
\centering
\begin{minipage}[b]{\linewidth}
  \rotatebox{90}{\parbox[c][0.02\linewidth][c]{0.08\linewidth}{\centering CLRNet}}
  \includegraphics[width=0.242\linewidth]{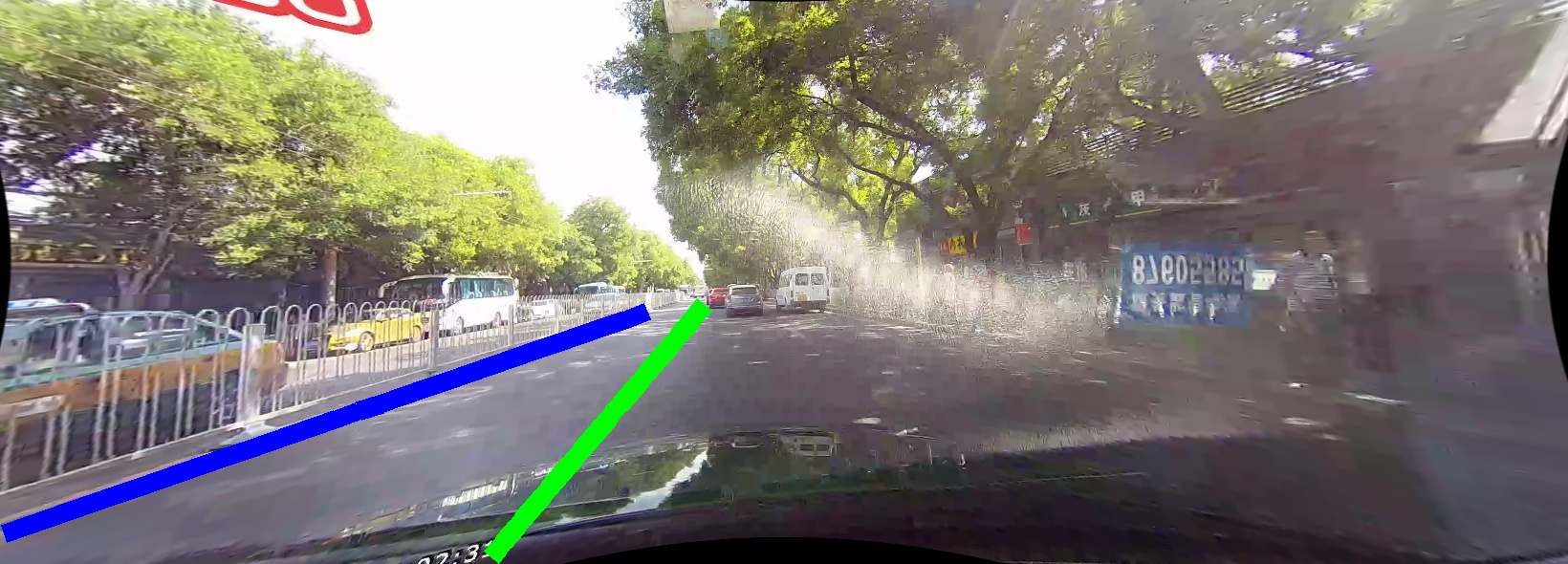}\hfill
  \includegraphics[width=0.242\linewidth]{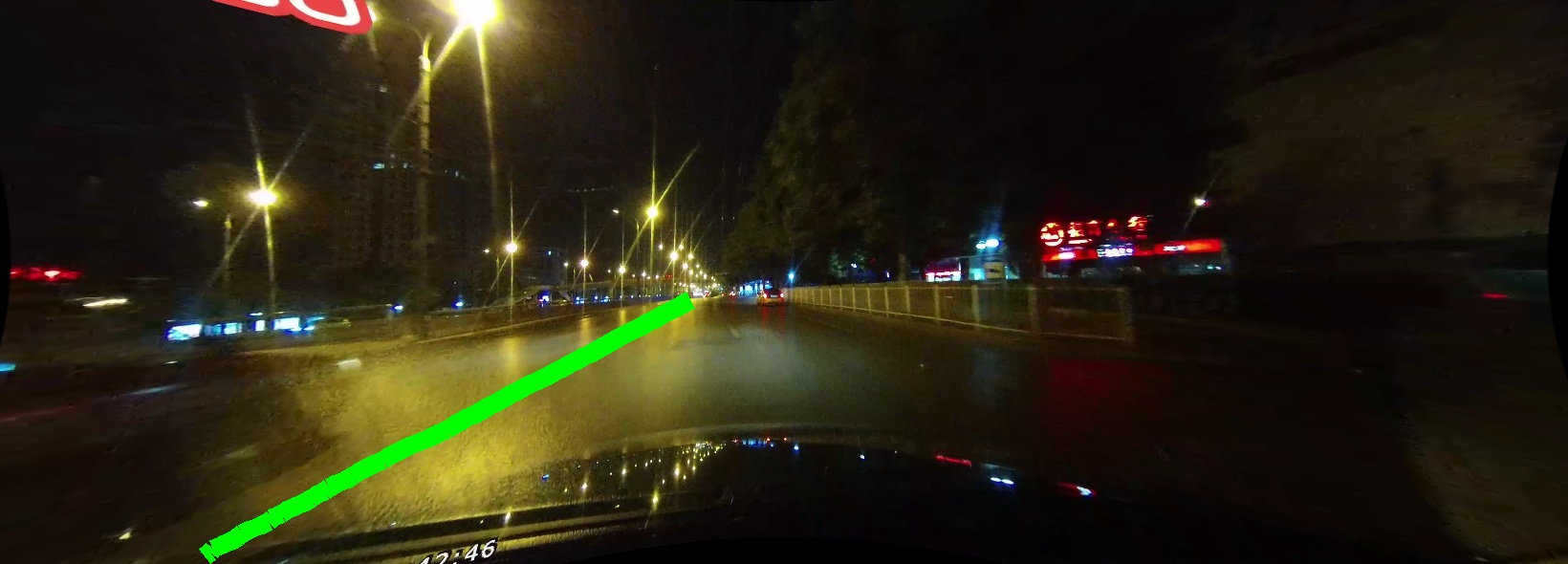}\hfill
  \includegraphics[width=0.242\linewidth]{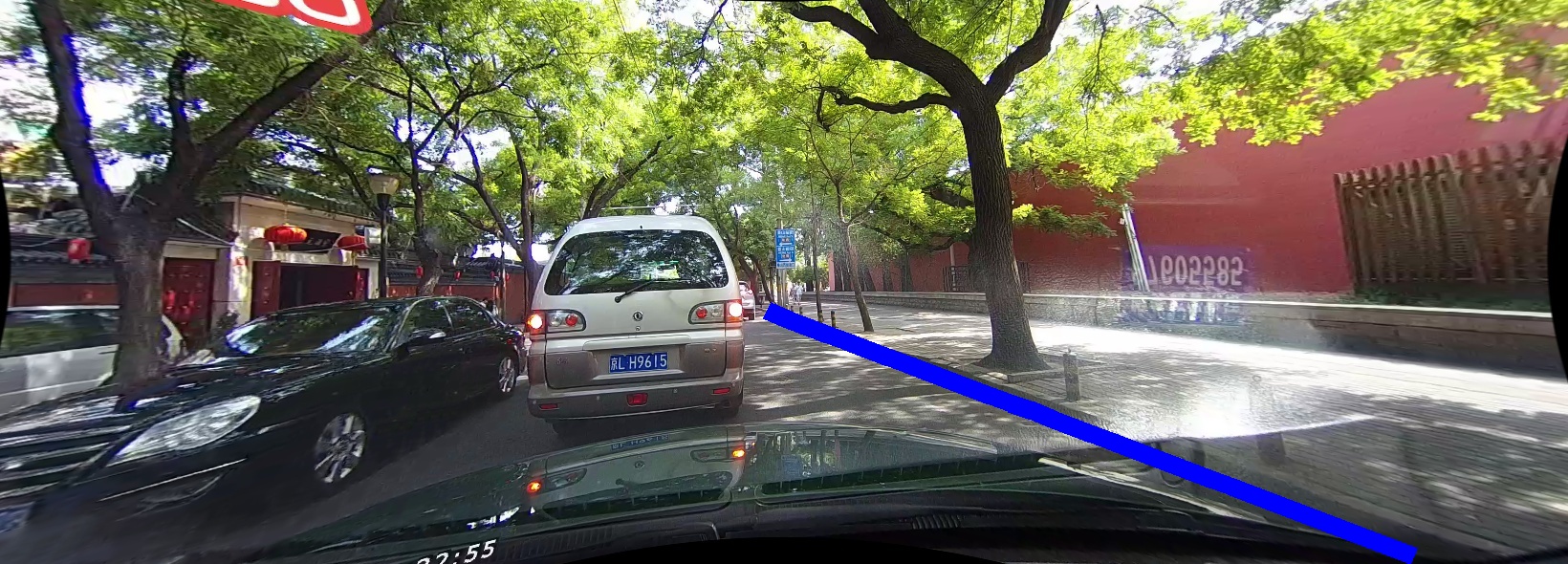}\hfill
  \includegraphics[width=0.242\linewidth]{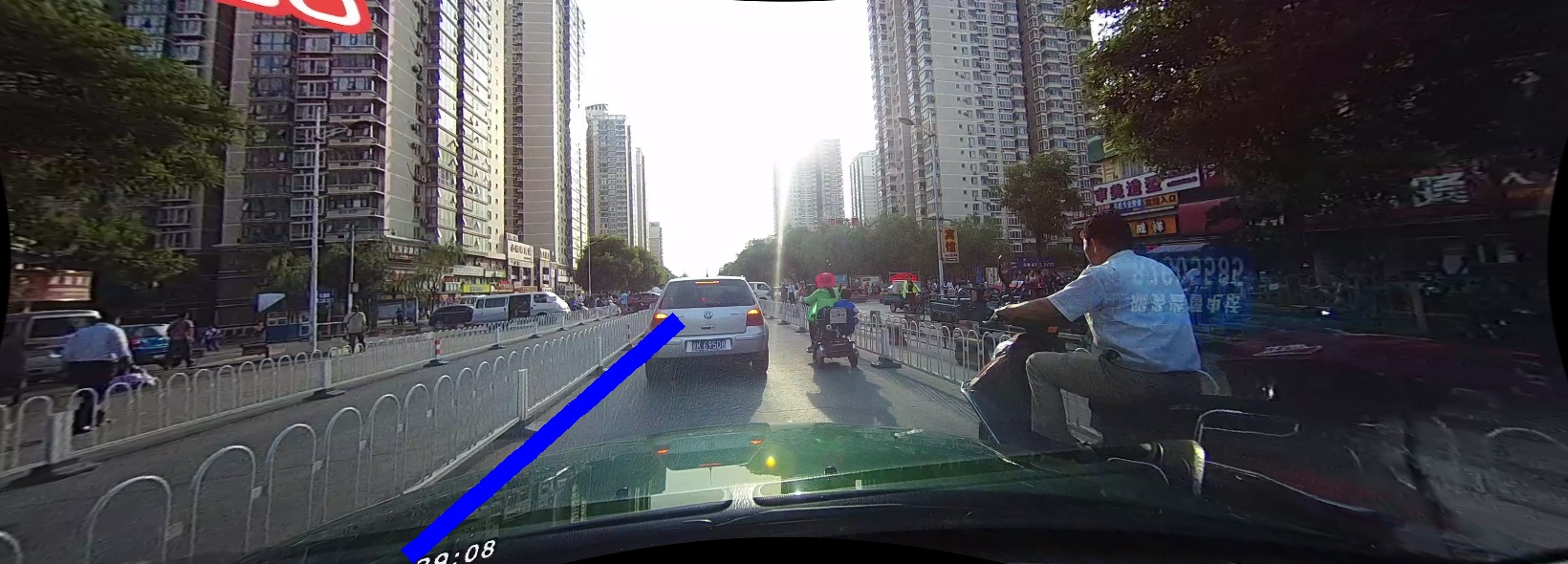}\hfill
\end{minipage}\hfill \\
\begin{minipage}[b]{\linewidth}
  \rotatebox{90}{\parbox[c][0.02\linewidth][c]{0.08\linewidth}{\centering \self}}
  \includegraphics[width=0.242\linewidth]{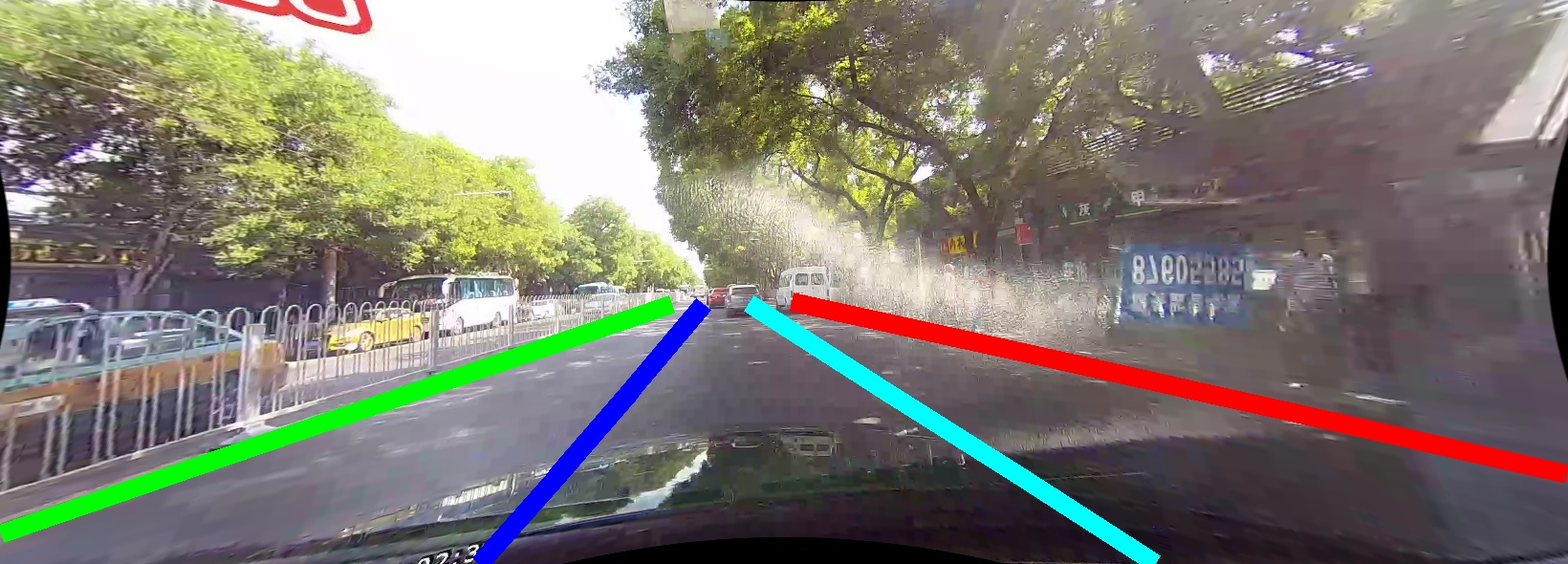}\hfill
  \includegraphics[width=0.242\linewidth]{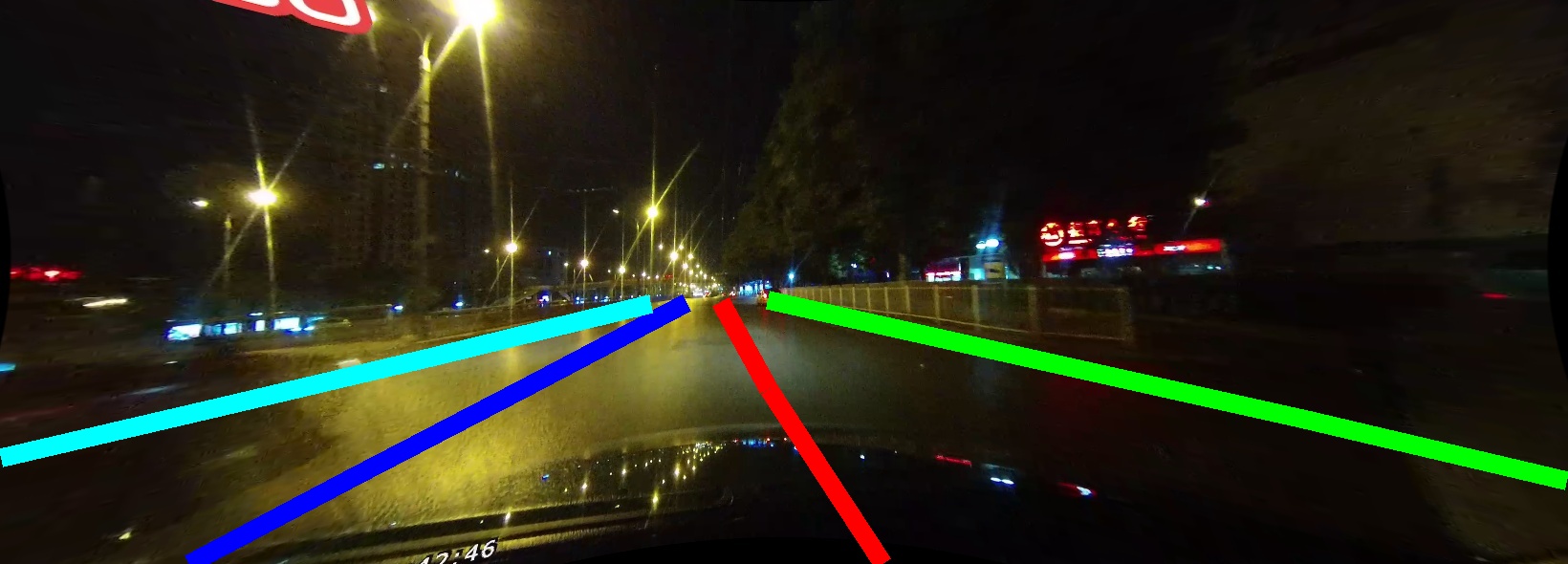}\hfill
  \includegraphics[width=0.242\linewidth]{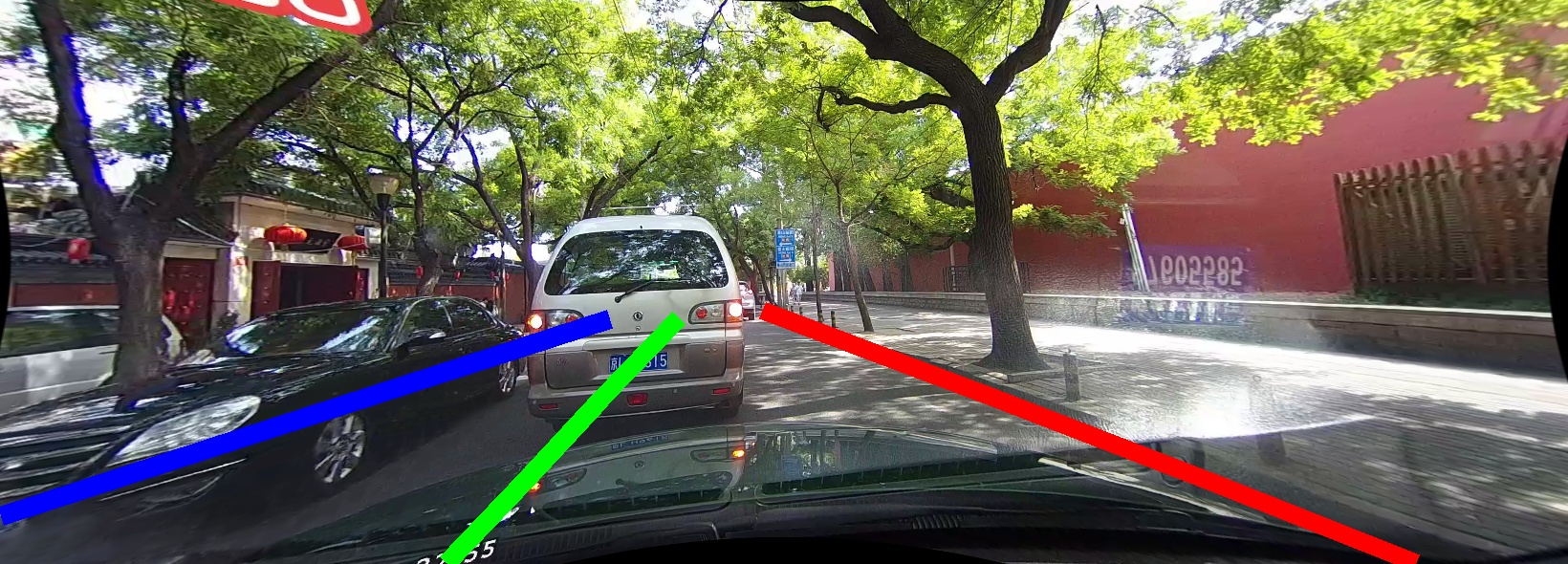}\hfill
  \includegraphics[width=0.242\linewidth]{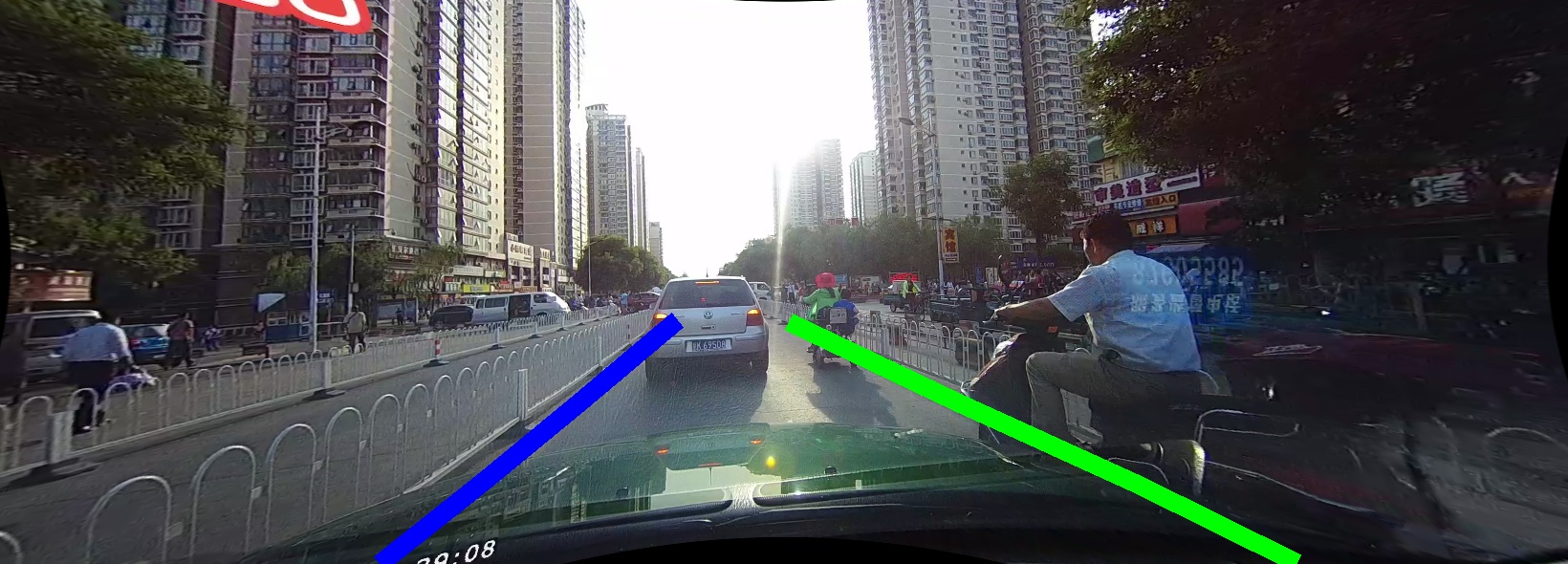}\hfill
\end{minipage}\hfill \\
\begin{minipage}[b]{\linewidth}
  \rotatebox{90}{\parbox[c][0.02\linewidth][c]{0.08\linewidth}{\centering GT}}
  \includegraphics[width=0.242\linewidth]{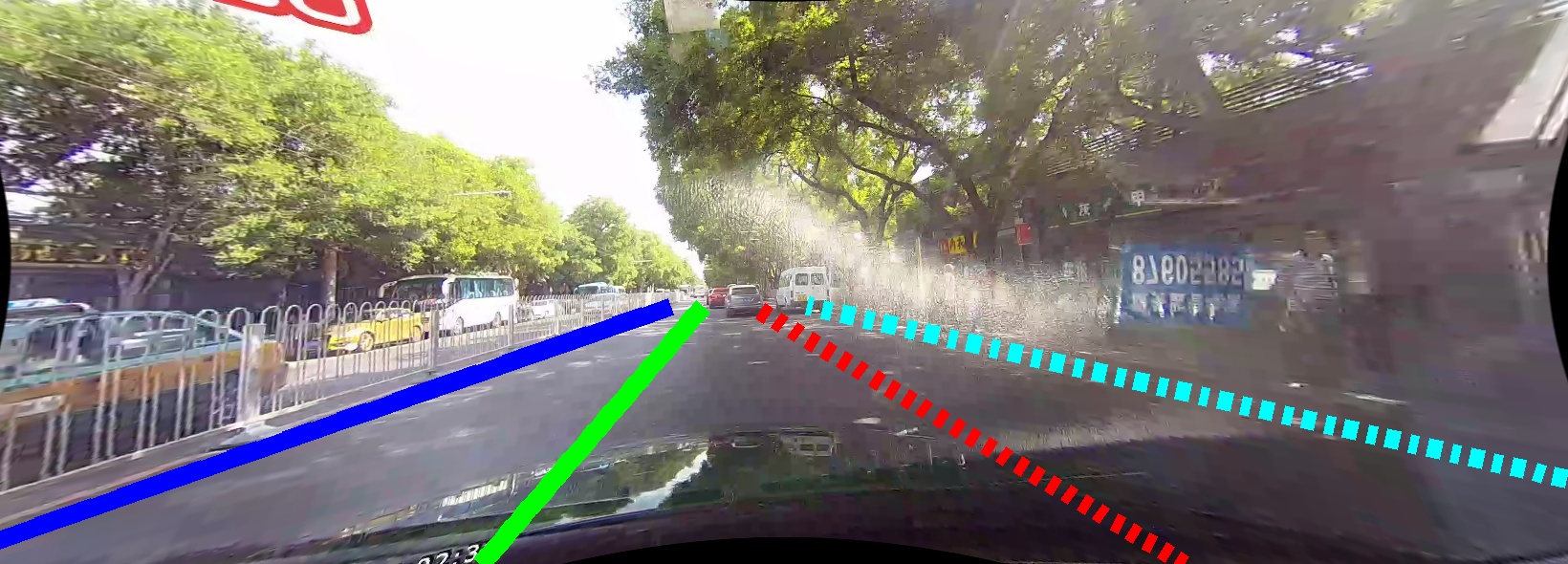}\hfill
  \includegraphics[width=0.242\linewidth]{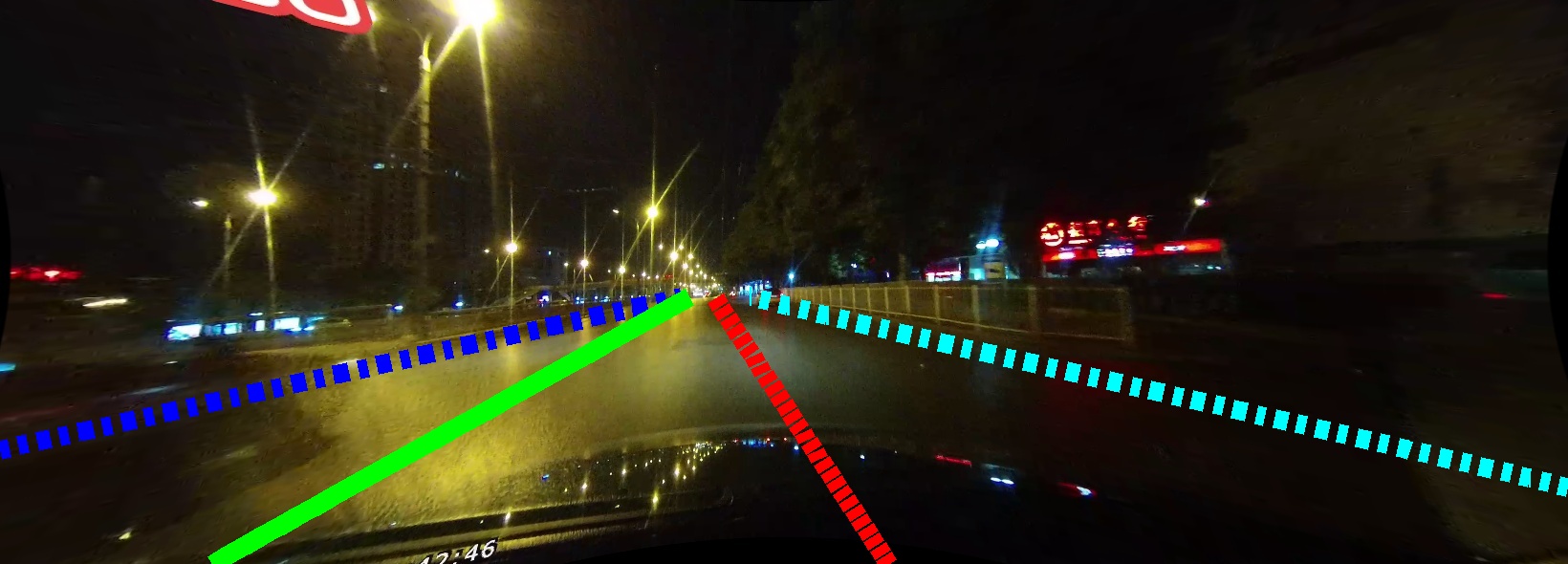}\hfill
  \includegraphics[width=0.242\linewidth]{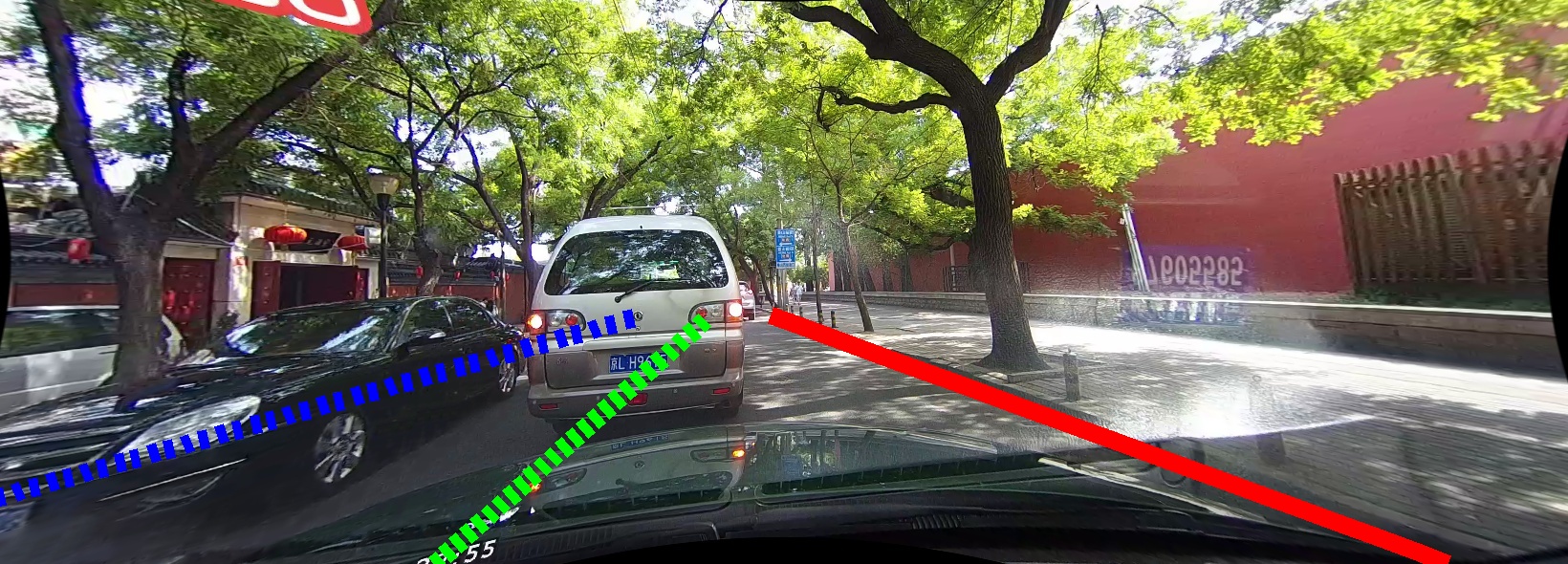}\hfill
  \includegraphics[width=0.242\linewidth]{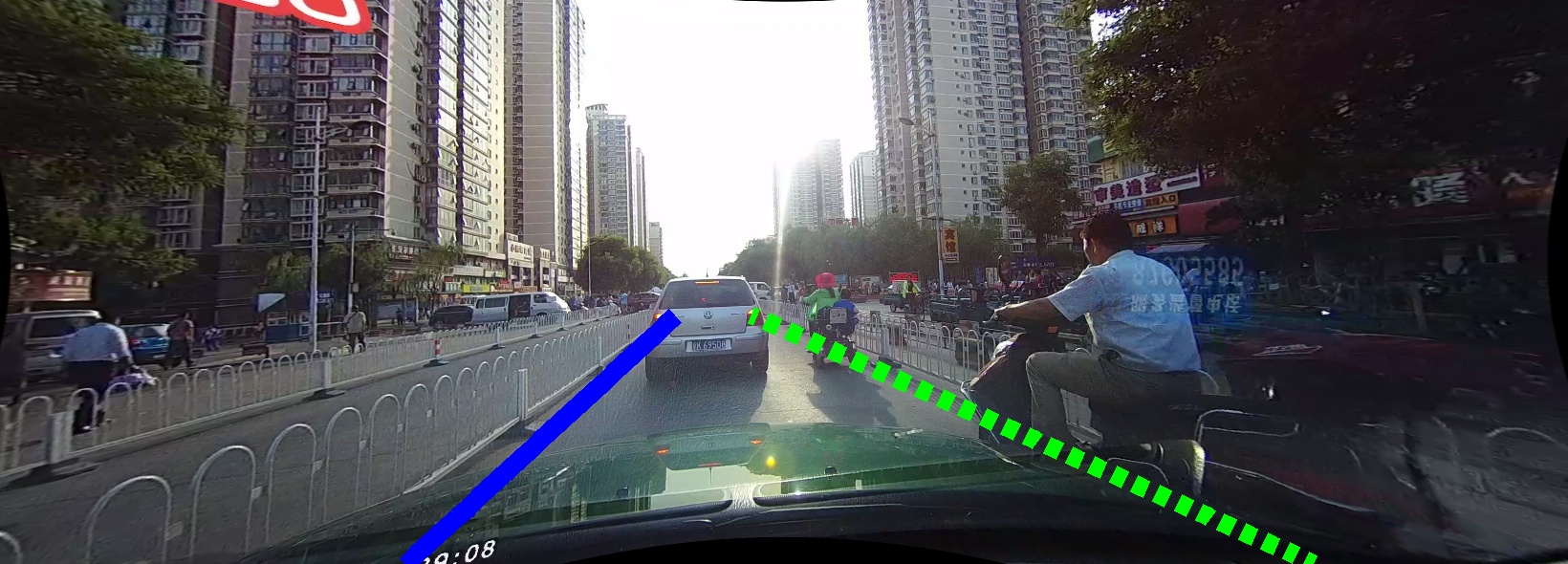}\hfill
\end{minipage}\hfill \\
\caption{Lane prediction results using the current state-of-the-art method (CLRNet) and \self for cases with
  little- or no-visual-clue, lens flare, weak lighting, occlusion, and
  hidden lines. CLRNet misses certain lanes, while \self correctly
  finds all instances. Examples with ground truth are from the CULane dataset.}
\label{fig:no-visual-clue}
\end{figure*}

\begin{figure*}[!tb]
\centering
\begin{minipage}[b]{\linewidth}
  \rotatebox{90}{\parbox[c][0.02\linewidth][c]{0.08\linewidth}{\centering CANet}}
  \includegraphics[width=0.242\linewidth]{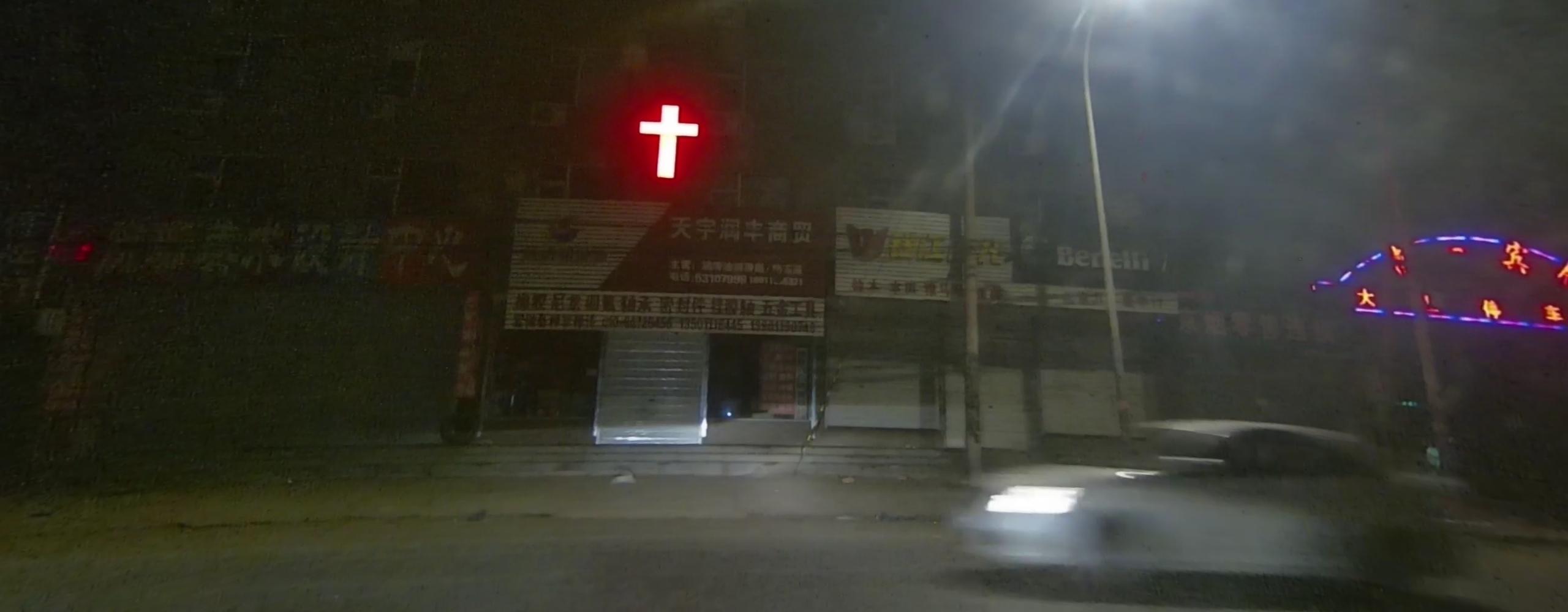}\hfill
  \includegraphics[width=0.242\linewidth]{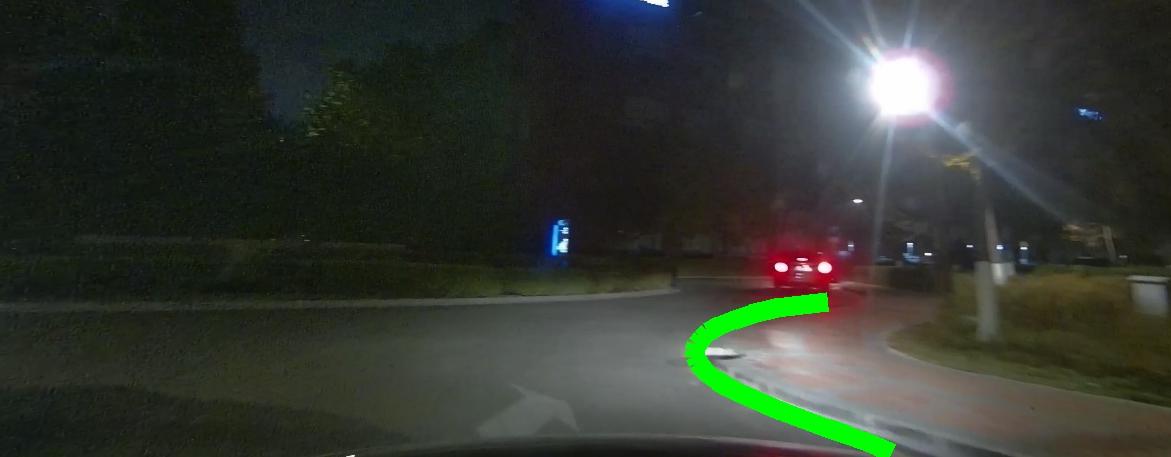}\hfill
  \includegraphics[width=0.242\linewidth]{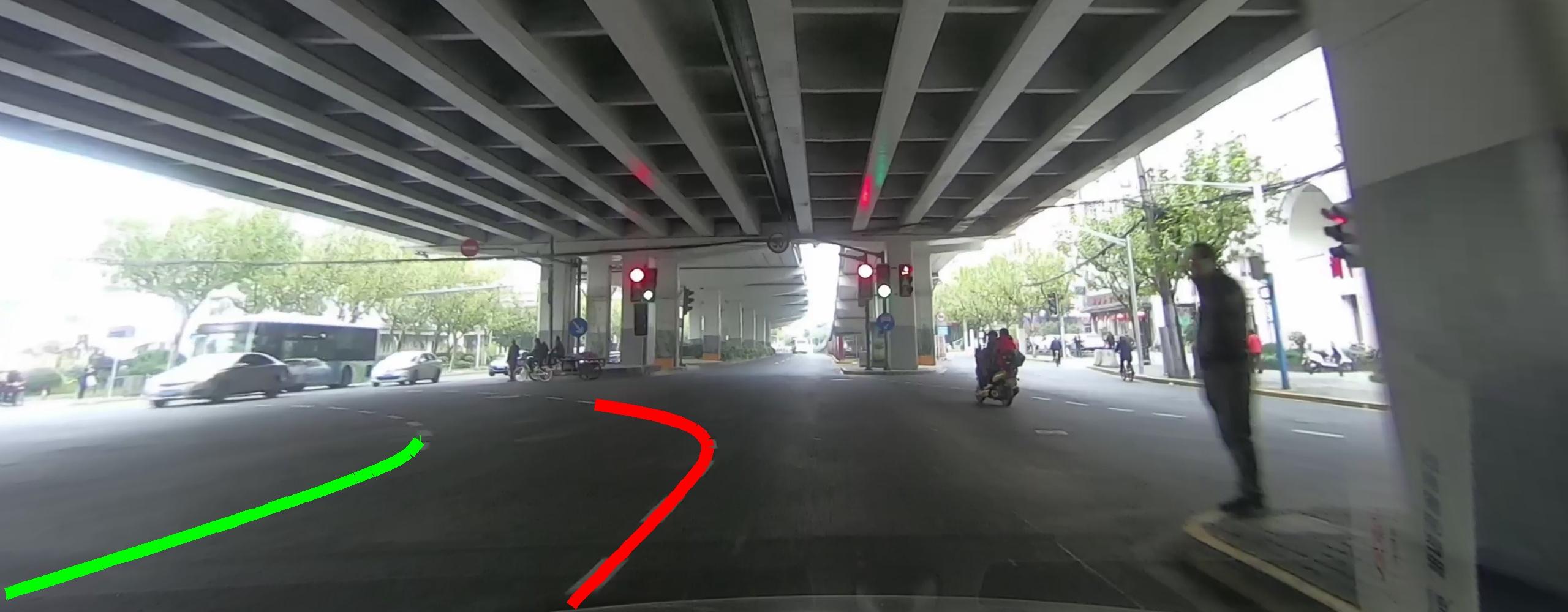}\hfill
  \includegraphics[width=0.242\linewidth]{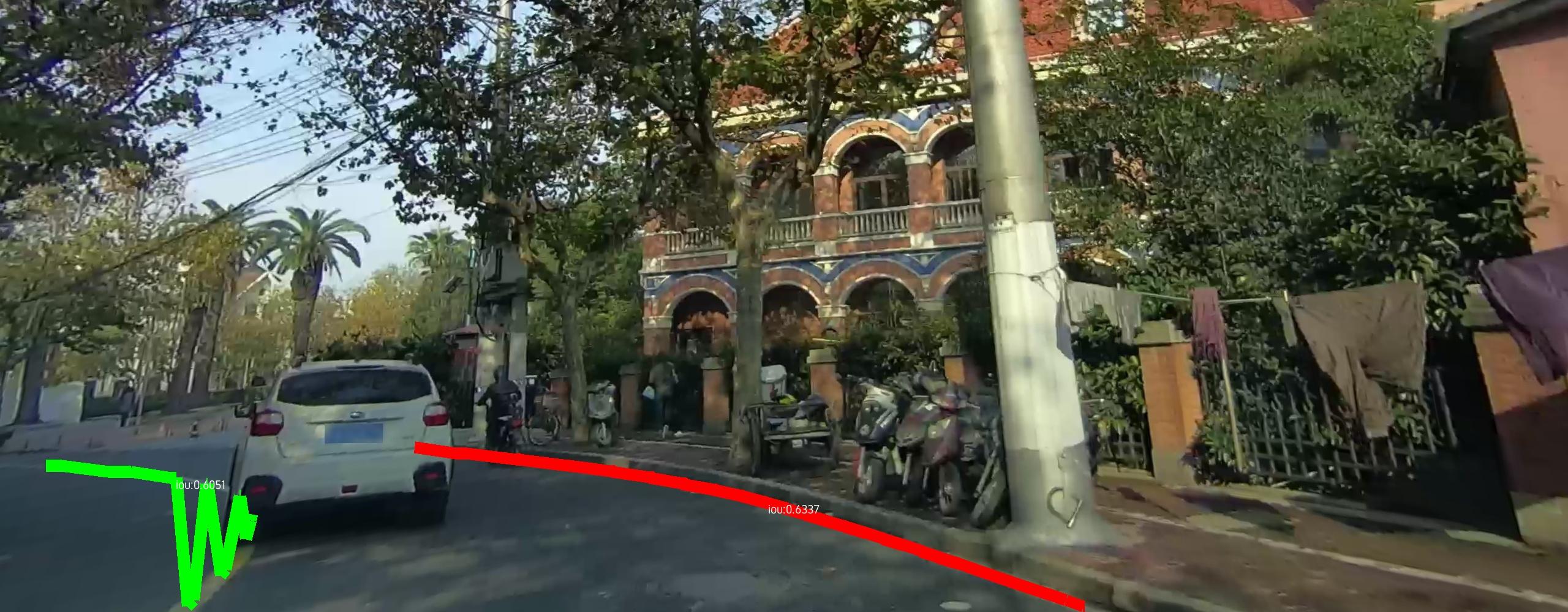}\hfill
\end{minipage}\hfill \\
\begin{minipage}[b]{\linewidth}
  \rotatebox{90}{\parbox[c][0.02\linewidth][c]{0.08\linewidth}{\centering \self}}
  \includegraphics[width=0.242\linewidth]{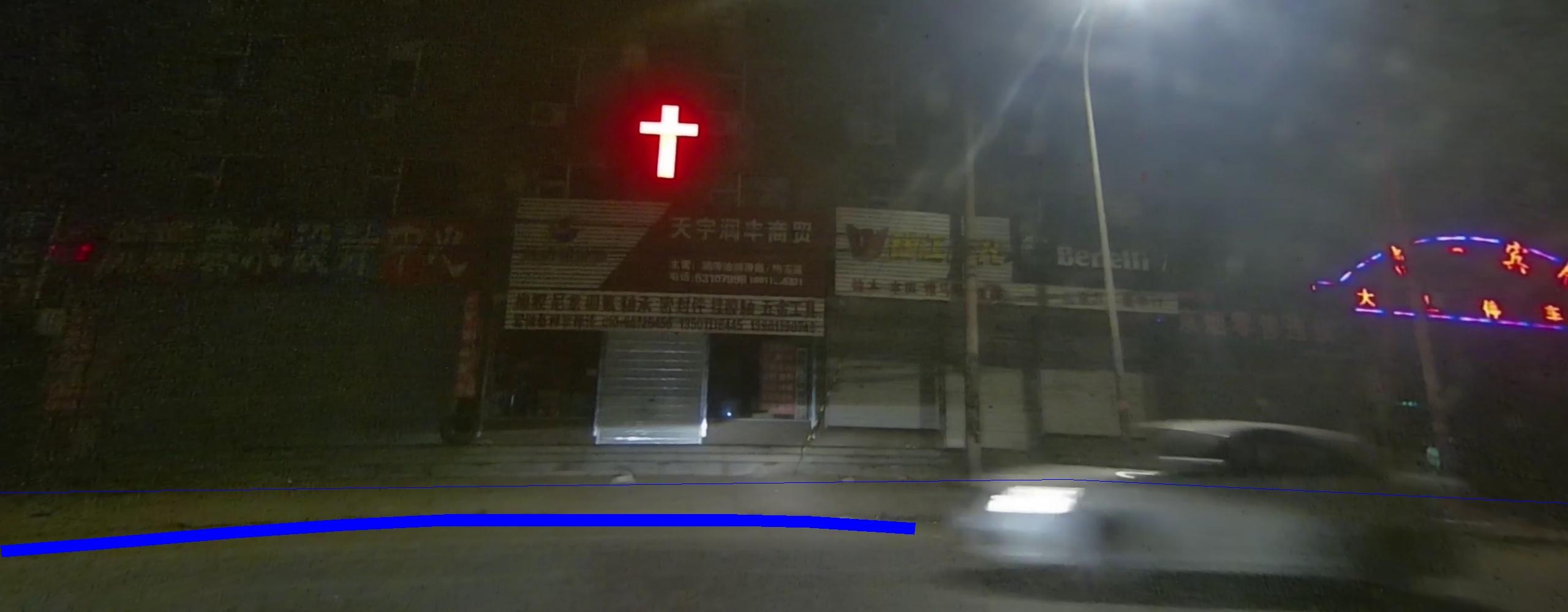}\hfill
  \includegraphics[width=0.242\linewidth]{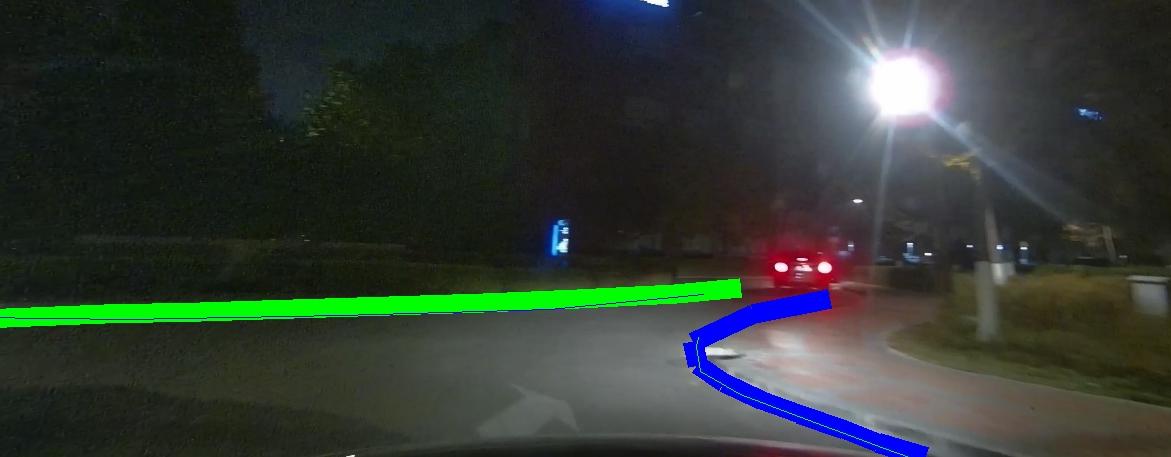}\hfill
  \includegraphics[width=0.242\linewidth]{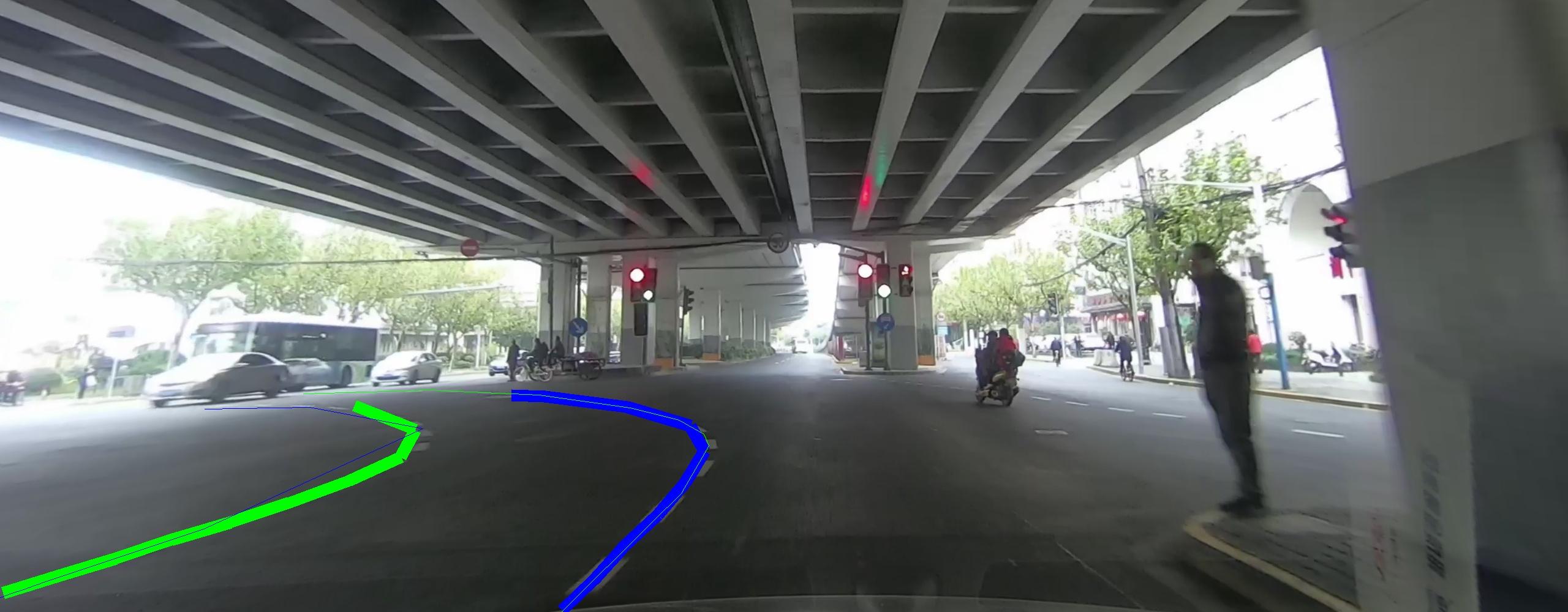}\hfill
  \includegraphics[width=0.242\linewidth]{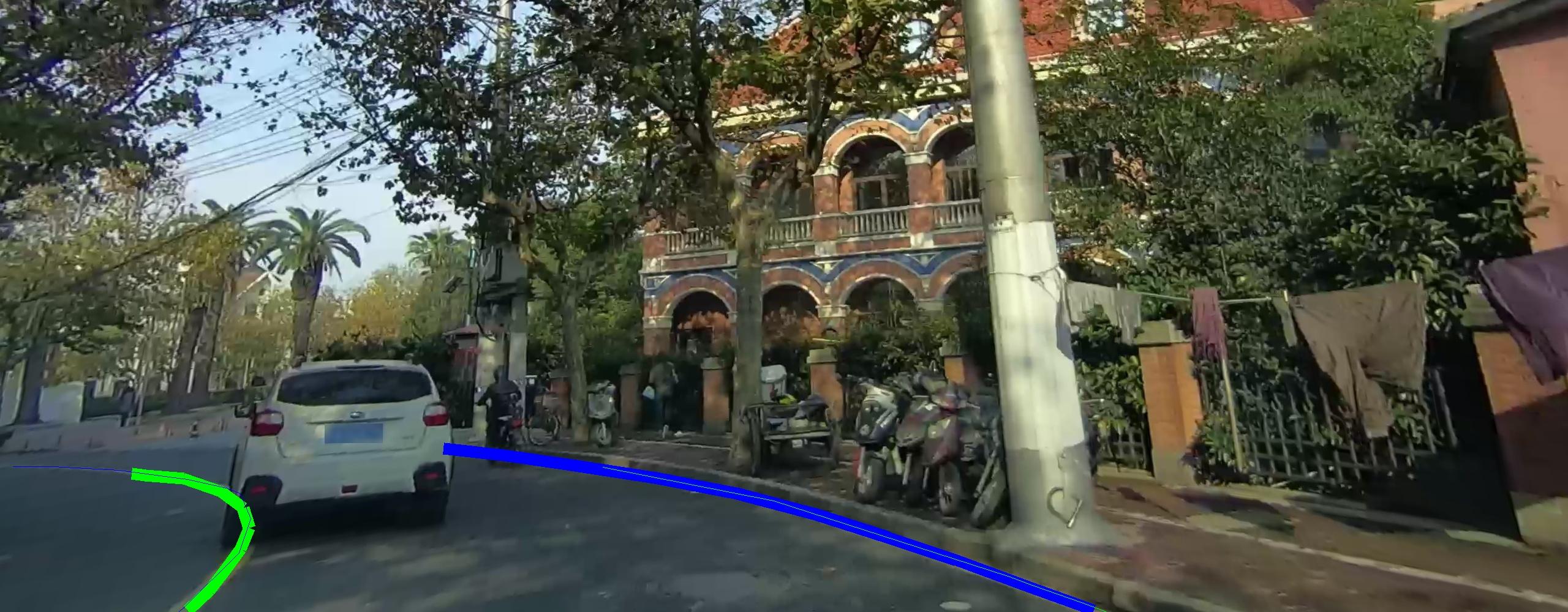}\hfill
\end{minipage}\hfill \\
\begin{minipage}[b]{\linewidth}
  \rotatebox{90}{\parbox[c][0.02\linewidth][c]{0.08\linewidth}{\centering GT}}
  \includegraphics[width=0.242\linewidth]{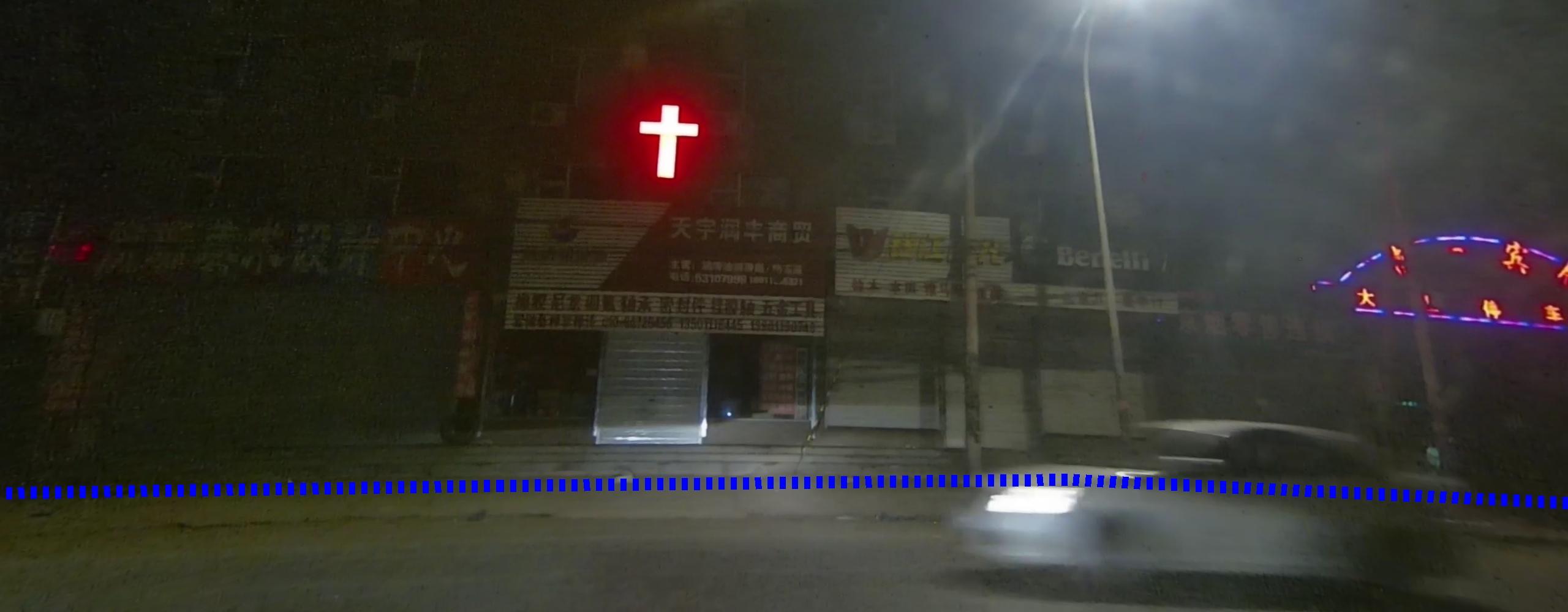}\hfill
  \includegraphics[width=0.242\linewidth]{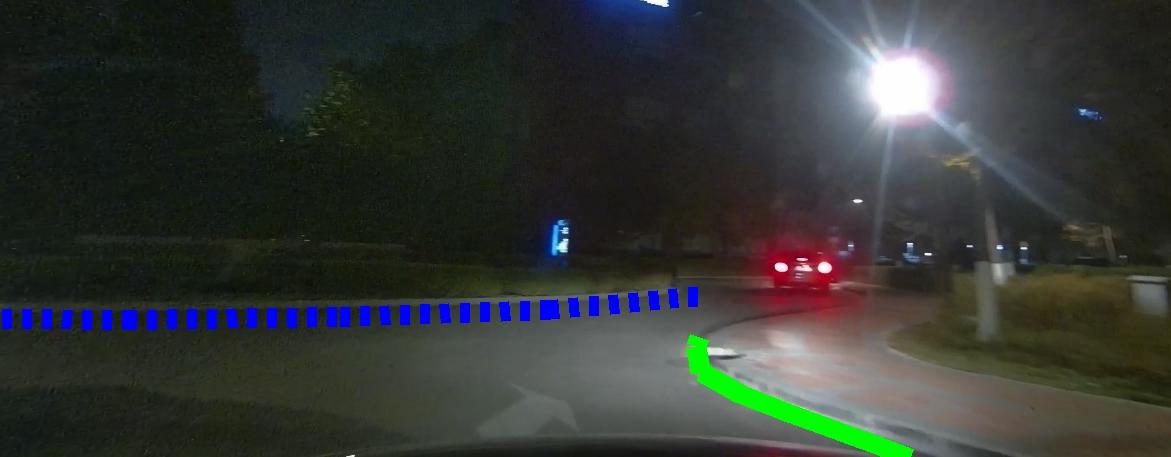}\hfill
  \includegraphics[width=0.242\linewidth]{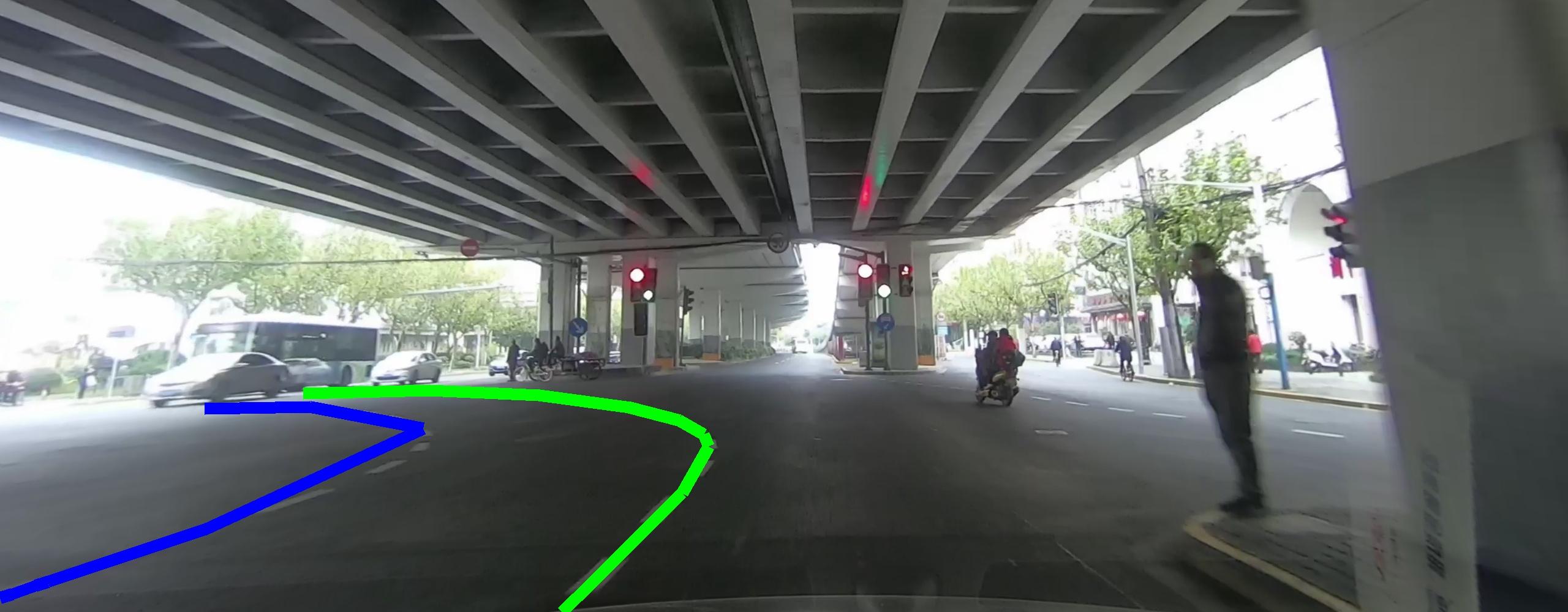}\hfill
  \includegraphics[width=0.242\linewidth]{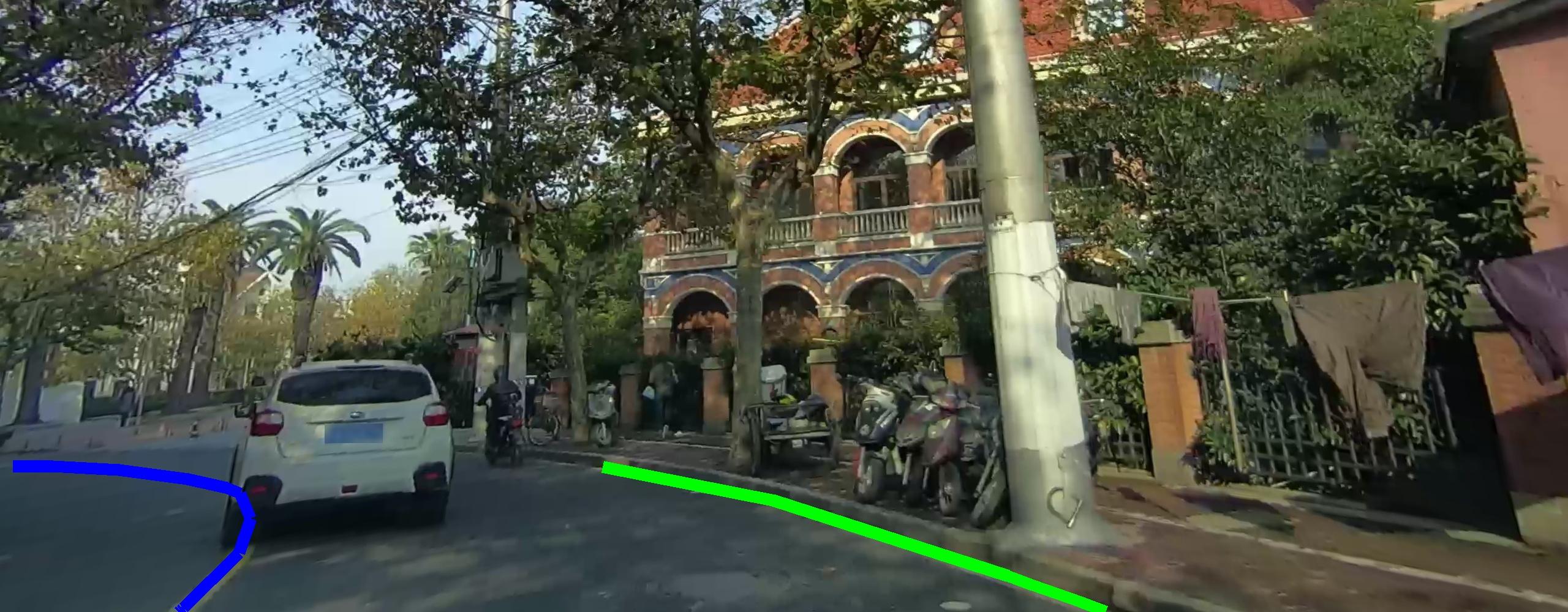}\hfill
\end{minipage}\hfill \\
\caption{Lane prediction results for CANet and \self for examples from the
  CurveLanes dataset. Limited by its lane representation, CANet cannot describe
  lanes in special cases like T-junctions, roundabouts, waiting areas, and sharp
  turns, while \self can address them all.}
  \label{fig:need-chain}
\end{figure*}

Some research leverages semantic segmentation~\cite{scnn,laneaf} to identify
lanes by classifying pixels or picking up keypoints. However, it is
difficult to separate different instances from the lane foreground produced by
semantic segmentation for lanes that are very close to each other. Instead, 
other research turns to a top-down approach. LaneATT~\cite{laneatt} first
predicts numerous candidates, then post-processes them with non-maximum
suppression (NMS). However, NMS struggles to accurately distinguish adjacent lanes, 
leading to false deletions. CondLaneNet~\cite{condlanenet} and
CANet~\cite{canet} obtain lane instances by detecting keypoint responses on
heatmaps, but due to the local perception characteristics of CNNs, keypoints
often respond weakly when visual features are far away from them, making them
prone to detection failure. HoughLaneNet~\cite{houghlanenet}
leverages DHT-based feature aggregation to detect lanes with weak visual
features but is only applicable to straight lanes. To mitigate these
challenges, recent studies employing transformer-based models~\cite{lstr,
priorlane,laneformer,o2sformer,detr_condlane} utilize a global attention 
mechanism to derive implicit semantic insights.
O2SFormer~\cite{o2sformer} proposed one-to-several label assignment to address
label semantic conflict. Chen~\cite{detr_condlane} improved the convolutional
kernel generation network of CondLaneNet with transformer:
 the object query, after being processed by attention calculation, results
in generated kernels possessing more global information. However, these methods
do not thoroughly utilize target locations to focus attention. Therefore, a 
substantial amount of the attention computation is squandered on the background, 
which is irrelevant to the target objects, thereby restricting the models' ability 
to effectively discern and focus on the lanes. Fig.~\ref{fig:no-visual-clue} illustrates 
challenging cases, particularly those with few or no visual cues. These cases 
are crucial for common downstream tasks, such as lane keeping and map-based lane 
information collection.

Additionally, current lane representations are unsuitable for cases like those in
Fig.~\ref{fig:need-chain}. Existing methods usually rely on handcrafted 
post-processing rules based on prior assumptions. Almost all
methods~\cite{laneaf, folo, ganet, ultra_fast,
 condlanenet,o2sformer,detr_condlane} assume that lanes extend from the bottom
of the image upwards, so fail to detect horizontal lanes like those in
the first three columns of Fig.~\ref{fig:need-chain}. Although
CANet used an adaptive post-processing decoder to avoid such
assumptions, it still cannot handle cases like the one in the last column 
of Fig.~\ref{fig:need-chain} due to its reliance on manual settings. 
It is crucial to develop improved representations that can accurately 
capture such lanes, as they are commonly encountered in everyday driving.
% in everyday driving 有点奇怪，感觉得表达成自动驾驶里的基本功能？

To tackle the aforementioned problems, we propose a novel top-down end-to-end
lane detection network, termed \self \emph{(lane detection 
transformer)}, based on the transformer architecture. 
Specifically, we propose a novel \emph{anchor-chain}
to represent the shapes of lanes and two new loss functions to supervise their
overall trend and detailed descriptions. Moreover, to enhance \self's ability and
efficiency during deep semantic information extraction, a novel \emph{multi-referenced
deformable cross-attention} (MRDA) algorithm is applied in the transformer decoder. 
Additionally, we incorporate auxiliary branches to extract more fine-grained target information.
As the second row of Fig.~\ref{fig:need-chain} shows, \self demonstrates
remarkable performance in handling diverse challenging scenarios. 
Extensive experimental results on multiple datasets~\cite{scnn, curvelanes} 
show that \self achieves state-of-the-art performance.

The main contributions of this paper are as follows:
\begin{itemize}
  \item A new lane representation method \emph{anchor-chain} is
        proposed. It conceptualizes lanes as a holistic entity of 
        interconnected nodes, in contrast to the conventional approach of 
        representing them as isolated dots (pixels or keypoints). This 
        allows us to bypass the need for rule-based post-processing and 
        the complexities associated with handling intricate lane geometries. 
        Moreover, the anchor-chain requires only a few points to
        denote the important turning points of a lane, so is efficient
        and precise.
  \item A \emph{multi-referenced deformable attention module} is proposed to
        transmit the position prior information contained in the anchor-chain to the
        network, evenly distributing attention around the targets. This,
        combined with the global semantic information extracted by the Encoder,
        enhances the model's perception ability toward the targets in case of
        little- or no-visual-clue.
  \item Two line IoU algorithms are devised, namely the \emph{point-to-point}' (P2P)
        IoU and the \emph{dense-sampling} (DS) IoU. They are applied in binary matching
        cost and loss during training, respectively. Compared to the traditional
        point-to-point L1 distance, the new algorithms introduce global
        optimization to improve training efficiency and inferencing performance.
  \item We evaluate \self using typical metrics,
        parameterized F1-score as well as synthetic metrics, on public datasets.
        Experimental results demonstrate that \self outperforms other methods overall.
\end{itemize}

\section{Related Work}
\label{sec:related-work}

We relate our work to both the existing lane detection approaches as well as to general
object detection methods.

\subsection{Lane Detection}
\label{sec:lane-detection}
Deep learning-based lane detection algorithms can be boiled down into two main
categories: \emph{bottom-up} and \emph{top-down}.

Bottom-up methods cluster or classify lane pixels or keypoints.
Pixel-level segmentation~\cite{scnn, lanenet, laneaf}, evolving from general
visual semantic segmentation, first extracts foreground lane pixels using
semantic segmentation and then clusters or classifies them using techniques like
pixel embedding to differentiate lane instances. In contrast,
keypoint-based detection methods~\cite{folo, ganet} can be regarded as sparse
versions of segmentation models that replace dense pixel classification with
discrete keypoints, which partially alleviates the problem of excessive focus on
segmentation boundaries in pixel-level segmentation. However, bottom-up methods
are usually unable to handle branching or merging lanes, and to accurately
distinguish adjacent boundaries of multiple closely located lanes. \self is
designed to address all such challenges.

Top-down methods first obtain target instances and then refine the
representation of the shape for each instance. Basically, there are three major
categories based on the lane representation: parameterized curve
fitting~\cite{lstr, bezierlanenet}, a tilted anchor~\cite{laneatt, o2sformer,
 clrnet, RVLD}, and row-wise classification~\cite{condlanenet, ultra_fast,
 houghlanenet, detr_condlane}.

Curve fitting models complex lanes as simple polynomial curves, which can be
very efficient due to the small number of descriptors that need to be predicted.
However, it is difficult for such curves to match sophisticated lanes in the real world,
leading to poor precision and flexibility. Tilted anchor based methods obtain
a large number of proposals by placing dense anchors, then filter out
unqualified and overlapping instances through NMS in
a post-processing stage. However, double solid lines and adjacent lanes are
placed close together under specific perspectives and may be incorrectly
deleted by NMS during deduplication, leading to critical information loss. In
contrast, row-wise classification improves efficiency based on the observation
that often, lanes appear vertically. Unfortunately, this also prevents its application to lanes that appear almost horizontally, like high-angle
lanes~\cite{canet} and U-turn lanes.

Unlike these three approaches, \self uses set prediction to
distinguish lanes that are close together; its anchor ensures precise
details while performing well in various complicated cases, such as U-turns,
T-junctions, and roundabouts, commonly found in real-world situations and that must
be addressed.

\begin{figure*}[t]
  \centering
  \includegraphics[width=\linewidth]{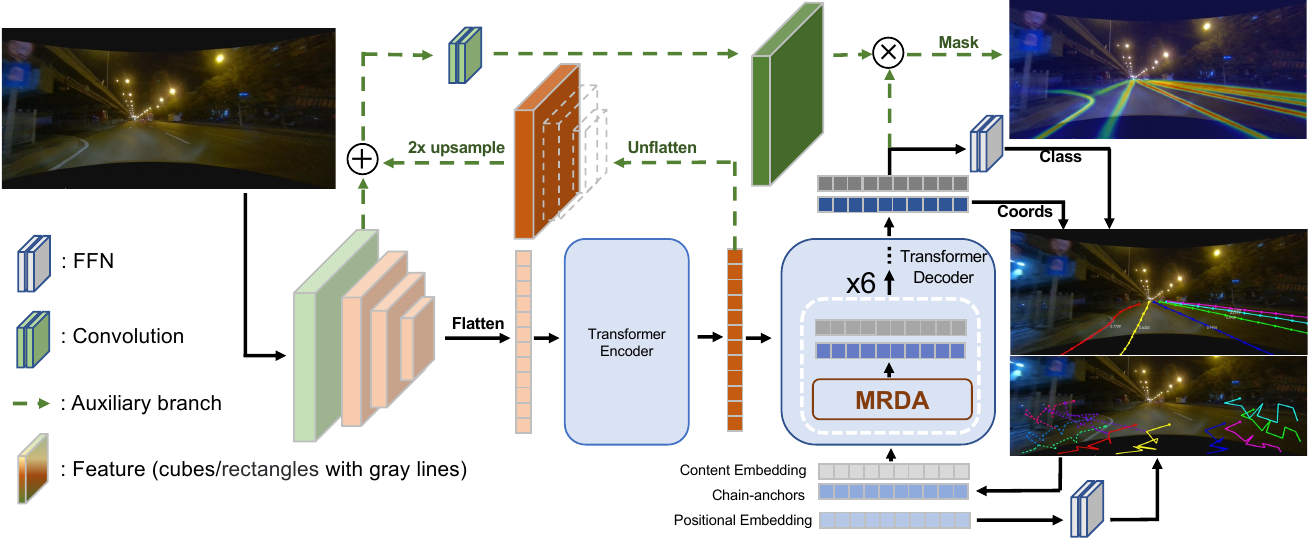}
  \caption{\self follows the structural paradigm of DETR. After 2D image
    features are extracted by the backbone, \self further extracts deep semantic
    information in the encoder through the self-attention mechanism. The input
    object queries to the decoder are composed of content embeddings and
    anchor-chains. In the computation of each decoder layer, the object queries
    update themselves through MRDA and interact with image features, including
    the correction of anchor-chains and differentiation of positive or negative
    objects. After 6 iterative updates, the positive anchor-chains are able to
    represent lane shapes accurately. Additionally, \self introduces a
    Gaussian heatmap auxiliary branch to enhance the ability of the object
    query to perceive lane details.}
  \label{fig:ldtr-pipeline}
\end{figure*}

\begin{figure*}[!t]
  \centering
  \subfigure[Original image.]{\label{fig:lane-rep-ori-img}%
    \includegraphics[width=.36\linewidth]{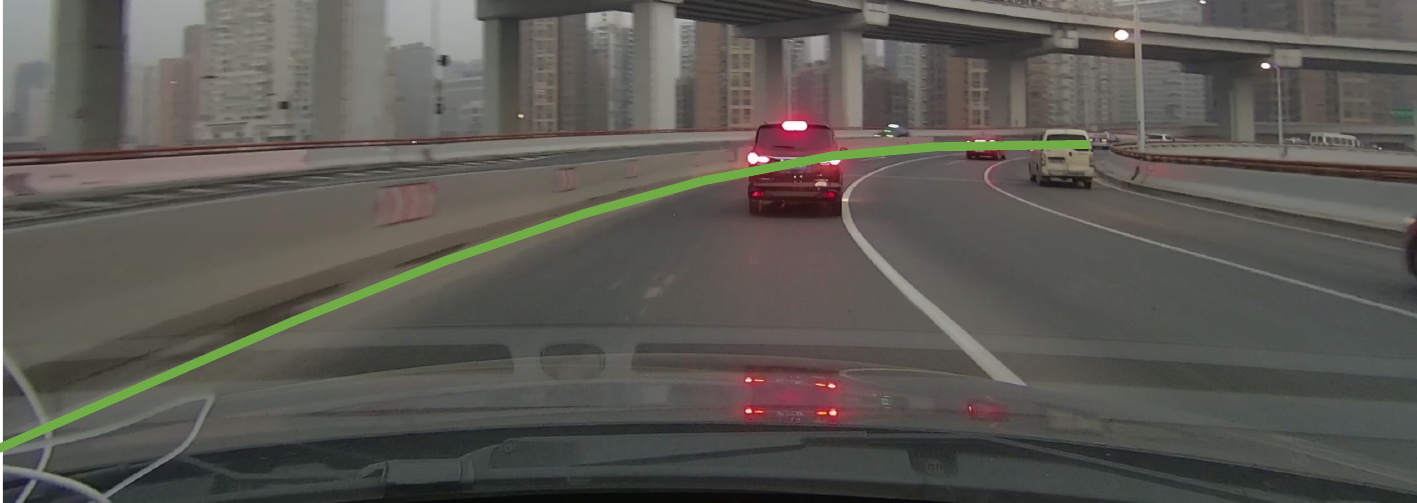}}\quad
  \subfigure[Tilted anchor.]{\label{fig:lane-rep-anchor}%
    \includegraphics[width=.36\linewidth]{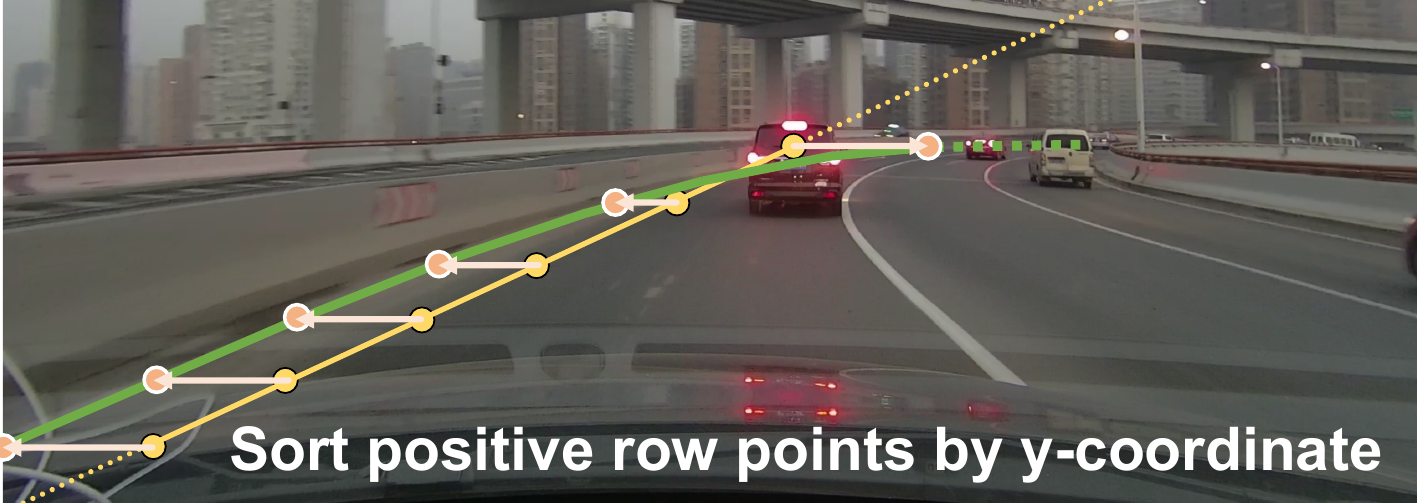}}\\
  \subfigure[Keypoints/Segmentation.]{\label{fig:lane-rep-kps}%
    \includegraphics[width=.36\linewidth]{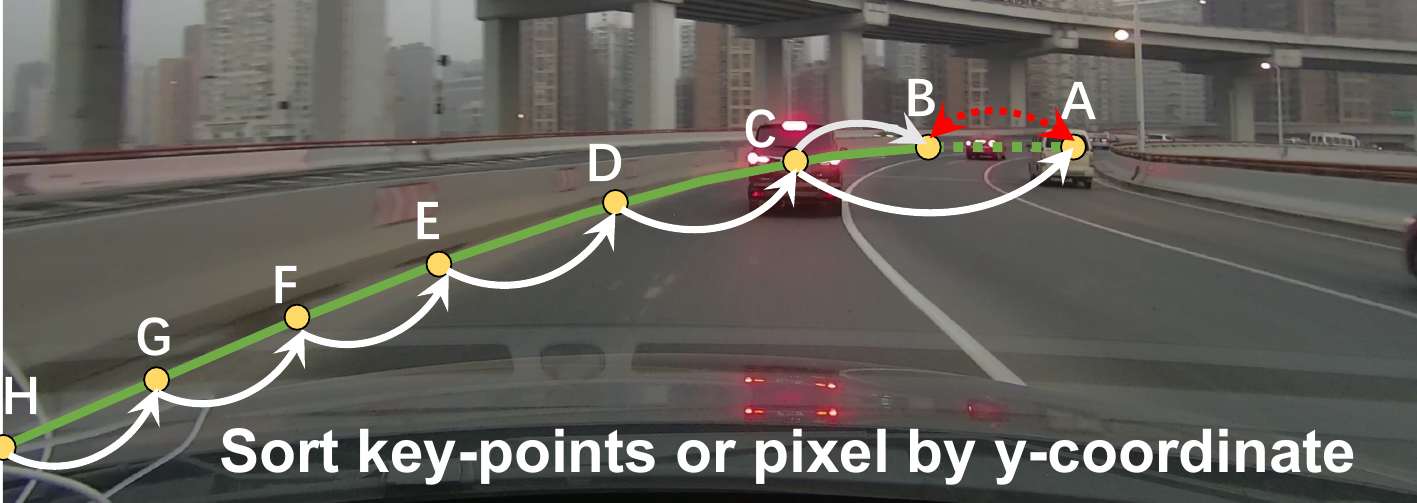}} \quad
  \subfigure[Anchor-chain.]{\label{fig:lane-rep-ca}%
    \includegraphics[width=.36\linewidth]{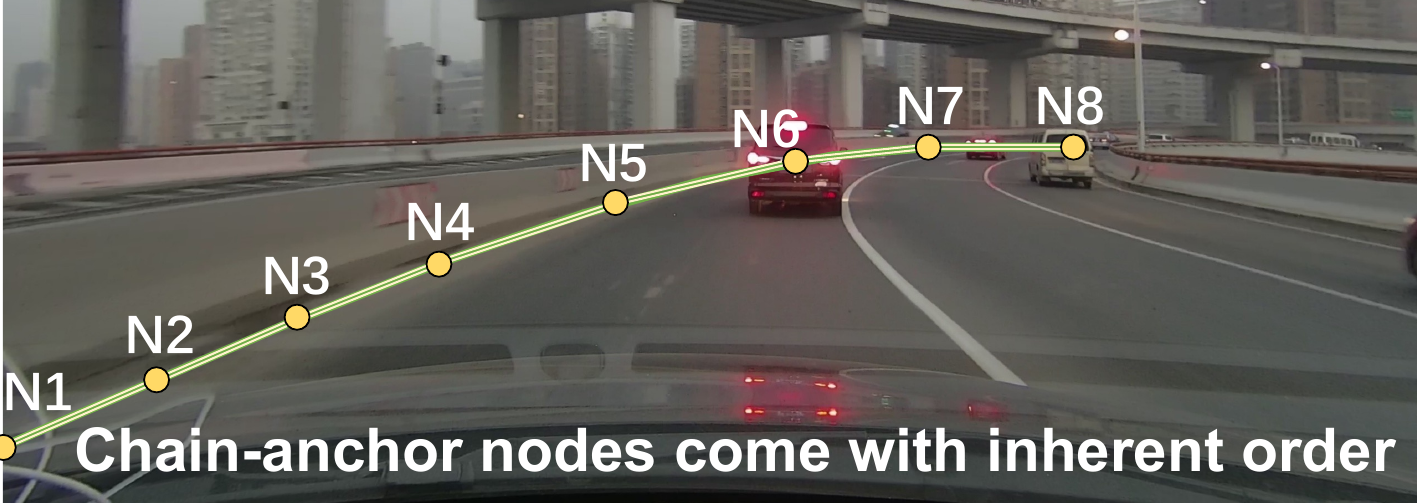}}%
  \caption{Various lane representations. It is hard for current methods to
    represent horizontal parts of lanes, but easy for our anchor-chain.}
  \label{fig:lane-reps}
\end{figure*}

\subsection{Object Detection}
\label{sec:object-detection}
Object detection is closely related to lane detection, and many of its ideas and
techniques can be leveraged directly. Early CNN-based
methods~\cite{faster-rcnn, yolo, yolov3, centernet, fcos} mostly required
rule-based post-processing operations, which can lead to poor performance
in some uncommon scenarios. DETR~\cite{detr} proposed a new end-to-end 
paradigm for object detection but suffers from long training
iterations and high computational cost. DETR inspired much following
research. Deformable DETR~\cite{deformable_detr} transformed the dense attention
operation in the original cross-attention mechanism into a more efficient sparse
attention mechanism by reference point sampling, reducing computational cost
while improving model training convergence speed. DAB-DETR~\cite{dabdetr}
explicitly modeled the object query as an anchor, using the position of the
bounding box to guide attention focus near the target, which further optimizes
 training and improves model performance.

The global attention mechanism in the DETR paradigm equips the model with
a powerful semantic awareness capability, which helps to improve model
performance in cases with few or no visual clues. Based on the
structures and optimization techniques of the DETR-family, this paper presents
an end-to-end lane detection model, \self.

\section{Method}
\label{sec:methods}

\subsection{Network Architecture}
\label{sec:network-architecture}

As a transformer-based model, \self is inspired by the DETR architecture
as shown in Fig.~\ref{fig:ldtr-pipeline}. The black solid line indicates the
lane prediction process. Firstly, \self takes a front view as the network input
and extracts features at different levels through a backbone network composed of
multiple CNN layers. The high-level features are reduced to one dimension and
input into the transformer encoder for further interaction and output. Secondly,
the transformer decoder takes a small fixed number of learnable content embeddings
and anchor-chains (see Section~\ref{sec:anchor-chain}) as input object queries, computes
MRDA (see Section~\ref{sec:multi-ref-deform-cross-attention}) with the
output of the encoder, and outputs modified content embeddings and anchors.
Finally, the modified content embeddings are passed to a shared parameter
feed-forward network (FFN), which predicts the presence or absence of targets,
and the anchor-chain accurately describes target positions.

\begin{figure*}[t]
  \centering
  \subfigure[Manually annotated ground-truth (GT)]{\label{fig:ca-manual-anno}%
    \includegraphics[width=.26\linewidth]{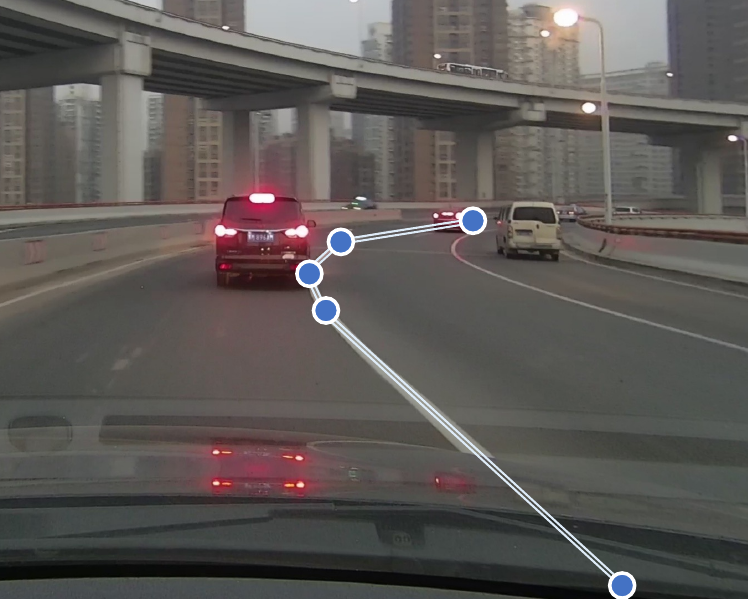}}\quad
  \subfigure[Uniformly sampled GT]{\label{fig:ca-sample-anno}%
    \includegraphics[width=.26\linewidth]{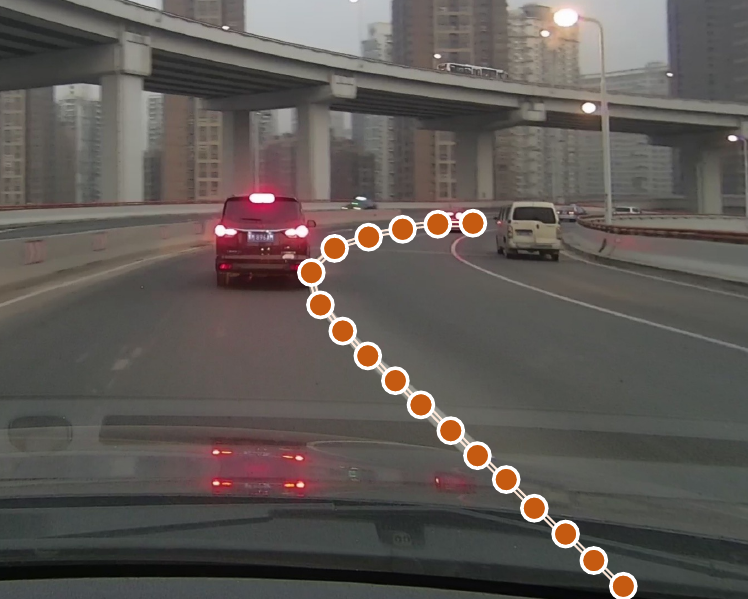}}\quad
  \subfigure[Initial anchor-chain]{\label{fig:ca-current-pred}%
    \includegraphics[width=.26\linewidth]{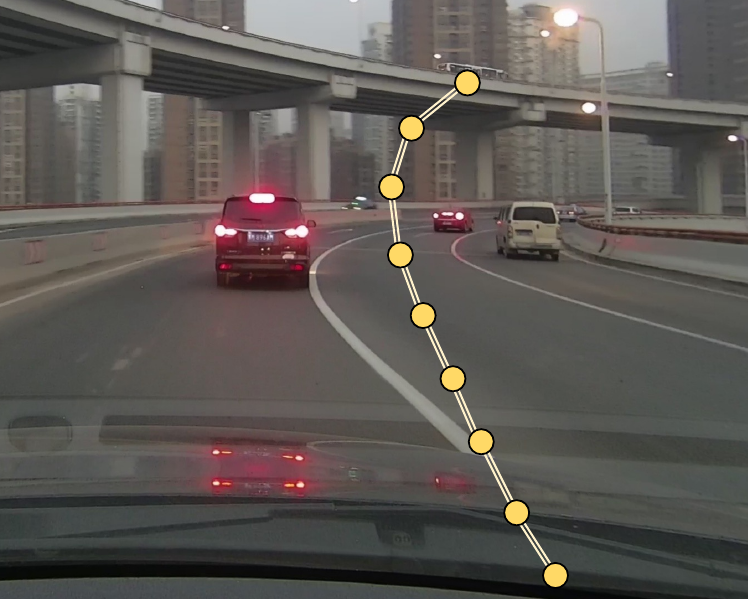}}

  \subfigure[Anchor-chain with manual GT]{\label{fig:ca-super-step1}%
    \includegraphics[width=.39\linewidth]{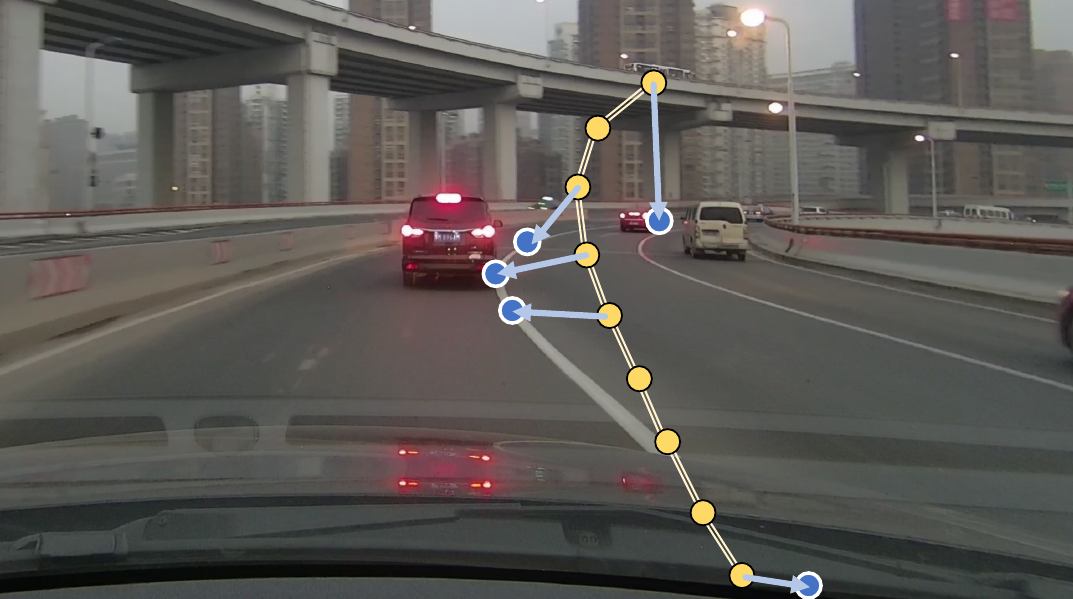}}\quad\quad
  \subfigure[Anchor-chain with uniform sampled GT]{\label{fig:ca-super-step2}%
    \includegraphics[width=.39\linewidth]{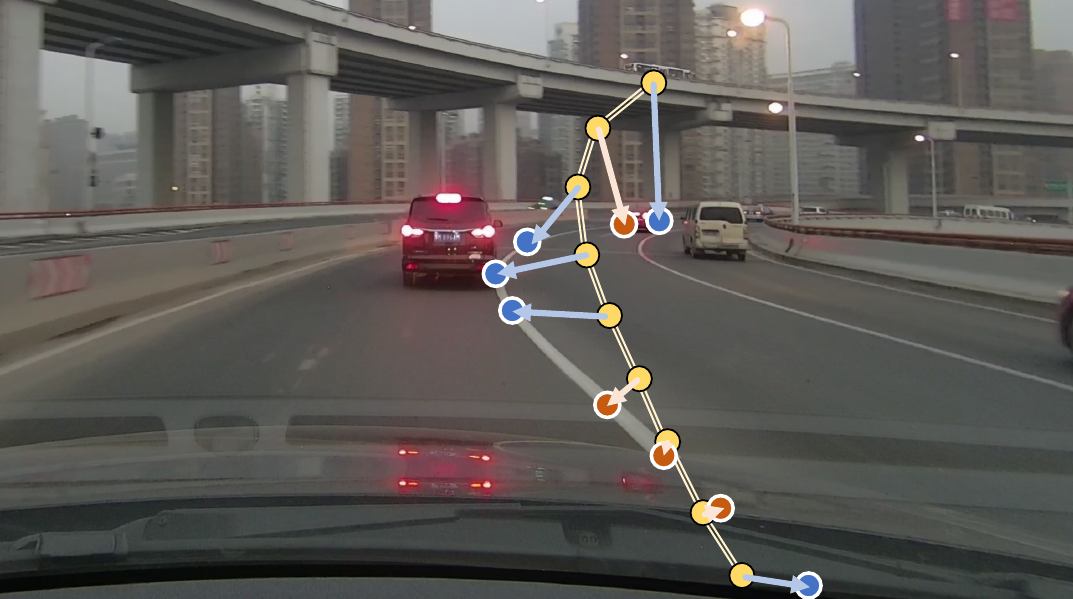}}
  \caption{The regression supervision approach of the anchor-chain enables it to
    efficiently utilize a small number of nodes to accurately describe curves, essentially
    similar to how humans recognize lanes.}
  \label{fig:anchor-chain}
\end{figure*}

\subsection{Anchor-chain}
\label{sec:anchor-chain}

Various lane representations are available, such as tilted anchor lines (see
Fig.~\ref{fig:lane-rep-anchor}~\cite{laneatt}), row-wise
classification~\cite{ultra_fast, condlanenet} predicting a set of points and
sorting them by y-coordinates~\cite{folo, ganet}, and keypoints with adaptive
decoders~\cite{canet} (see Fig.~\ref{fig:lane-rep-kps}). Most share the
same assumption that lanes extend vertically in the view. Models with
such an assumption usually perform well for such lanes but
fail to handle horizontal lanes. Particularly, the tilted anchor does not support
curved lanes, while a keypoint-based anchor cannot sort the keypoints properly
in case of strongly curving lanes in which $y$-coordinates are out of order.

To address these challenges, \self describes a lane as a whole with
a \emph{anchor-chain}, $\mathrm{Lane}_{\mathrm{ca}} = \{(x_1, y_1), \dots, (x_N , y_N)\}$,
where $N$ is the number of nodes in the anchor-chain, and $x_{i}$, $y_{i}$ are 
normalized relative coordinates with values in the range [0, 1] as shown in
Fig.~\ref{fig:anchor-chain}.

Regarding supervision, \self uses two types of ground truth: an ordered set of
manually annotated nodes, $\mathrm{Lane}_m$ (see Fig.~\ref{fig:ca-manual-anno}), and
a densely sampled set of nodes, $\mathrm{Lane}_s$
(Fig.~\ref{fig:ca-sample-anno}), obtained by uniform sampling along 
$\mathrm{Lane}_m$. To predict the lane, $\mathrm{Lane}_{\mathrm{pr}}$
(see Fig.~\ref{fig:ca-current-pred}, the initial anchor-chain), \self first matches
it to $\mathrm{Lane}_m$ using the Hungarian algorithm~\cite{hungarian}. Nodes
in $\mathrm{Lane}_{\mathrm{ca}}$ that have not been successfully matched are then matched
with $\mathrm{Lane}_s$ using the same algorithm, ensuring that each node in
$\mathrm{Lane}_{\mathrm{pr}}$ is matched to one corresponding ground truth node, forming the
final anchor-chain. Finally,
\self employs L1 distance to supervise the horizontal and vertical coordinates
of each predicted node on the anchor-chain; the loss is:
\begin{equation}
  \label{eq:reg_loss}
    L_{\mathrm{reg}} = -\frac{1}{N}\sum\limits_{(x,y)\in \mathrm{Lane}}|\hat{P}_{xy} - P_{xy}|
\end{equation}
where $\hat{P}_{xy}$ and $P_{xy}$ denote predicted and ground truth
nodes, respectively.

Unlike the uniform sampling in the Point Set approach~\cite{maptr}, \self samples
$M (M \gg N)$ nodes on the annotated lane and performs Hungarian matching between
these nodes and predictions. It matches each predicted node to the closest
ground truth node, giving the nodes more degrees of freedom. This allows
the anchor-chain to learn implicit human preferences in the annotations during
training and to distribute the nodes at higher information density at turning points.
Besides, thanks to the low prior assumption setting, the anchor-chain can describe
lanes of any shape and requires no longer prior conditions.

In addition, the anchor-chain can also provide fine-grained position information for
the network. 
The cross-attention module needs to gather features from the entire
feature map, so it is necessary to provide appropriate position priors for
each query to focus attention on the locality surrounding the targets.
\self explicitly models the query position as an anchor, which is a similar approach
to DAB-DETR. The anchor-chain can effectively help the network
aggregate features from nearby regions of different parts of a target using
multi-referenced deformable cross-attention modules.

\begin{figure}[!t]
  \centering
  \includegraphics[width=\columnwidth]{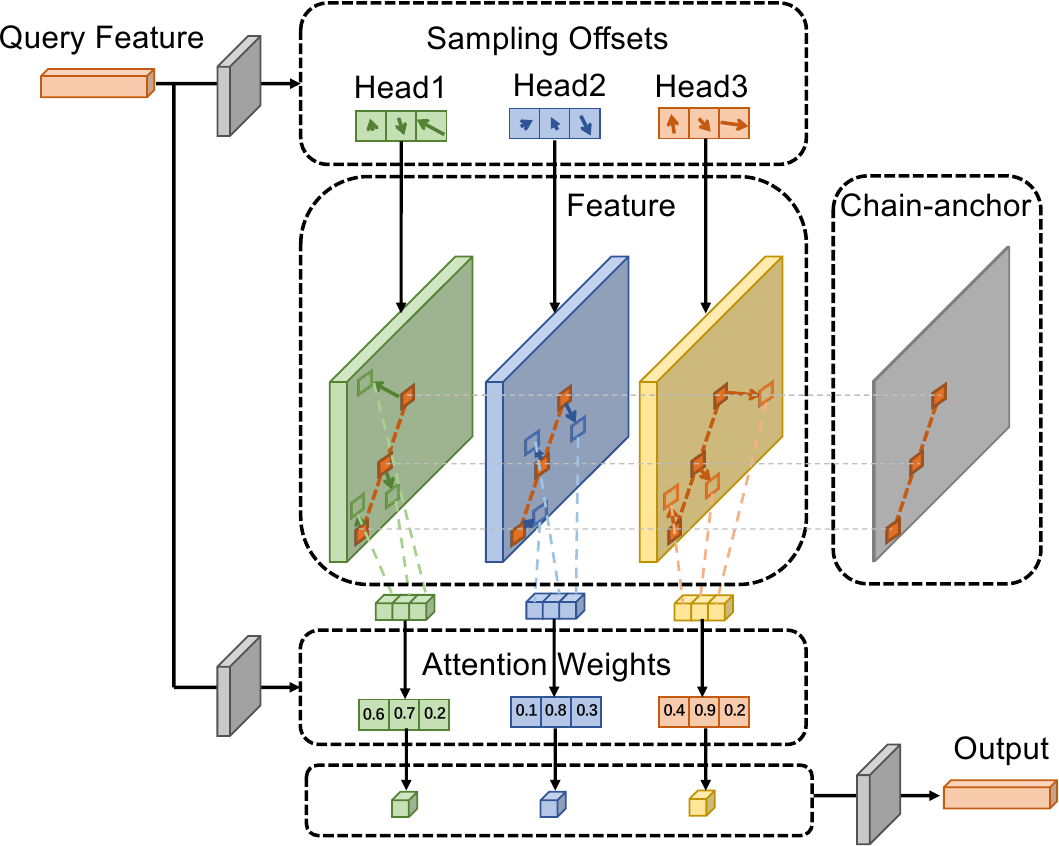}
  \caption{The multi-referenced deformable cross-attention module leverages positional information from the anchor-chain to guide the attention distribution.}
  \label{fig:mrda}
\end{figure}

\subsection{Multi-Referenced Deformable Cross-Attention}
\label{sec:multi-ref-deform-cross-attention}

As most adjacent features contain similar appearance information, traditional
cross-attention modules bring a lot of additional computation, most of which is
useless computation on backgrounds. The improved deformable attention
module~\cite{deformabledetr} samples partial information based on a learnable
offset field. However, due to the lack of explicit supervision to guide 
sampling along the object contour, the deformable attention module can still
lead to wasted computation when sampling around the center of elongated lanes,
and cannot balance sampling of endpoints far away from the center point.
Therefore, \self uses points (anchors) distributed along the lanes as reference
points and samples only part of the information around each point, as shown in
Fig.~\ref{fig:mrda}. By assigning only a small number of keys along the lanes
for each query, the convergence speed and computational efficiency can be
improved significantly.

\subsection{Line IoU}
\label{sec:line_iou}
\subsubsection{Background}
L1 distance loss (see Equation~\eqref{eq:reg_loss}) can independently optimize the
position of each node on the chain, but lacking an overall error calculation of
the target nodes leads to slow convergence. IoU loss is a widely adopted overall
loss function in object detection/segmentation, computed by bounding box overlap
or pixel-level intersection. For thin and long objects such as lanes, the bounding
box method suffers from large errors while the latter does not support gradient
backpropagation using the current lane representation.

Hence, CLRNet~\cite{clrnet} suggested \emph{line IoU}, an approximate IoU algorithm for
lanes. However, it assumes lanes are all vertical and only calculates the
horizontal errors between two lines. Thus line IoU results for a pair of lines
vary a lot when they appear at different angles. In addition, the existing line
IoU algorithm is bound to a row-wise classification head and can only describe
lanes that extend vertically in the view, which cannot be used to optimize
our anchor-chain that can describe lane lines of any shape. To address such
limitations, we propose two line IoU algorithms inspired by the Anchor-chain:
\emph{point-to-point} and \emph{dense-sampling} Line IoU.

\begin{figure}[t]
  \centering
  \includegraphics[width=0.5\columnwidth]{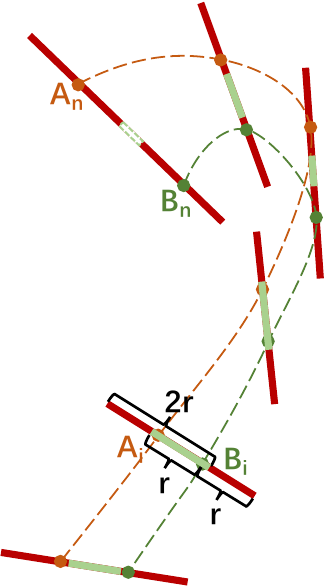}
  \caption{Point-to-point (P2P) line IoU.}
  \label{fig:p2p-iou}
\end{figure}

\begin{figure*}[!t]
  \centering
  \subfigure[Vertical splitting.]{\label{fig:dense-iou-split-v}%
    \includegraphics[width=.45\linewidth]{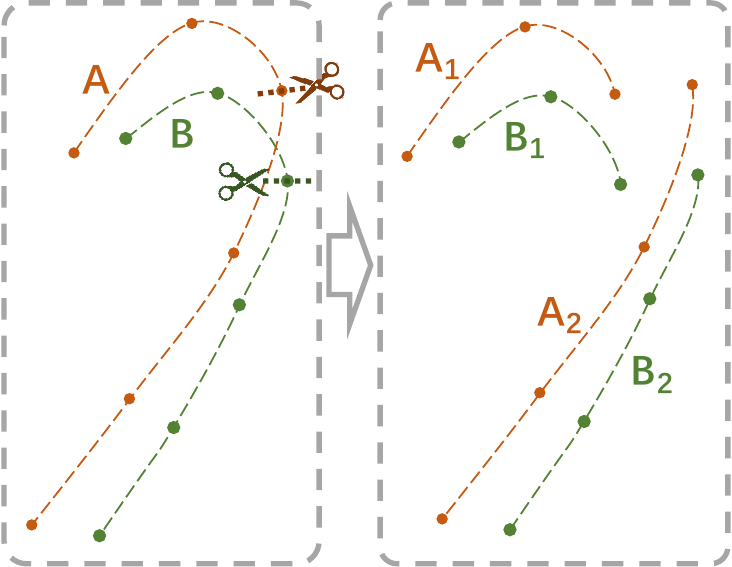}}\quad
  \subfigure[Horizontal splitting.]{\label{fig:dense-iou-split-h}%
    \includegraphics[width=.45\linewidth]{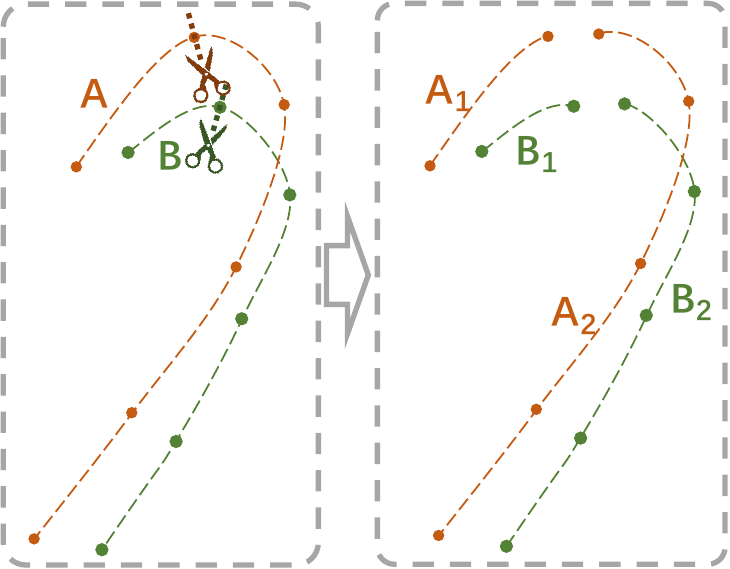}}
  \caption{Split U-turn lines.}
  \label{fig:dense-iou-split}
\end{figure*}

\subsubsection{Point-to-Point Line IoU}
\label{sec:p2p-line-iou}

To calculate the IoU of two lines $A$ and $B$, the Point-to-Point (P2P)
algorithm picks the same number ($N$) of keypoints on them uniformly and pairs
the $i$-th points together $(A_{i}, B_{i})$. Given a fixed length $r$, draw two
line segments along the direction of $(A_{i}, B_{i})$ with length $2r$, taking
$A_{i}$ and $B_{i}$ as midpoints, respectively (thick solid red lines in
Fig.~\ref{fig:p2p-iou}). Let $L_{{\cup}}^{A_{i}B_{i}}$ be the distance between the far
endpoints (the union) of the two segments, $L_{\cap}^{A_{i}B_{i}}$ the distance
between the near endpoints (the intersection, green lines in Fig.~\ref{fig:p2p-iou})
and $||A_{i}B_{i}||_{2}$ the length of $(A_{i}, B_{i})$. If the two segments
overlap, $L_{\cap}^{A_{i}B_{i}} = 2r-||A_{i}B_{i}||_{2}$ (solid green
lines), while they are separated from each other (like $(A_{n}, B_{n})$ in
Fig.~\ref{fig:p2p-iou}). This expression is still used instead of 0, describing
how far away they are in negative values (thick white line with the dashed green
border between $A_{n}$ and $B_{n}$). This is helpful for gradients and
optimization. Then, LIoU$_{\mathrm{P2P}}$ is defined as
Equation~(\eqref{eq:p2p-line-iou}).
\begin{equation}
  \label{eq:p2p-line-iou}
  \mathrm{LIoU}_{\mathrm{P2P}}(A, B) = \frac{\sum_{i=1}^{N} L^{A_iB_i}_{\cap}}{\sum_{i=1}^{N} L^{A_iB_i}_{{\cup}}} = \frac{\sum_{i=1}^{N} (2r-||A_iB_i||_2)}{\sum_{i=1}^{N} (2r+||A_iB_i||_2)}
\end{equation}

Fig.~\ref{fig:p2p-iou} illustrates how LIoU$_{\mathrm{P2P}}$ is computed.
LIoU$_{\mathrm{P2P}}$ ranges from -1 to 1. When the two lines overlap
completely, it is 1, while when they are infinitely apart, it converges to -1.

P2P line IoU describes the trend similarity between lanes and is not affected by
the lane orientation. Thanks to its stability in the optimization process, \self
leverages P2P line IoU to compute the matching error as a supplement to the L1
distance and classification costs.

\subsubsection{Dense-sampling Line IoU}
\label{sec:dense-sampling-line}

DS line IoU consists of two steps:

\emph{Step 1: Line splitting.} DS line IoU samples keypoints in both $x$ and
$y$ directions. To be compatible with curved lanes and backtracking lanes, it
first splits the lines in the sampling direction into multiple one-way line segments. Considering Fig.~\ref{fig:dense-iou-split}, after the lines
$A$ and $B$ are split, they respectively consist of line segments $\{A_1, A_2\}$ and
$\{B_1, B_2\}$, each of which contains multiple sample points.

\begin{figure*}[!t]
  \centering
  \subfigure[Vertical Sampling]{\label{fig:dense-iou-v}%
    \includegraphics[width=.3\linewidth]{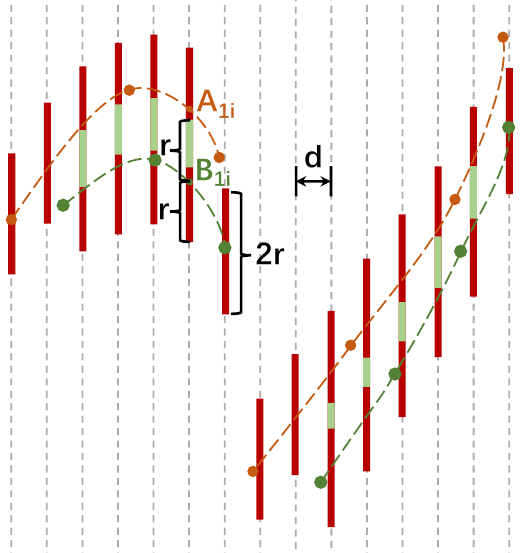}}\quad
  \subfigure[Horizontal Sampling]{\label{fig:dense-iou-h}%
    \includegraphics[width=.3\linewidth]{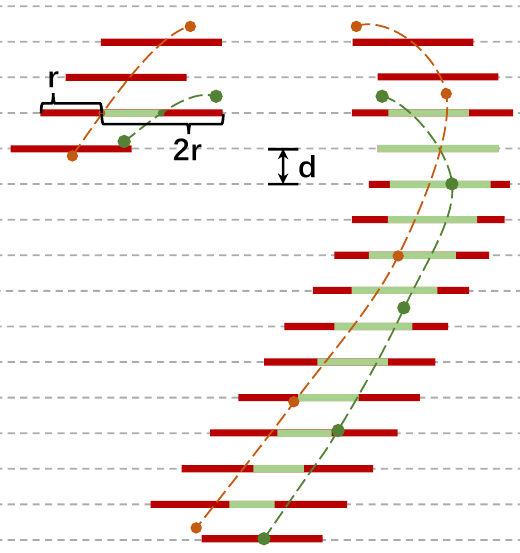}}
  \caption{Dense-Sampling (DS) Line IoU.}
  \label{fig:dense-iou}
\end{figure*}

\emph{Step 2: Segment-wise calculation.} For each unidirectional sub-line, the
algorithm sets reference lines for every distance $d$ in the sampling direction.
If a reference line intersects both line segments, the two intersection points are
paired together as in P2P, in which case DS shares the definitions of $L_{\cap}$
and $L_{{\cup}}$ with P2P in Equation~\eqref{eq:dense-iou-pp}.
\begin{equation}
  \label{eq:dense-iou-pp}
  \begin{split}
    L^{A_{ij}B_{ij}}_{\cap} &= 2r-||A_{ij}B_{ij}||_2\\
    L^{A_{ij}B_{ij}}_{\cup_2} &= 2r+||A_{ij}B_{ij}||_2
  \end{split}
\end{equation}
where $i$ is the segment index, $j$ the reference point index, 2 in $\cup_2$ refers to the number of intersection points.

Otherwise, if a reference line intersects only one sub-line, $L_{\cap}$ is set
to 0, and $L_{\cup_1}$ is set as below.
\begin{equation}
  \label{eq:dense-iou-p}
    L^{A}_{\cup_1} = L^{B}_{\cup_1} = 2r
\end{equation}
Fig.~\ref{fig:dense-iou} illustrates how the reference lines intersect the
segments and cases of different intersection points.

Then, DS line IoU is defined as the ratio of accumulated intersection distances
to the union ones as in P2P:
\begin{equation}
%  \label{eq:dense-line-iou}
 \mathrm{LIoU}_{\mathrm{DS}}(A, B) = \frac{\sum\limits_{i=1}^{O} \sum\limits_{j=1}^{N_i} L^{A_{ij}B_{ij}}_{\cap}}{ \sum\limits_{i=1}^{O}\sum\limits_{j=1}^{N_i} L^{A_{ij}B_{ij}}_{\mathrm{\cup_2}} + \sum\limits_{i=1}^{U_A} L^{A}_{\mathrm{\cup_1}} + \sum\limits_{i=1}^{U_B} L^{B}_{\mathrm{\cup_1}} }
\end{equation}
where $O$ is the number of $\cup_2$ segments, $N_i$ the number of
sampling points in the $i$-th $\cup_2$ segment, and $U_A$ and $U_{B}$ are the
numbers of $\cup_1$ segments on lines $A$ and $B$, respectively.

Changing the sampling interval $d$ can make a flexible balance between precision and
speed. \self sets $d$ to 8 pixels by default. While DS requires more
computation than P2P, it is straightforward to parallelize the algorithm on the GPU.

\self applies DS line IoU to the overall loss of the target as shown in
Equation~\eqref{eq:iou-loss}, which can record subtle differences between the
predictions and GT. Since the entire chain of nodes is considered as a whole, the
predictions of the horizontal and vertical coordinates are optimized in $x$ and
$y$ directions independently.
\begin{equation}
  \label{eq:iou-loss}
    L_{\mathrm{iou}} = 1 - \mathrm{LIoU}_{\mathrm{DS}}(P_{\mathrm{chain}}, \hat{P}_{\mathrm{chain}})
\end{equation}
where $P_{\mathrm{chain}}$ and $\hat{P}_{\mathrm{chain}}$ are the ground truth and predicted
chains, respectively.

\subsection{Gaussian Heatmap Auxiliary Branch}
\label{sec:auxiliary_branch}

Multi-task training has been widely used to enhance the generalization ability
of single-task models. Therefore, \self introduces a Gaussian heatmap
branch~\cite{canet} as an auxiliary training branch. The structure and workflow
of the branch are similar to that of Mask-DINO~\cite{mask_dino}, as indicated by the
green dotted line in Fig.~\ref{fig:ldtr-pipeline}.

The Gaussian heatmap auxiliary training branch only performs forward and
backward propagation during the training process, and the gradient is
backpropagated to update the network weights. During inferencing, this
branch is discarded for efficiency. Networks with auxiliary branches have more parameters, making it easier to learn how to fit the relationship
between input images and ground truth from the initial state, effectively
improving the rate of convergence and stability of the optimization direction of
the model. \self uses the same $L_{\mathrm{mask}}$ and $L_{\mathrm{offset}}$ as CANet
to supervise training of the auxiliary branch.

\subsection{Total Loss}
\label{sec:total-loss}
In addition, \self adopts Focal loss to supervise the classification head following
DETR. It is used to determine whether each query corresponds to a target in
the input. The classification loss is:
 \begin{equation}
\label{eq:cls-loss}
L_{\mathrm{cls}} = \frac{-1}{N_Q}\sum\limits_{q \in Q}
\begin{cases}
  (1-\hat{P}_{q})^\gamma\log{\hat{P}_{q}} & P_{q} = 1 \\
  (1-P_{q})^\lambda \hat{P}_{q}^\gamma\log(1-\hat{P}_{q}) & \mathrm{otherwise}
\end{cases}
\end{equation}
where $Q$ represents the set of all queries, $P_{q}$ and $\hat{P}_{q}$ denote
the prediction and binary match ground truth for Query $q$, respectively, and
$N_Q$ is the number of queries. The total loss function in \self is then:
\begin{equation}
\label{eq:total_loss}
L_{\mathrm{total}} = a L_{\mathrm{cls}} + b L_{\mathrm{reg}} + c L_{\mathrm{iou}} + L_{\mathrm{mask}} + L_{\mathrm{offset}}
\end{equation}
The hyperparameters $a$, $b$, and $c$ are set to 1, 5, and 1, respectively.

\section{Experiments}
\label{sec:experiments}

\subsection{Datasets}
\label{sec:datasets}
To evaluate \self, extensive experiments were conducted on two widely used
datasets for lane detection, CULane~\cite{scnn} and
CurveLanes~\cite{curvelanes}. As a comprehensive dataset, CULane contains images
from urban street views, rural roads, and highways under diverse conditions, e.g.\ with
glare and occlusion. Many lanes have little- or no-visual-clue, and lane detection requires a
deep understanding of the overall scene by the models as demonstrated in
Fig.~\ref{fig:no-visual-clue}. While lanes in CurveLanes are more obvious than
those in CULane, CurveLanes has more topologically complicated lanes, such as
forks, convergences, sharp turns, and T-junctions as seen in Fig.~\ref{fig:need-chain},
which were not well addressed in earlier datasets.

\begin{figure*}[t!]
\centering
  \subfigure[Predictions are determined as false positives for lanes lacking visual
  clues.]{\label{fig:fp-no-visual-clue}%
    \begin{minipage}[b]{\linewidth}
     \centering
      \includegraphics[width=0.4\linewidth]{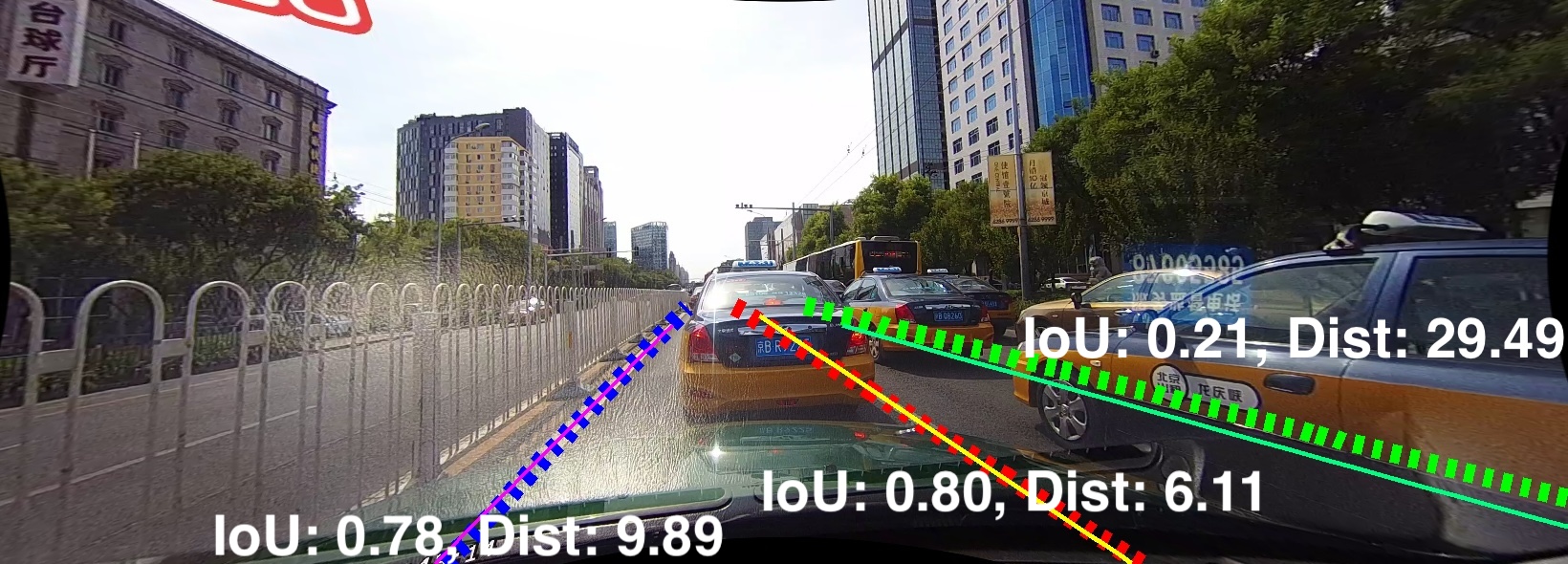}\quad
      \includegraphics[width=0.4\linewidth]{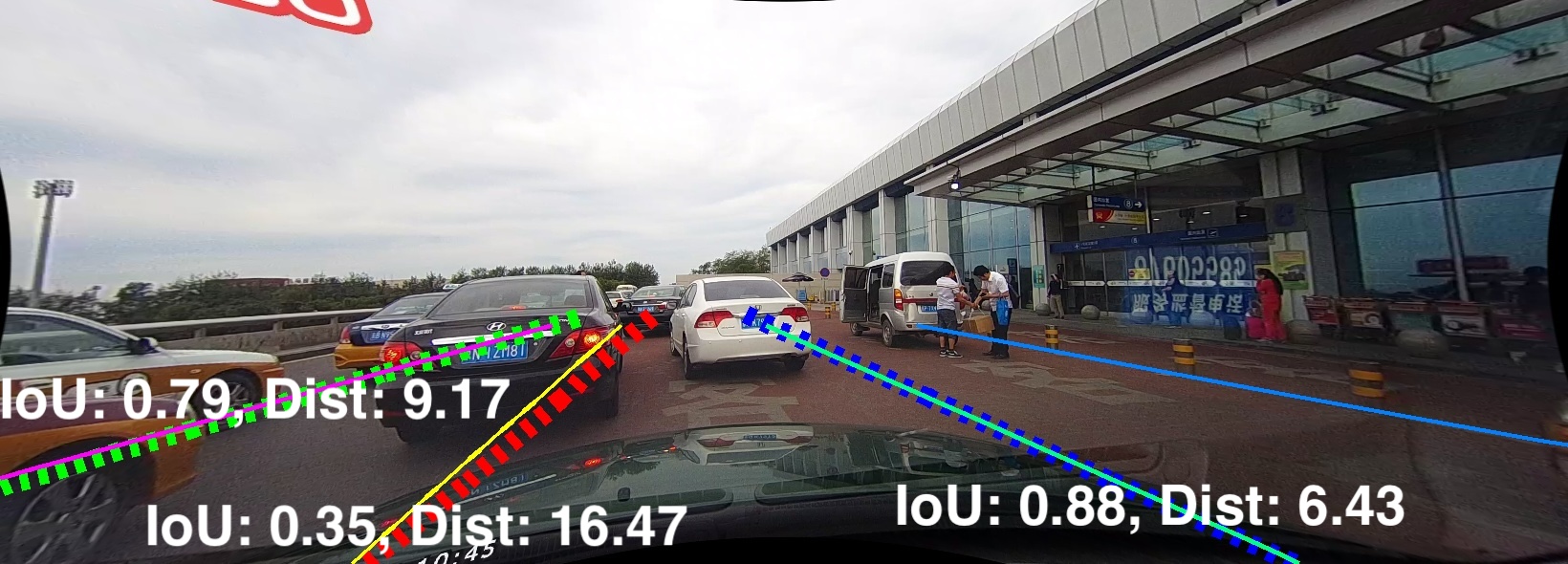}\\
      \includegraphics[width=0.4\linewidth]{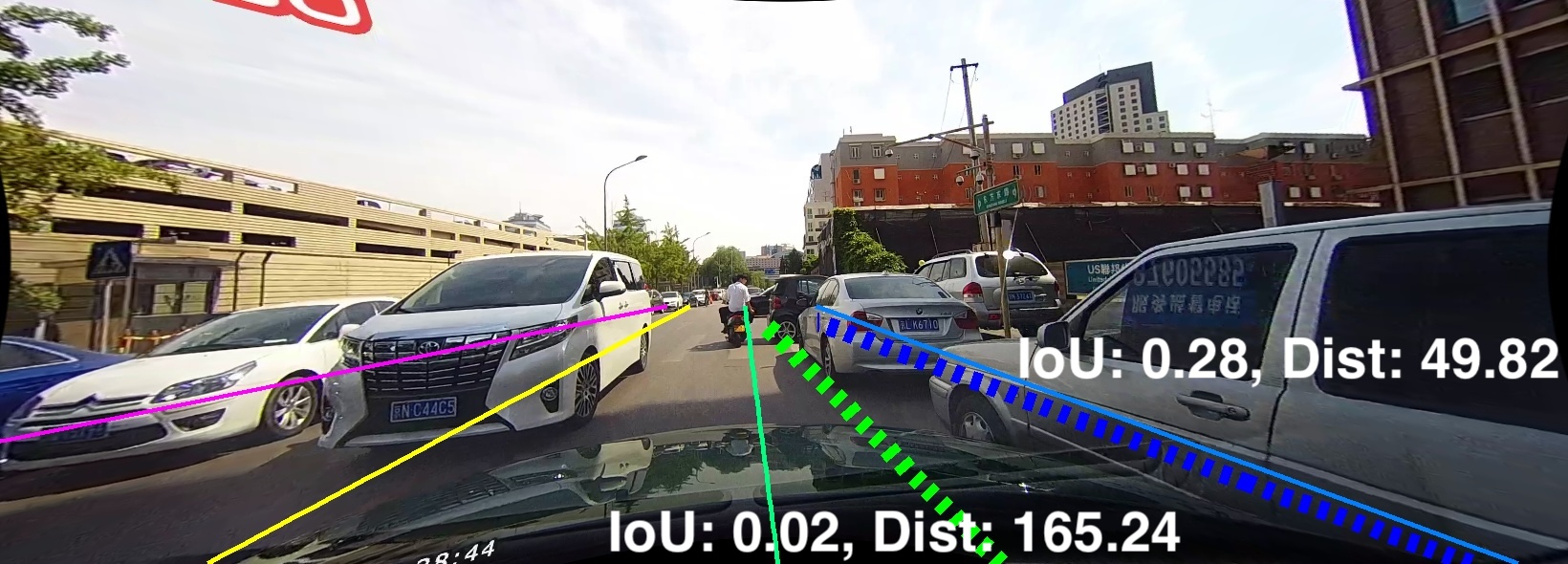}\quad
      \includegraphics[width=0.4\linewidth]{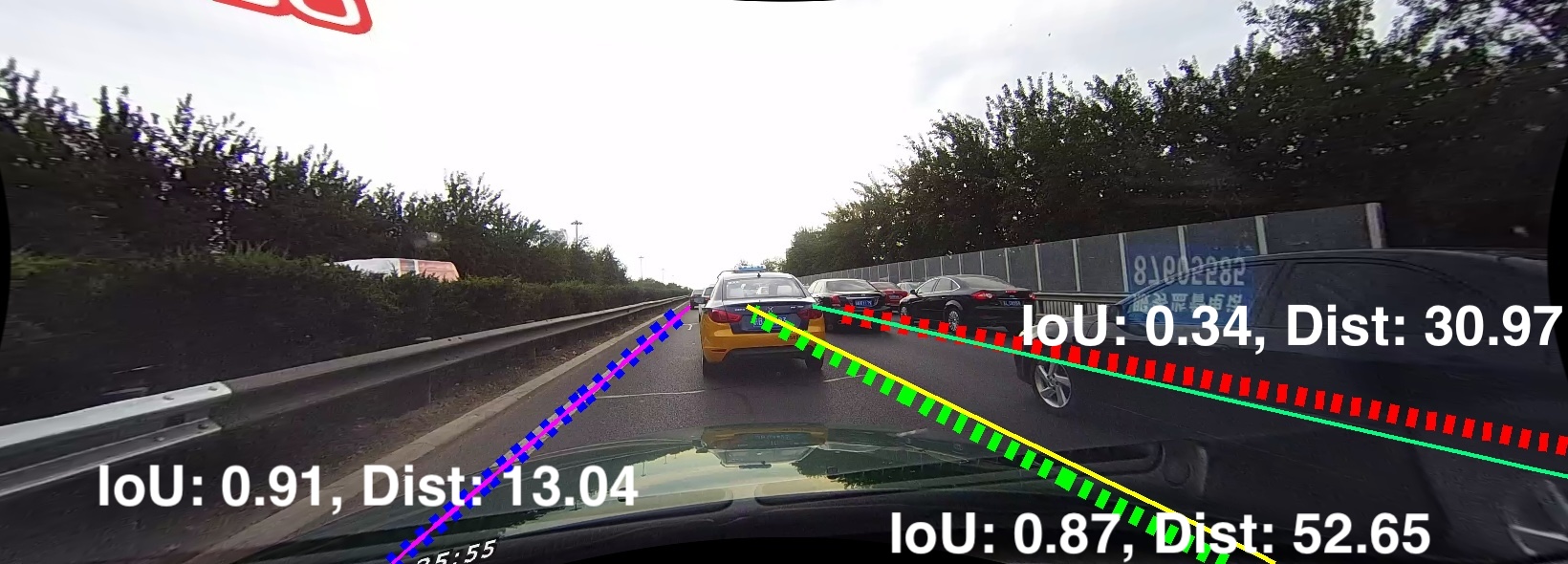}
    \end{minipage}}\\
  \subfigure[Incorrect predictions with high IoU.]{\label{fig:bad-iou}%
    \begin{minipage}[b]{\linewidth}
     \centering
      \includegraphics[width=0.4\linewidth]{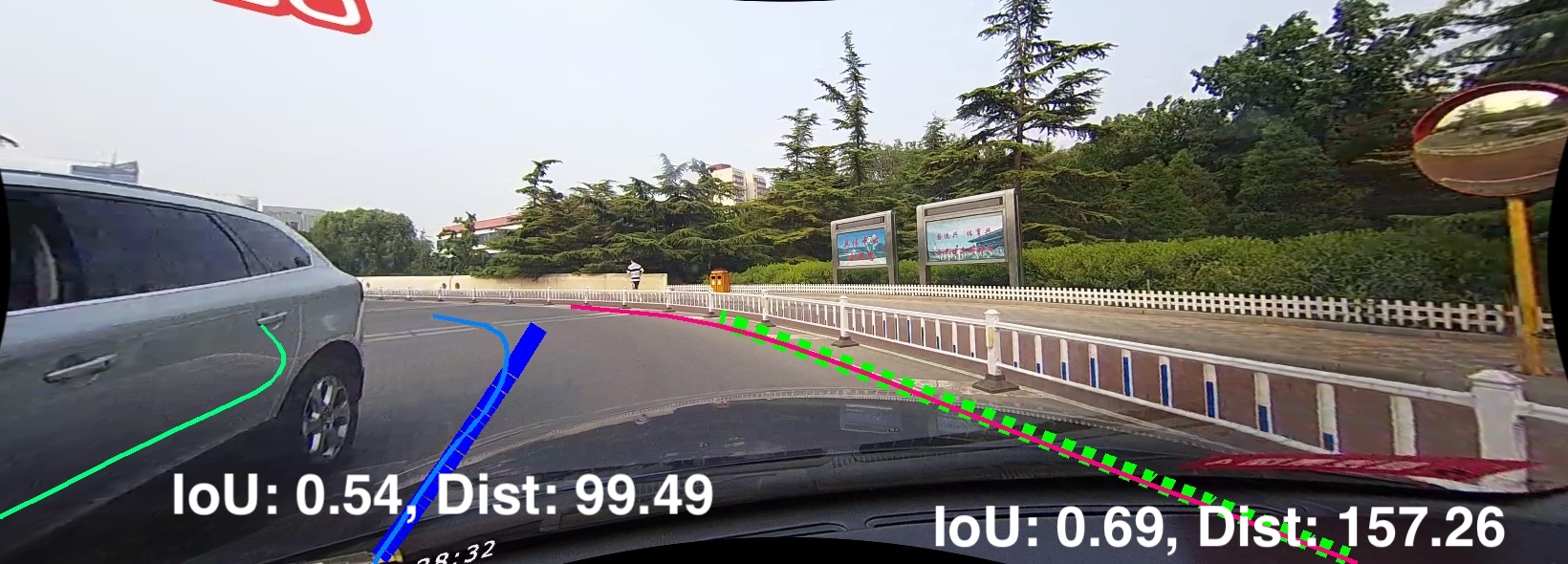}\quad
      \includegraphics[width=0.4\linewidth]{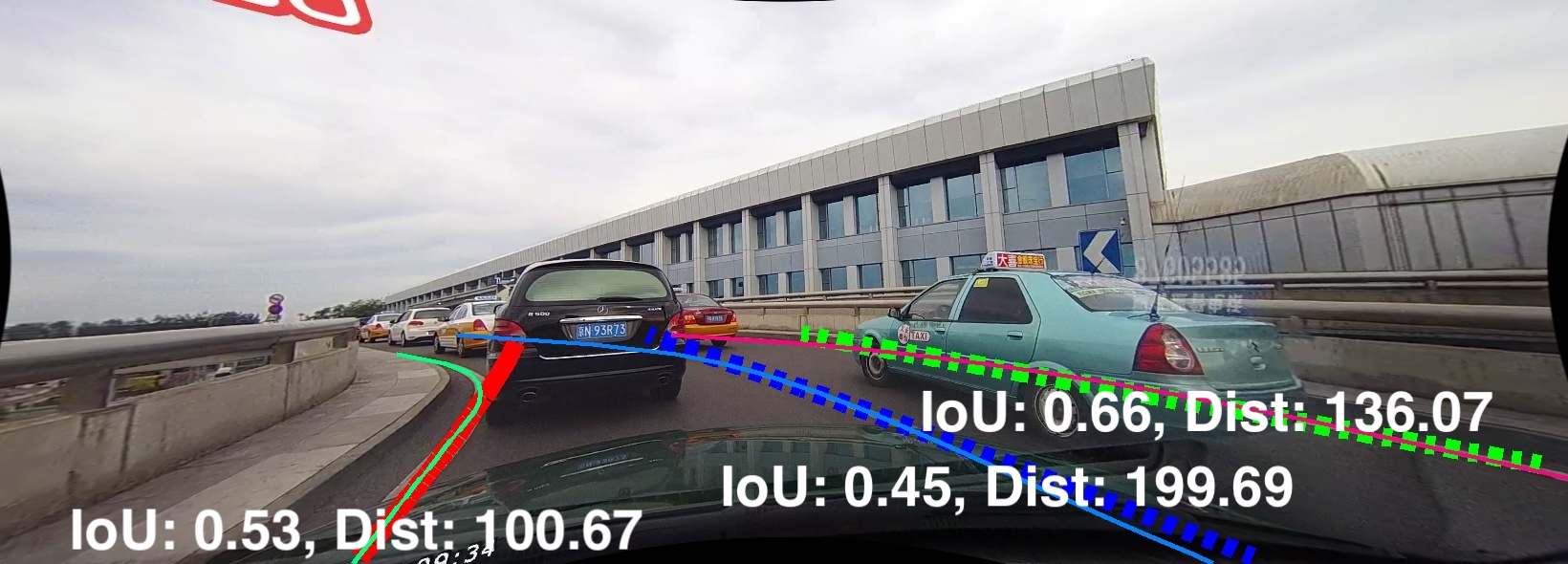}\\
      \includegraphics[width=0.4\linewidth]{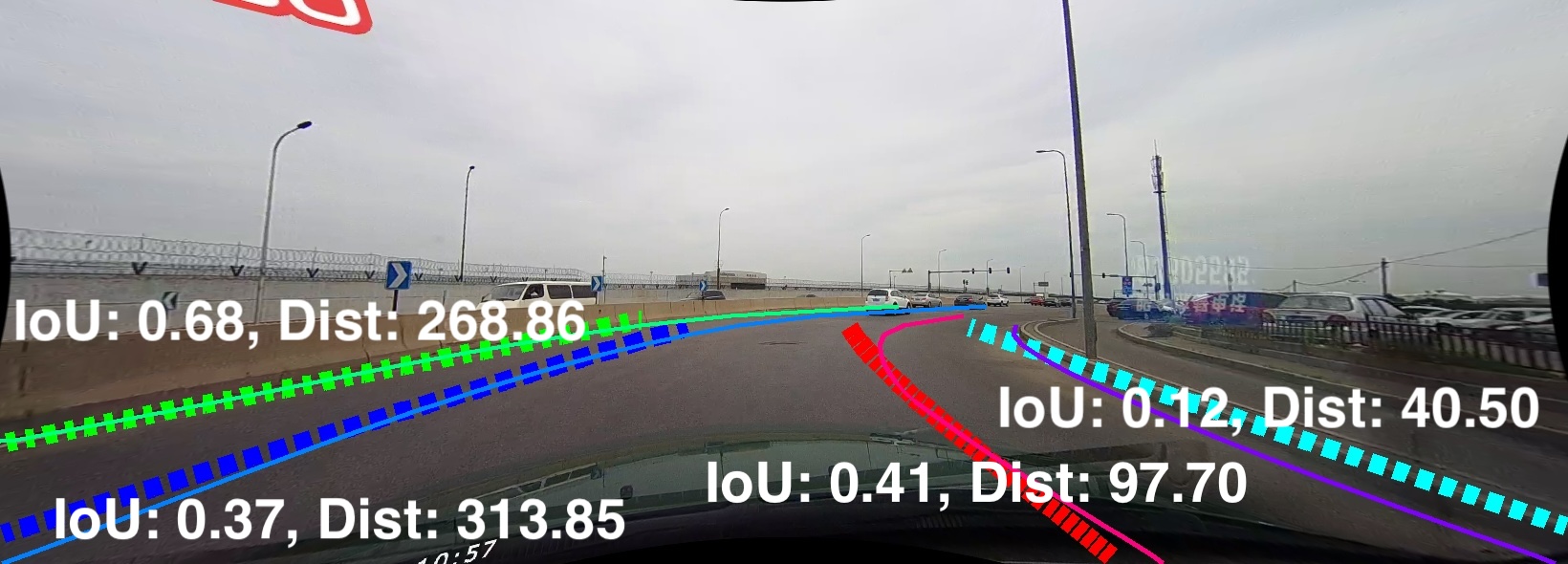}\quad
      \includegraphics[width=0.4\linewidth]{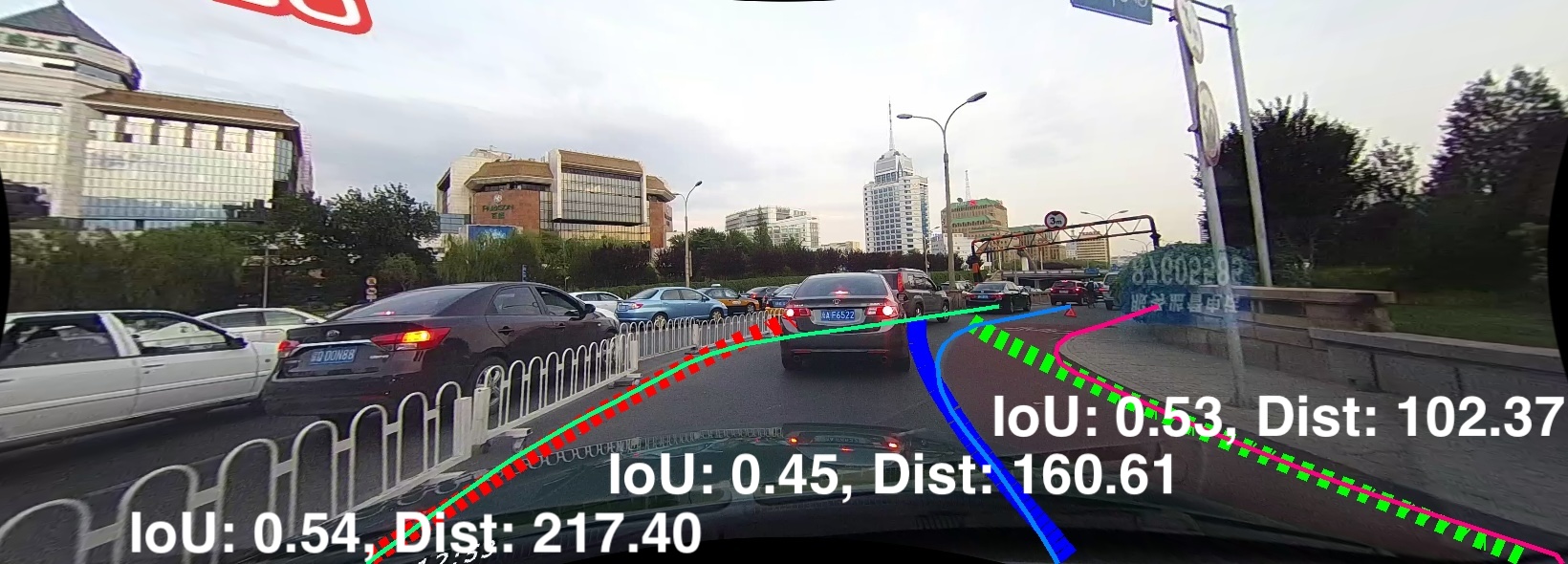}
    \end{minipage}}
  \caption{Abnormal predictions with typical IoU threshold. Thin solid lines are
    ground truth, and thick dotted ones are predicted lanes. Dist indicates the
    Fréchet distance between the prediction and corresponding ground truth.}
  \label{fig:abnormal-cases}
\end{figure*}

\subsection{Performance Metrics}
\label{sec:perf-indic}

\subsubsection{Parameterized F1-score}
\label{sec:param-f1-score}

F1-score is the default evaluation metric for lane detection models.
Predictions and ground truth are expanded into fixed-width masks and the IoU
between masks greater than the threshold is regarded as TP. The default IoU
threshold of 0.5 originated from general object detection but is too high for
lane detection. Because lanes are long, thin objects, slight jitter in predicted
points may lead to huge IoU variation. IoU is so sensitive that many
predictions may be rejected even though they are suitable for downstream tasks.
The situation would get worse for cases in Fig.~\ref{fig:abnormal-cases}.

On the one hand, many lanes lack obvious visual features to provide
sufficient information for models to accurately locate. Some predicted lanes
 run almost parallel to the ground truth at a tiny distance, like
the predictions with IoU of 0.34, 0.35, 0.28, and 0.21 in
Fig.~\ref{fig:fp-no-visual-clue}. These predictions should be allowed but are
dropped because of the small IoU. Hence, it is reasonable to decrease the
threshold of IoU to allow predictions with inaccurate but acceptable locations.
Looking at Fig.~\ref{fig:fp-no-visual-clue}, the threshold could be adjusted
from 0.5 to 0.2.

On the other hand, certain high IoU predictions should not be accepted because
IoU only focuses on pixel overlap but lacks measurement of lane trends, as
Fig.~\ref{fig:bad-iou} illustrates. Specifically, two situations cause such
misjudgments: incomplete prediction and incorrect trend prediction, both of
which are harmful to real-world downstream tasks. Simply decreasing the IoU
threshold may exacerbate the occurrence of such misjudgments, so Fréchet
distance is introduced to filter out predictions that deviate significantly from
the real trend. The original Fréchet distance expects the lines to be roughly
the same length and calculates the maximum shortest distance bidirectionally.
However, concerning lane detection, the predicted lanes may be longer than the
ground truth. Though the additional predicted segment implies the lane trend, it
is difficult to justify and should be weighted less.

Thus, we modify Fréchet distance to calculate unidirectionally from ground truth
to predictions only, to make sure every point in ground truth counts but not
vice versa. Predicted lanes are then considered to match only if they
satisfy a hybrid constraint: IoU$\geq \alpha$ and Fréchet distance$\leq \beta$.
Thus, the F1-score will depend on two parameters, IoU threshold ($\alpha$) and
Fréchet distance threshold ($\beta$): F1= F1($\alpha$, $\beta$). The
classic F1-score configuration corresponds to F1(0.5, $+\infty$), while in our
experiments we set $\beta$ to be 4\% of the image width, i.e., F1(0.2, 60) for
CULane and F1(0.2, 10) for CurveLanes. For simplicity, F1 without qualification means F1(0.5,
$+\infty$) in the following unless otherwise specified. Not only lane detection,
but also other object detection tasks could benefit from the idea of parameterized
F1-score.

The video in the supplementary material compares the prediction
results of the current state-of-the-art model CLRNet and \self on CULane using two different
evaluation metrics. The video visually demonstrates that \self has better
instance recall in situations where visual cues are lacking. However, these
additional recalls compared to CLRNet are often regarded as false positives (red
lines).

\subsubsection{MIoU and MDis}
\label{sec:miou-mdis}
In Fig.~\ref{fig:abnormal-cases}, all predictions of IoU below 0.5 are
dropped by F1 because they are regarded as false positives, though they are suitable for
downstream tasks and should be accepted instead. Consequently, if a model
predicts more lanes of this kind, its precision will become worse.
This contradiction stems from the definition of precision, and more insightful
indicators are required to investigate lane detection models together.

Thus, we suggest turning to \emph{MIoU} (Mean IoU) and
\emph{MDis} (Mean Fréchet Distance), defined as 
\begin{equation}
  \label{eq:miou-mdis}
  \mathrm{MIoU} = \frac{\sum_{i\in U_{\mathrm{TP}}}{\mathrm{IoU}_i}}{N_{\mathrm{TP}}}, \quad \mathrm{MDis} = \frac{\sum_{i\in U_{\mathrm{TP}}}\mathrm{Dis}_i}{N_{\mathrm{TP}}},
\end{equation}
to
compare position precision and trend similarity of predicted lanes under similar
recall rates. Both metrics describe how closely the predictions relate to ground
truth overall, which makes more sense than precision with respect to lane
detection. Here, $N_{\mathrm{TP}}$ is the number of true positive
predictions, and $U_{\mathrm{TP}}$ is the set of all true positive instances.

\begin{table}[t]
  \caption{Performance of state-of-the-art models and \self on CULane. Prec = Precision; PF1 = F1(0.2, 60).}
  \label{tab:diff-eval-thr-comparation}
  \centering
    \begin{tabular}{lrrrrrr}
    \hline % --------------------------------------------------------------------------------
    Model  & F1             & Prec      & Recall         & MIoU           & MDis           & PF1    \\
    \hline % --------------------------------------------------------------------------------
    CANet  & 79.86          & \textbf{88.03} & 73.08          & \textbf{82.61} & 15.83          & 81.85          \\
    CLRNet & \textbf{80.47} & 87.13          & 74.77          & 82.13          & 22.62          & 82.11          \\
    \self  & 78.16          & 81.53          & \textbf{75.06} & 82.23          & \textbf{12.89} & \textbf{84.18} \\
    \hline % --------------------------------------------------------------------------------
\end{tabular}
\end{table}
\begin{table}[t]
  \setlength{\tabcolsep}{5pt}
  \caption{\self performance by adopting the two IoU algorithms in different
    cost and loss combinations. Prec = Precision; PF1 = F1(0.2, 10).}
  \label{tab:iou-ablation}
  \centering
  \begin{tabular}{crrrrrrr}
    \hline % --------------------------------------------------------------------------------
    Cost & Loss & F1             & Prec           & Recall         & MIoU           & MDis          & PF1            \\
    \hline % --------------------------------------------------------------------------------
    N/A  & N/A  & 86.24          & 88.20          & 84.36          & 79.86          & 2.88          & 88.60          \\
    P2P  & P2P  & 87.15          & 90.92          & 83.69          & 81.25          & 2.77          & 88.51          \\
    DS   & DS   & 87.74          & 90.28          & \textbf{85.33} & 81.19          & 2.86          & 89.00          \\
    P2P  & DS   & \textbf{87.96} & \textbf{91.19} & 84.95          & \textbf{81.31} & \textbf{2.85} & \textbf{89.01} \\
\hline % --------------------------------------------------------------------------------
\end{tabular}
\end{table}

\begin{table*}[t!]
  \caption{Ablation experiments of different components in \self on CurveLanes.}
  \label{tab:ldtr-ablation}
  \centering
  \begin{tabular}{llllllll}
    \hline % --------------------------------------------------------------------------------
    No. & Model         & F1                       & Precision                & Recall                   & MIoU                     & MDis                    & F1(0.2, 10)              \\
    \hline % --------------------------------------------------------------------------------
    1 & Anchor-chain  & 85.57                    & 87.41                    & 83.80                    & 78.86                    & 3.01                    & 88.44                    \\
    2 & +MRDA         & 86.24$_{+0.67}$          & 88.20$_{+0.79}$          & 84.36$_{+0.56}$          & 79.86$_{+1.00}$          & 2.88$_{-0.13}$          & 88.60$_{+0.16}$          \\
    3 & +Line-IoU     & 87.96$_{+1.72}$          & 91.19$_{+2.99}$          & 84.95$_{+0.59}$          & 81.31$_{+1.45}$          & 2.85$_{-0.03}$          & 89.01$_{+0.41}$          \\
    4 & +Auxiliary    & \textbf{88.44}$_{+0.48}$ & \textbf{91.55$_{+0.36}$} & \textbf{85.53}$_{+0.58}$ & \textbf{81.39}$_{+0.08}$ & \textbf{2.69}$_{-0.16}$ & \textbf{89.66}$_{+0.65}$ \\
\hline % --------------------------------------------------------------------------------
\end{tabular}
\end{table*}
\begin{figure*}[t!]
  \subfigure[Original deformable attention (ODA).]{\label{fig:oriattn}%
    \begin{minipage}[b]{\linewidth}
      \includegraphics[width=0.495\linewidth]{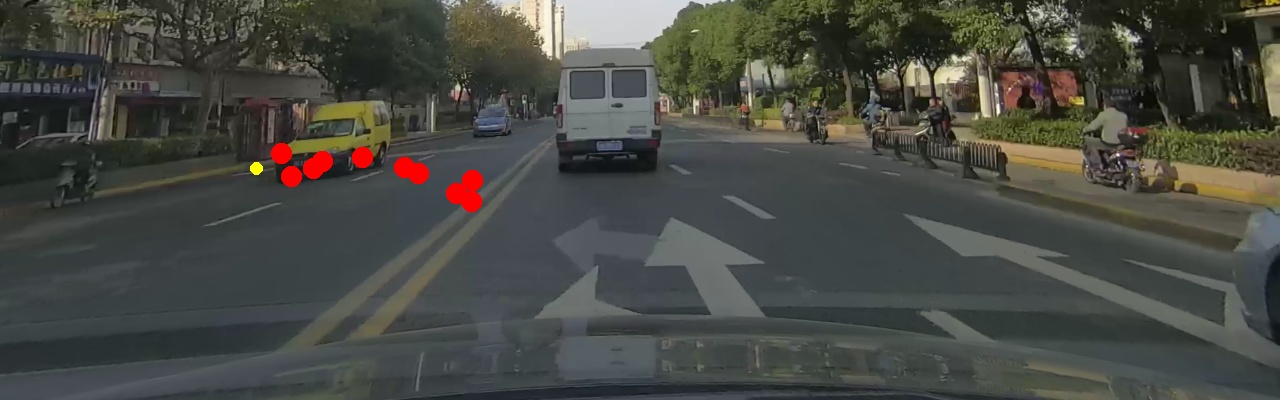}
      \includegraphics[width=0.495\linewidth]{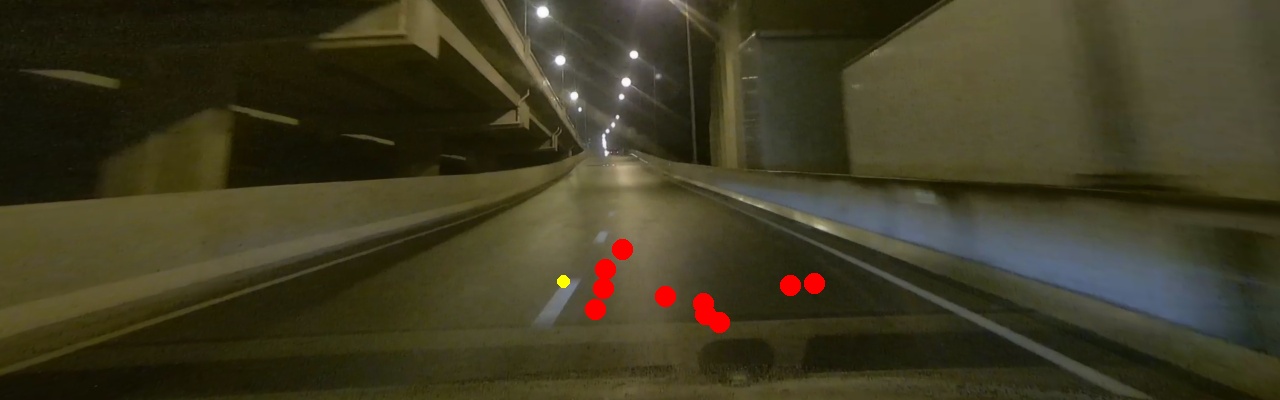}\\[1pt]
      \includegraphics[width=0.495\linewidth]{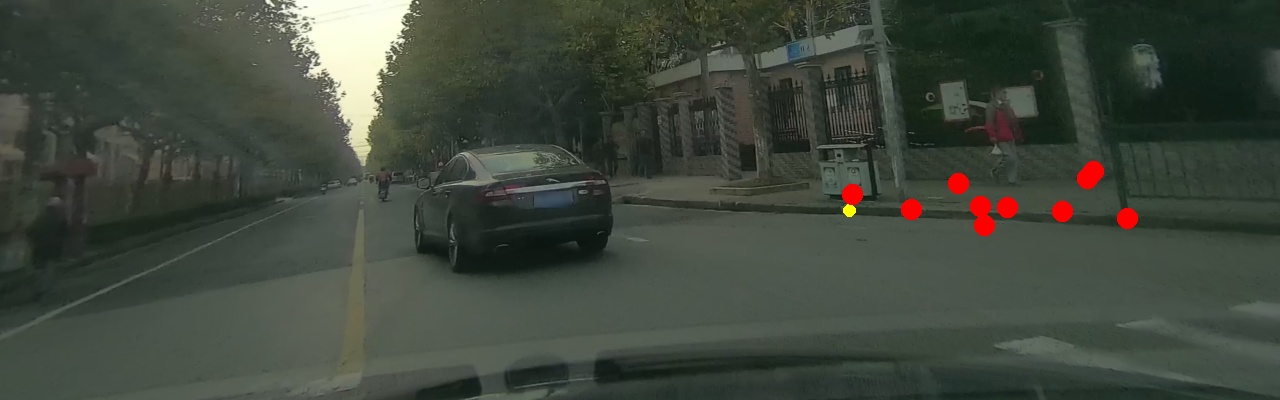}
      \includegraphics[width=0.495\linewidth]{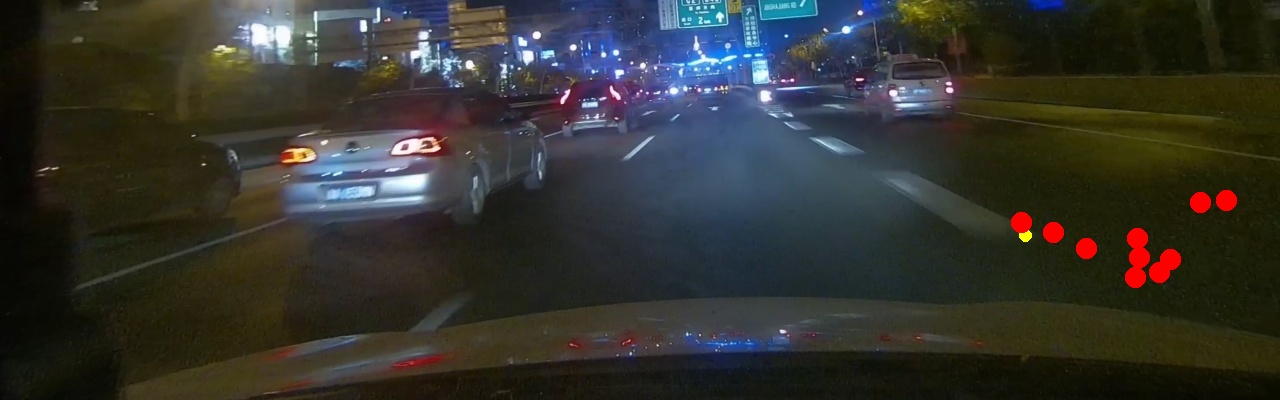}
    \end{minipage}}\\
  \subfigure[Multi-referenced deformable attention (MRDA).]{\label{fig:mrda-samples}%
    \begin{minipage}[b]{\linewidth}
      \includegraphics[width=0.495\linewidth]{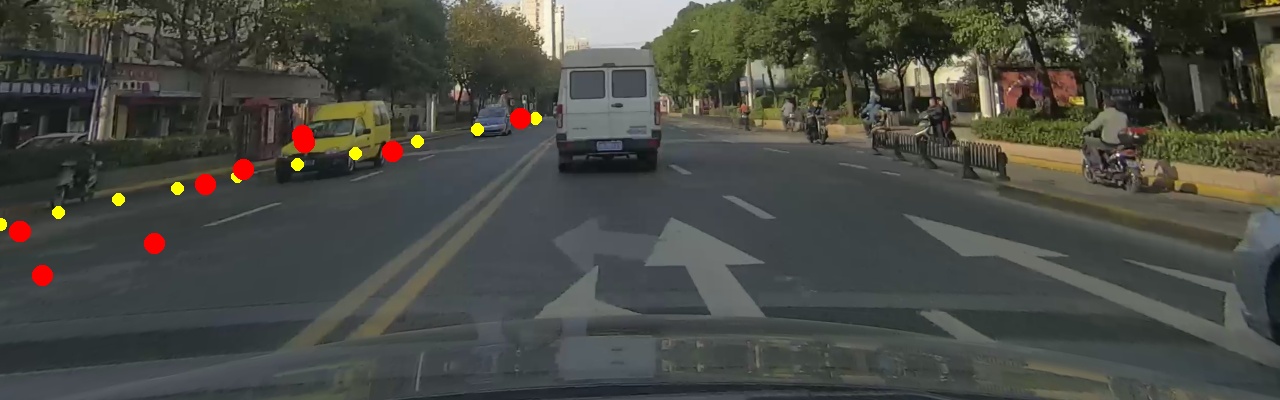}
      \includegraphics[width=0.495\linewidth]{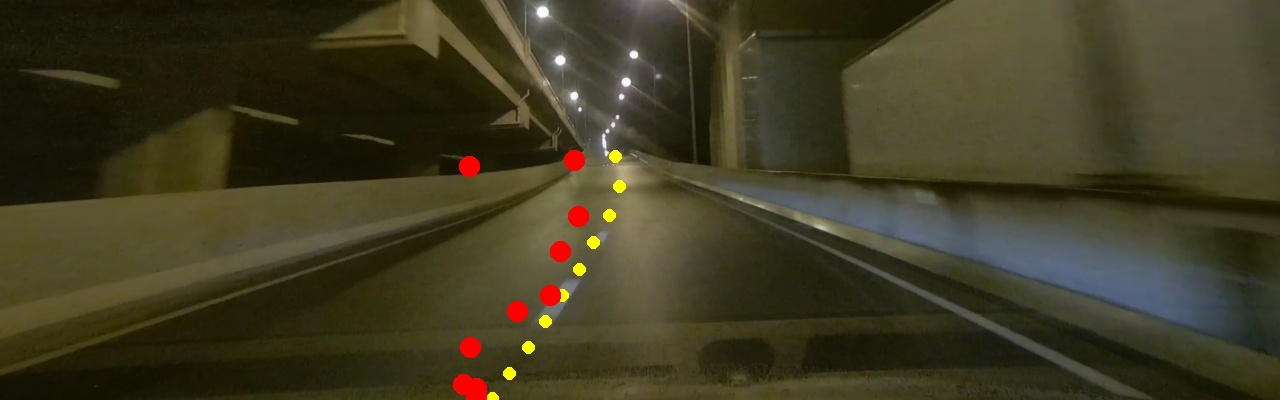}\\[1pt]
      \includegraphics[width=0.495\linewidth]{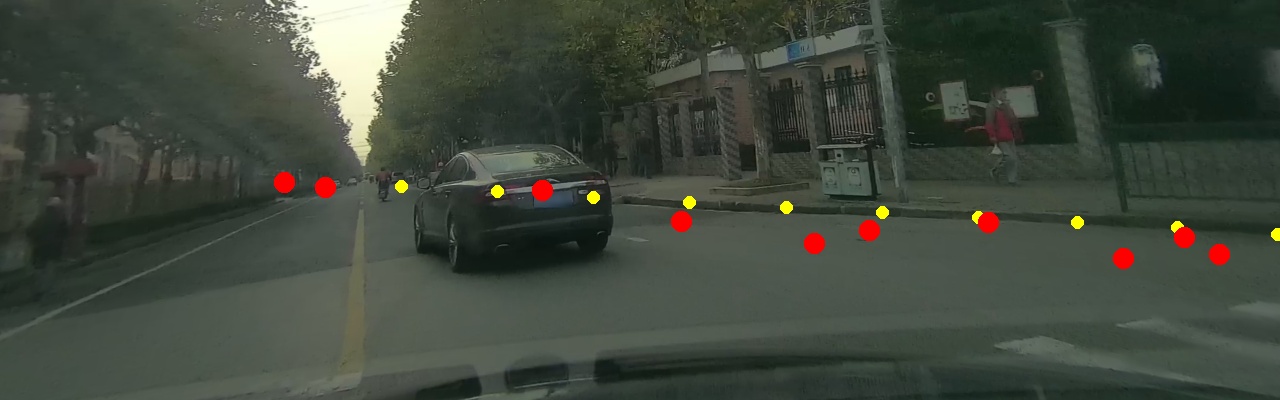}
      \includegraphics[width=0.495\linewidth]{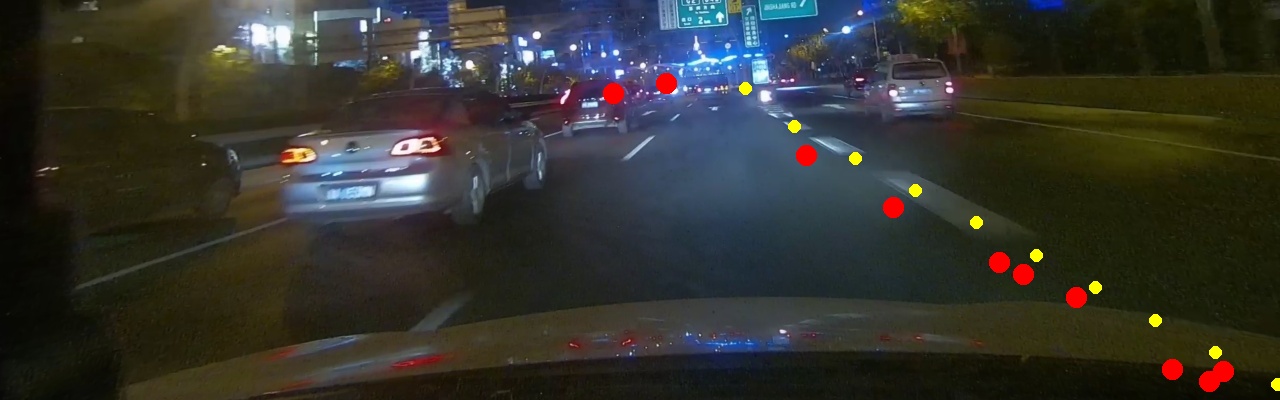}
    \end{minipage}}
  \caption{Distribution of reference points (yellow) and sample points
    (red) in cross-attention modules. In MRDA, the attention tends to
    be distributed along the line, while in ODA, the attention is often
    concentrated near the central point.}
  \label{fig:mrda-vs-oriattn}
\end{figure*}

\subsubsection{Necessity of New Metrics}
\label{sec:necess-verif-new}
To verify the effectiveness of the synthetic metrics, experimental results for
current leading models and \self are presented in
Table~\ref{tab:diff-eval-thr-comparation}. Some valuable insights can be
gained from the results. Although CLRNet reaches the best F1 and precision,
\self has a better recall rate. Additionally, with superior MIoU and MDis, \self
predicts lane positions more accurately. CANet has the highest precision and
MIoU due to its adaptive decoder's great description ability for common lanes,
and a little degradation in recall rate. Overall, the highest F1(0.2, 60) of
\self indicates that some false positive predictions otherwise are correctly recalled, while
the smallest MDis guarantees these predictions are safe.

\subsection{Evaluation of Line IoU Algorithms}
\label{sec:evaluation-two-iou}
Line IoU can be used as a binary matching cost to improve the stability of
matching, or as a loss function to optimize model training.
Table~\ref{tab:iou-ablation} presents \self performance using the two proposed
IoU algorithms in different combinations. Both algorithms adopted as either cost
or loss function outperform the baseline, while P2P is more suitable for cost
and DS for loss. This is because P2P is insensitive to shape jitter, so using
P2P as the cost for bipartite matching can make the training more stable, while
DS can capture subtle differences and is intrinsically suitable for the loss function,
 to optimize prediction details.

\subsection{Ablation Study}
\label{sec:ablation-study}
Table~\ref{tab:ldtr-ablation} presents the results of an ablation study
of the different components of \self, using the CurveLanes dataset. Unlike other lane
description methods that directly perform local computation on the image feature
map, the transformer encoder-decoder query-based structure adopted in this paper
does not have a strong mapping relationship between the query representing each
lane instance and positions in the image. If replacing anchor-chain with
other methods, the entire decoder needs to be removed, making it impossible to
control variables effectively. Therefore, we have to keep anchor-chain in the
baseline. The baseline (1st row) is the DETR-based model with anchor-chain,
which utilizes the classification loss ($L_{\mathrm{cls}}$ as
Equation~\eqref{eq:total_loss}) and regression loss ($L_{\mathrm{reg}}$ as
Equation~\eqref{eq:reg_loss}).

In the 2nd row, the original deformable attention is replaced with MRDA, whose
attention positions are shown in Fig.~\ref{fig:mrda-vs-oriattn}. Because of the
more precise location prior, MRDA can focus on more comprehensive detailed
features, which boosts all metrics in the results.

Particularly, when line IoU is adopted (3rd row) in the loss and matching error, the
calculation optimizes each lane instance as a whole, thus significantly
improving the accuracy and MIoU. In the last row, F1(0.2, 60) and MDis
are improved remarkably, indicating the auxiliary branch can enhance the model
to extract global semantic information.

\begin{table*}[t]
  \caption{Comparative testing on CULane. The first two groups (CNN-based and
    Transformer-based) are measured using the F1 metric, while selected state-of-the-art, SOTA, methods are assessed using F1(0.2, 60) in the final group.}
  \label{tab:culane}
  \centering
  \begin{tabular}{clcccccccccc}
  \hline % --------------------------------------------------------------------------------
  \multicolumn{1}{l}{}
  & Method                           & Total             & Normal            & Crowded           & Dazzle            & Shadow            & No line           & Arrow             & Curve             & Cross            & Night             \\
  \hline % -----------------------------------------------------------------------------------------------------------
  \multirow{8}{*}{\begin{tabular}[c]{@{}c@{}}CNN-based\\(F1)\end{tabular}}
  & SCNN        & 71.60             & 90.60             & 69.70             & 58.50             & 66.90             & 43.40             & 84.10             & 64.40             & 1990             & 66.10             \\
  & CurveLane   & 74.80             & 90.70             & 72.30             & 67.70             & 70.10             & 49.40             & 85.80             & 68.40             & 1746             & 68.90             \\
  & LaneATT     & 77.02             & 91.74             & 76.16             & 69.47             & 76.31             & 50.46             & 86.29             & 64.05             & 1264             & 70.81             \\
  & CondLaneNet & 79.48             & 93.47             & 77.44             & 70.93             & \underline{80.91} & \underline{54.13} & 90.16             & 75.21             & 1201             & 74.80             \\
  & GANet       & 79.63             & \underline{93.67} & 78.66             & \underline{71.82} & 78.32             & 53.38             & 89.86             & \textbf{77.37}    & 1352             & 73.85             \\
  & CANet       & \underline{79.86} & 93.60             & \underline{78.74} & 70.07             & 79.35             & 52.88             & \underline{90.18} & \underline{76.69} & \underline{1196} & \underline{74.91} \\
  & CLRNet      & \textbf{80.47}    & \textbf{93.73}    & \textbf{79.59}    & \textbf{75.30}    & \textbf{82.51}    & \textbf{54.58}    & \textbf{90.62}    & 74.13             & \textbf{1155}    & \textbf{75.37}    \\
  \hline % -----------------------------------------------------------------------------------------------------------
  % \cline{1-1} % ----------------
  \multirow{3}{*}{\begin{tabular}[c]{@{}c@{}}Transformer-\\based\\(F1)\end{tabular}}
  & PriorLane   & 76.27             & \underline{92.36} & 73.86             & 68.26             & \underline{78.13} & \underline{49.60} & \underline{88.59} & \textbf{73.94}    & 2688             & 70.26             \\
  & LaneFormer  & \underline{77.06} & 91.77             & \underline{75.41} & \underline{70.17} & 75.75             & 48.73             & 87.65             & 66.33             & \textbf{19}      & \underline{71.04} \\
  & \self       & \textbf{78.16}    & \textbf{93.22}    & \textbf{75.91}    & \textbf{72.57}    & \textbf{79.53}    & \textbf{53.02}    & \textbf{88.70}    & \underline{70.41} & \underline{1352} & \textbf{73.66}    \\
  \hline % -----------------------------------------------------------------------------------------------------------
  \multirow{3}{*}{\begin{tabular}[c]{@{}c@{}}SOTA\\(F1(0.2, 60))\end{tabular}}
  & CANet      & 81.85             & \underline{95.45} & 80.57             & 77.43             & 77.99             & 55.45             & 91.68             & \textbf{71.22}    & \underline{1196} & 77.10             \\
  & CLRNet     & \underline{82.11} & 94.54             & \underline{80.85} & \underline{80.96} & \underline{81.14} & \underline{58.66} & \underline{91.85} & 58.33             & \textbf{1155}    & \underline{78.80} \\
  & \self     & \textbf{84.18}    & \textbf{96.12}    & \textbf{83.27}    & \textbf{81.49}    & \textbf{87.39}    & \textbf{62.93}    & \textbf{92.06}    & \underline{62.77} & 1352             & \textbf{81.52}    \\
  \hline % -----------------------------------------------------------------------------------------------------------
  \end{tabular}
\end{table*}

\subsection{Performance on Datasets}

\subsubsection{Results on CULane}
\label{sec:results-culane}
As mentioned in Section~\ref{sec:datasets}, in CULane, many lanes have little- or
no-visual-clue, so it is suitable to distinguish the recall
capabilities of different models. Table~\ref{tab:culane} presents our
comprehensive experimental results. Models are clustered according to their
basic techniques into CNN-based and transformer-based groups, and best results are
marked in bold in each group. \self is ahead of other Transformer-based
models in almost all categories, validating the effectiveness of \self's network structure
design. However, it lags behind CLRNet because CNN-based models often predict
fewer false positives, which results in higher precision and F1. As discussed in
Section~\ref{sec:perf-indic}, F1 and precision are not the most appropriate metrics for downstream
tasks using lane detection, so an additional test measured F1(0.2, 60)
for various leading methods. \self outperforms both CANet and CLRNet by
2.33 and 2.07 overall, respectively, especially for scene types with fewer visual
clues such as Crowded, Dazzle, and Shadow.
It is worth noting that \self 's F1(0.2, 60) score is significantly improved
(by 6.02 percentage points) compared to its overall F1 score. This is because,
for the no-visual-clue scenarios that exist in CULane, \self 's predictions
are recalled more frequently using a more reasonable true positive standard. CLRNet
and CANet have weaker recall performance in this scenario, resulting in poorer
performance than \self as assessed by F1(0.2, 60).

\begin{table}[ht]
  \caption{Performance comparison on CurveLanes. The first group models are
    measured in F1, while the second group is in F1(0.2, 10) (marked with
    ``*'').}
  \label{tab:curvelanes}
  \centering
    \begin{tabular}{lcccr}
      \hline % --------------------------------------------------------------------------------
      Models        & F1             & Precision      & Recall         & FPS    \\
      \hline % --------------------------------------------------------------------------------
      SCNN          & 65.02          & 76.13          & 56.74          & 7.5    \\
      ENet-SAD      & 50.31          & 63.60          & 41.60          & 75     \\
      PointLaneNet  & 78.47          & 86.33          & 72.91          & 71     \\
      CurveLane   & 82.29          & 91.11          & 75.03          & -      \\
      CondLaneNet & 86.10          & 88.98          & 83.41          & 48     \\
      % CLRerNet      & 86.47          & 91.66          & 81.83          & 185   \\
      CANet       & 87.87          & \textbf{91.69} & 84.36          & 36.6   \\
      \self         & \textbf{88.44} & 91.55          & \textbf{85.53} & 25.2   \\
      \hline % --------------------------------------------------------------------------------
      CANet*      & 88.48          & 92.33          & 84.95          & 36.6   \\
      \self*        & \textbf{89.66} & \textbf{92.82} & \textbf{86.72} & 25.2   \\
      \hline % --------------------------------------------------------------------------------
    \end{tabular}
\end{table}  % tab:curvelanes

\subsubsection{Results on CurveLanes}
\label{sec:results-curvelanes}
Compared to CULane, the CurveLanes dataset covers a wider range of scenes and
has more complicated lane shapes, so provides a better test of 
lane shape modeling ability. Table~\ref{tab:curvelanes} shows the test
results in detail. \self performs best in terms of both F1 and F1(0.2, 10),
especially for recall rate, which is emphasized by \self.

\begin{table}[t]
  \caption{Average performance comparison on CULane and CurveLanes. As CLRNet
    did not provide metrics and trained weights on CurveLanes, it is not
    included.}
  \label{tab:avg_performance}
  \centering
  \begin{tabular}{llccc}
  \hline
  Datasets & Models      & AF1               & AP                & AR                \\
  \hline % --------------------------------------------------------------------------------
  \multirow{3}{*}{CULane}     
  & CANet & \underline{62.97} & \textbf{69.85}    & \underline{57.33} \\
  & CLRNet  & 58.99             & 63.87             & 54.81             \\
  & \self   & \textbf{63.65}    & \underline{66.39} & \textbf{61.12}    \\
  \hline % --------------------------------------------------------------------------------
  \multirow{2}{*}{CurveLanes} 
  & CANet & 58.55             & 61.10             & 56.22             \\
  & \self   & \textbf{62.76}    & \textbf{65.06}    & \textbf{60.62}    \\
    \hline % --------------------------------------------------------------------------------
  \end{tabular}
\end{table}  % tab:avg_performance

\subsubsection{Average Performance}
\label{sec:average-perf}
All previous experiments were executed with fixed hyperparameters $\alpha$ and
$\beta$. To evaluate the capability of \self with different hyperparameter
configurations, we borrowed the COCO~\cite{coco} object detection dataset performance
indicators AP and AR and define AF1 (average F1-score) similarly. By conducting
extensive experiments with many $\alpha$, $\beta$ combinations,
Table~\ref{tab:avg_performance} shows that \self surpasses other networks on
average, implying the effectiveness of its architectural design, independent of
specific parameter settings.

\begin{figure*}[t]
  \centering
  \subfigure[IoU threshold=0]{%\label{fig:dense-iou-split-v}%
    \includegraphics[width=.33\linewidth]{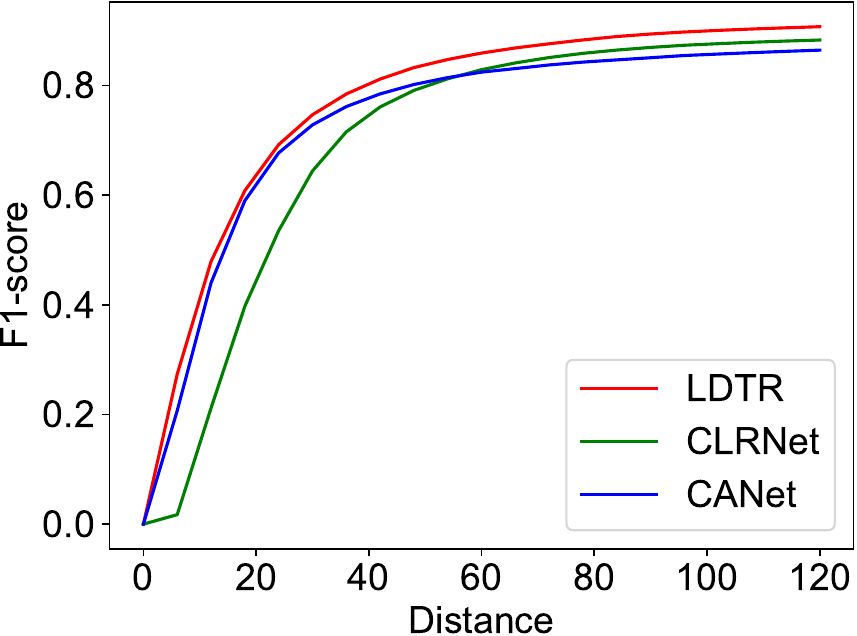}}\hfill
  \subfigure[IoU threshold=0.2]{%\label{fig:dense-iou-split-v}%
    \includegraphics[width=.33\linewidth]{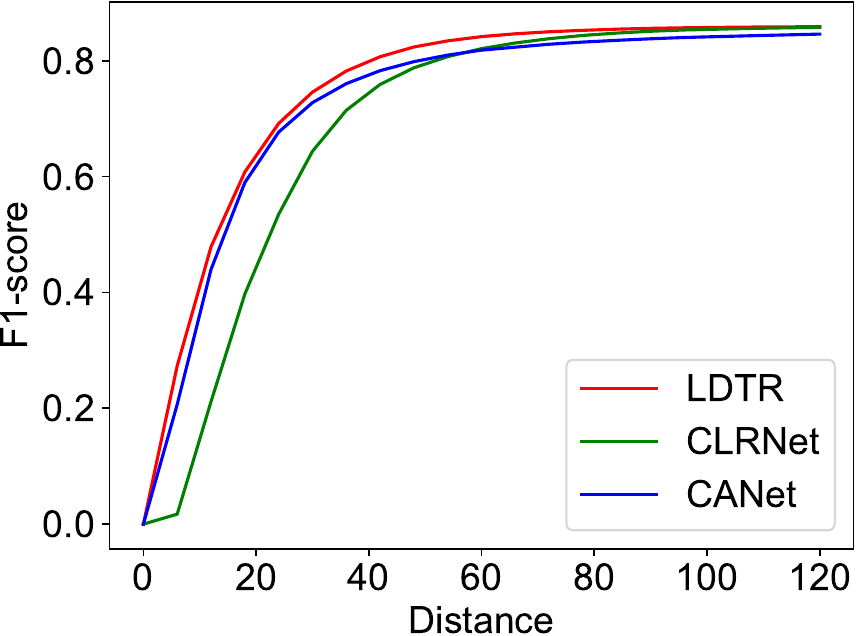}}\hfill
  \subfigure[IoU threshold=0.5]{%\label{fig:dense-iou-split-v}%
    \includegraphics[width=.33\linewidth]{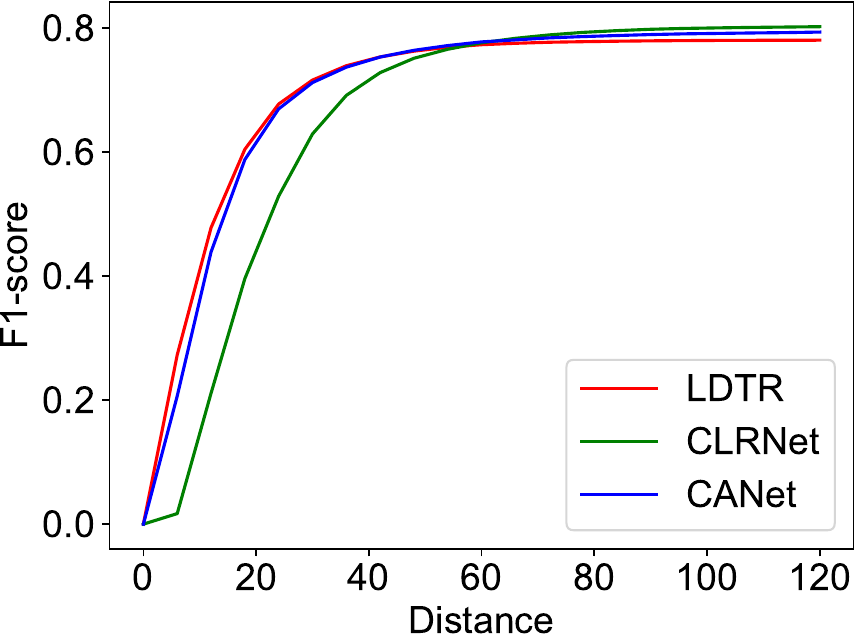}}\hfill
  \caption{F1-scores of different models vs. Fréchet distance thresholds on
    CULane.}
  \label{fig:different_dis_thr_culane}
\end{figure*}  % fig:different_dis_thr_culane
\begin{figure*}[t]
  \centering
  \subfigure[IoU threshold=0]{%\label{fig:dense-iou-split-v}%
    \includegraphics[width=.33\linewidth]{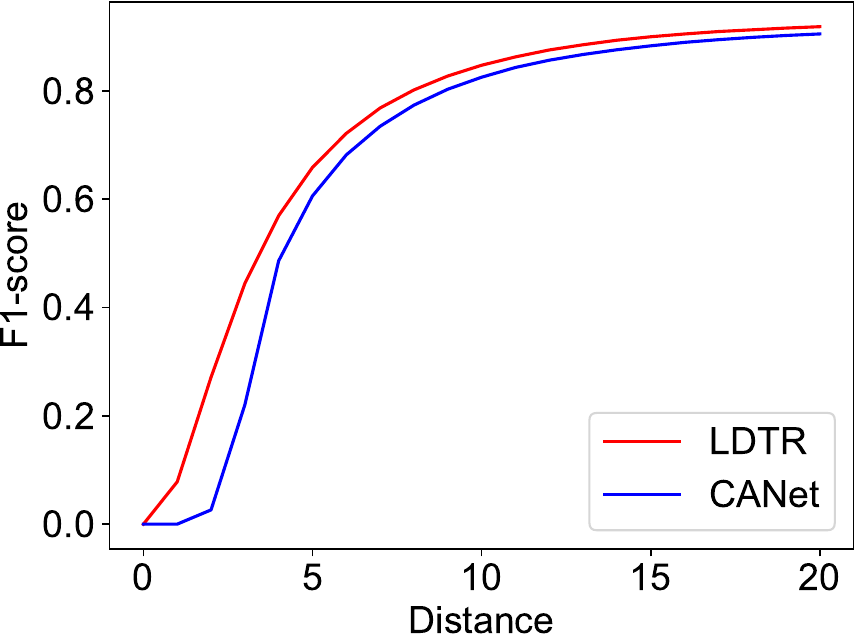}}\hfill
  \subfigure[IoU threshold=0.2]{%\label{fig:dense-iou-split-v}%
    \includegraphics[width=.33\linewidth]{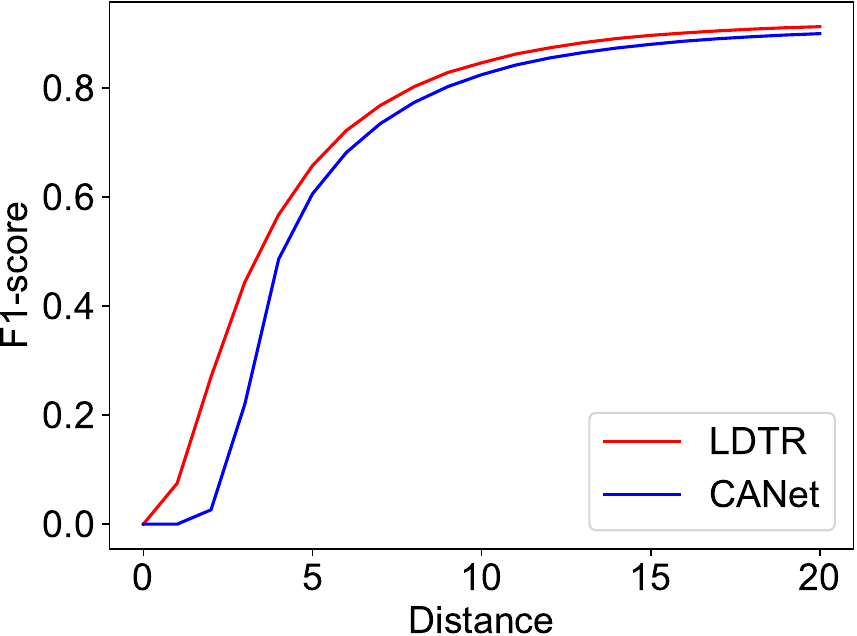}}\hfill
  \subfigure[IoU threshold=0.5]{%\label{fig:dense-iou-split-v}%
    \includegraphics[width=.33\linewidth]{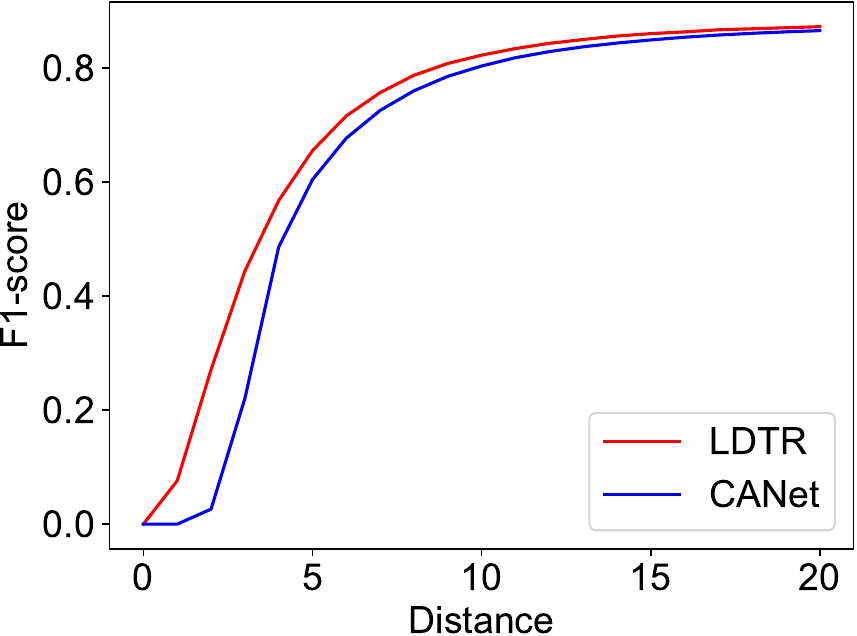}}\hfill
  \caption{F1-scores of different models vs. Fréchet distance thresholds on
    CurveLanes. As CLRNet did not provide metrics and trained weights on
    CurveLanes, it is not included.}
  \label{fig:different_dis_thr_curvelanes}
\end{figure*}  % fig:different_dis_thr_curvelanes

To thoroughly understand how the Fréchet distance threshold is determined,
Figs.~\ref{fig:different_dis_thr_culane} and~\ref{fig:different_dis_thr_curvelanes} present how F1-score varies with Fréchet distance thresholds with different IoU settings for the two datasets,
respectively. \self generally performs better than the other two models,
especially, when the IoU threshold is small. As the IoU threshold increases,
no-visual-clue lanes are more likely to be dropped, thus the advantage of \self
gradually diminishes. These figures indicate that \self exceeds in F1-score over a wide range of
 Fréchet distance thresholds.

\begin{figure*}[!t]
\centering
\begin{minipage}[b]{\linewidth}
  \rotatebox{90}{\parbox[c][0.05\linewidth][c]{0.11\linewidth}{\centering CLRNet F1(0.5, $+\infty$)}}
  \includegraphics[width=0.313\linewidth]{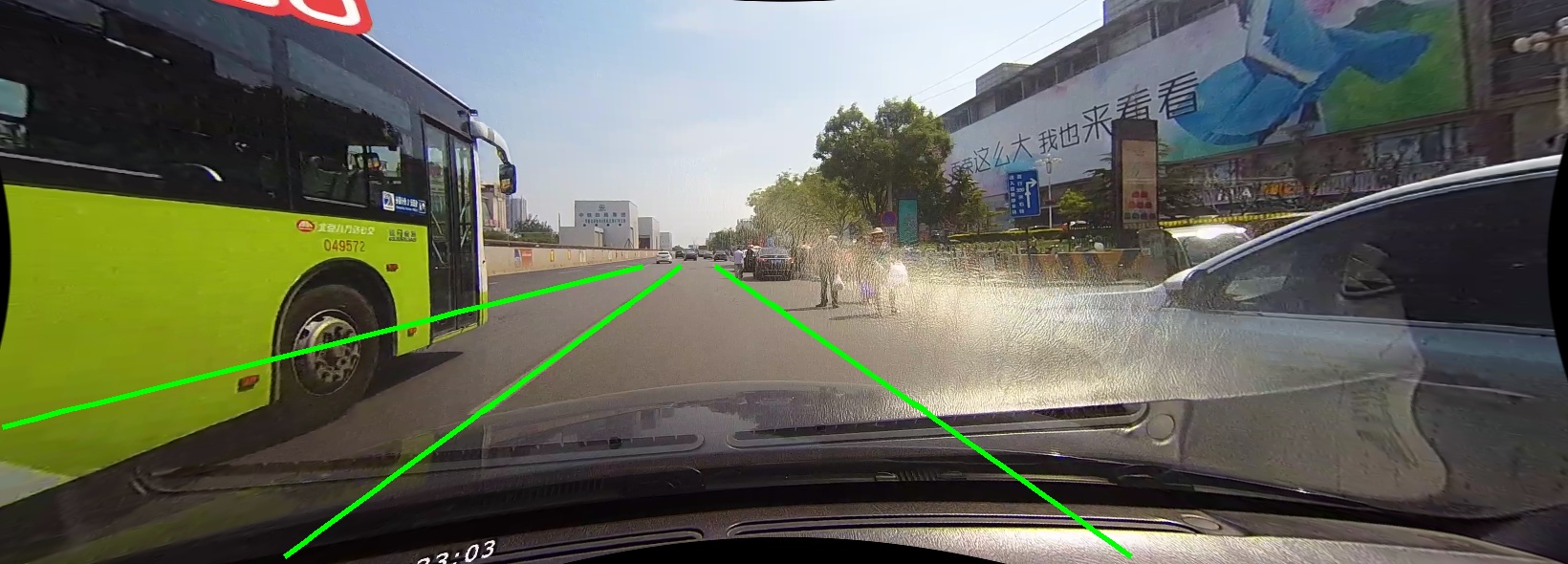}\hfill
  \includegraphics[width=0.313\linewidth]{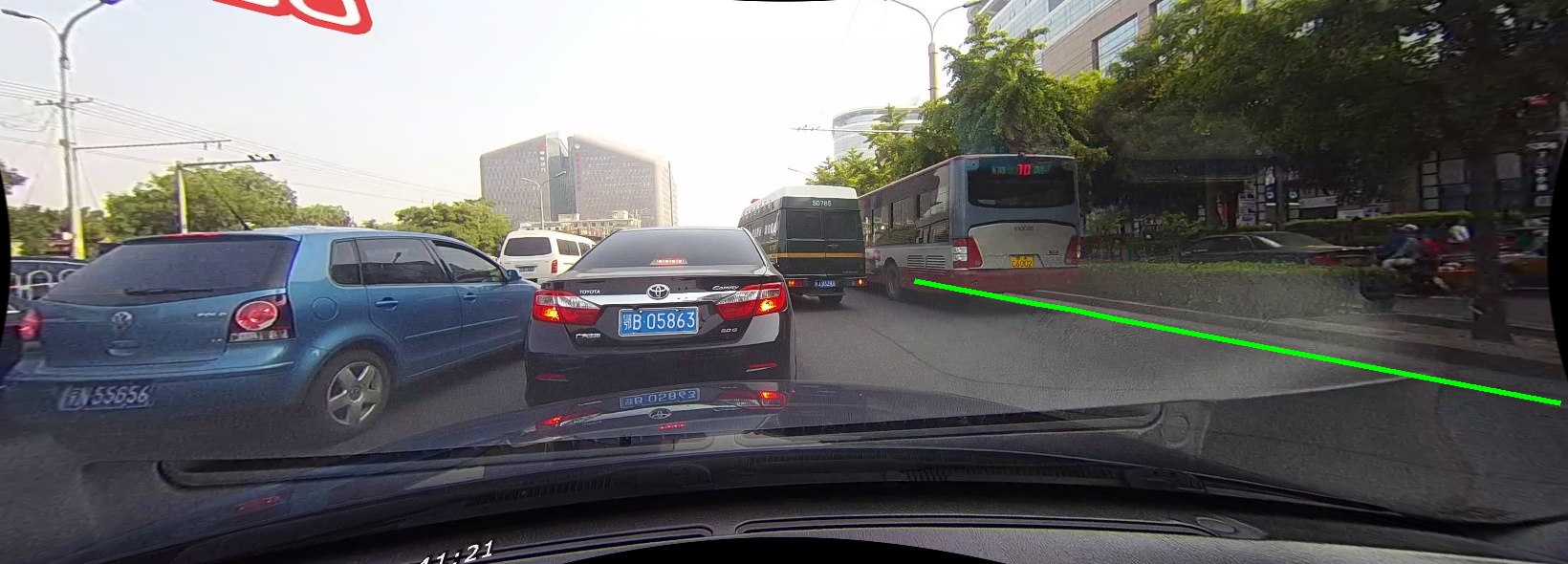}\hfill
  \includegraphics[width=0.313\linewidth]{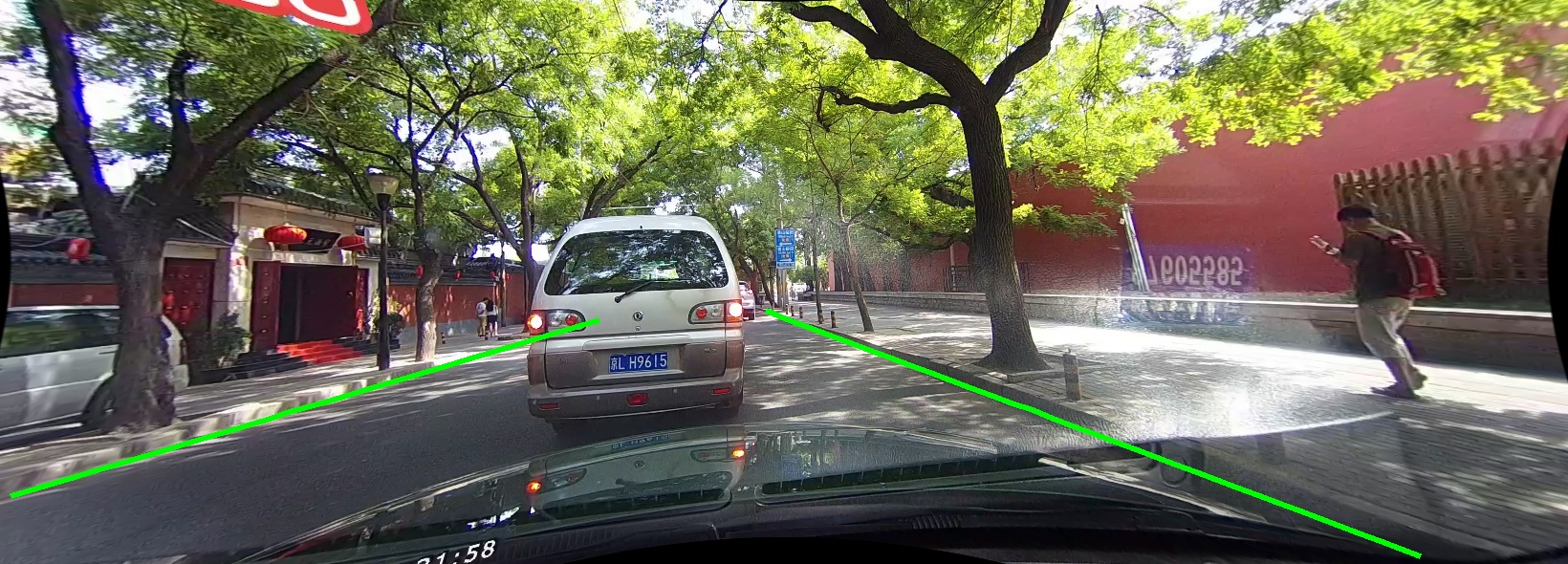}\hfill
\end{minipage}\hfill \\
\begin{minipage}[b]{\linewidth}
  \rotatebox{90}{\parbox[c][0.05\linewidth][c]{0.11\linewidth}{\centering \self F1(0.5, $+\infty$)}}
  \includegraphics[width=0.313\linewidth]{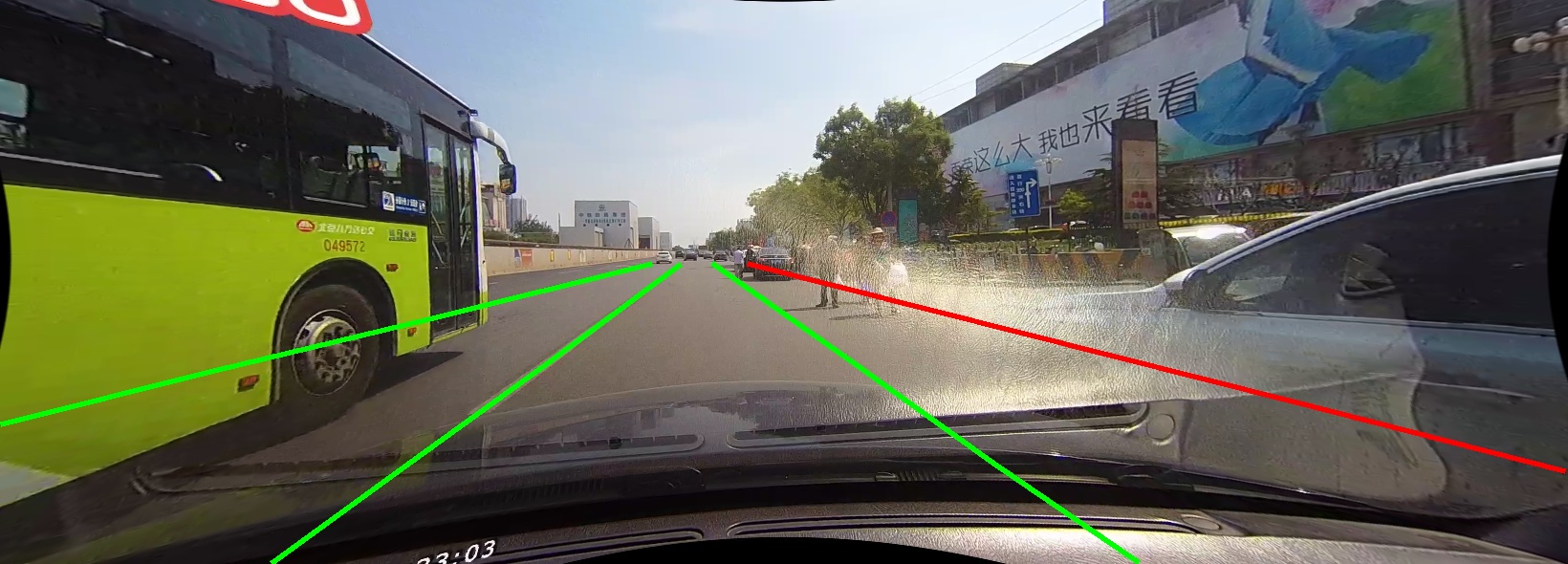}\hfill
  \includegraphics[width=0.313\linewidth]{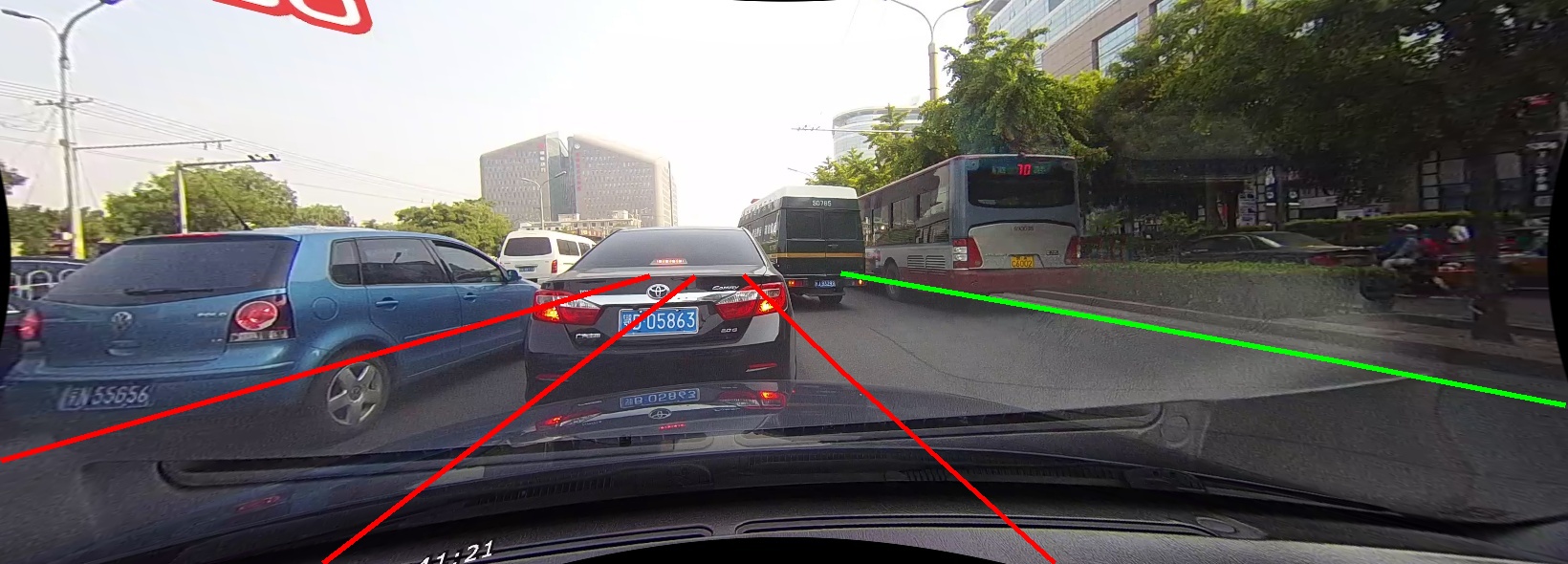}\hfill
  \includegraphics[width=0.313\linewidth]{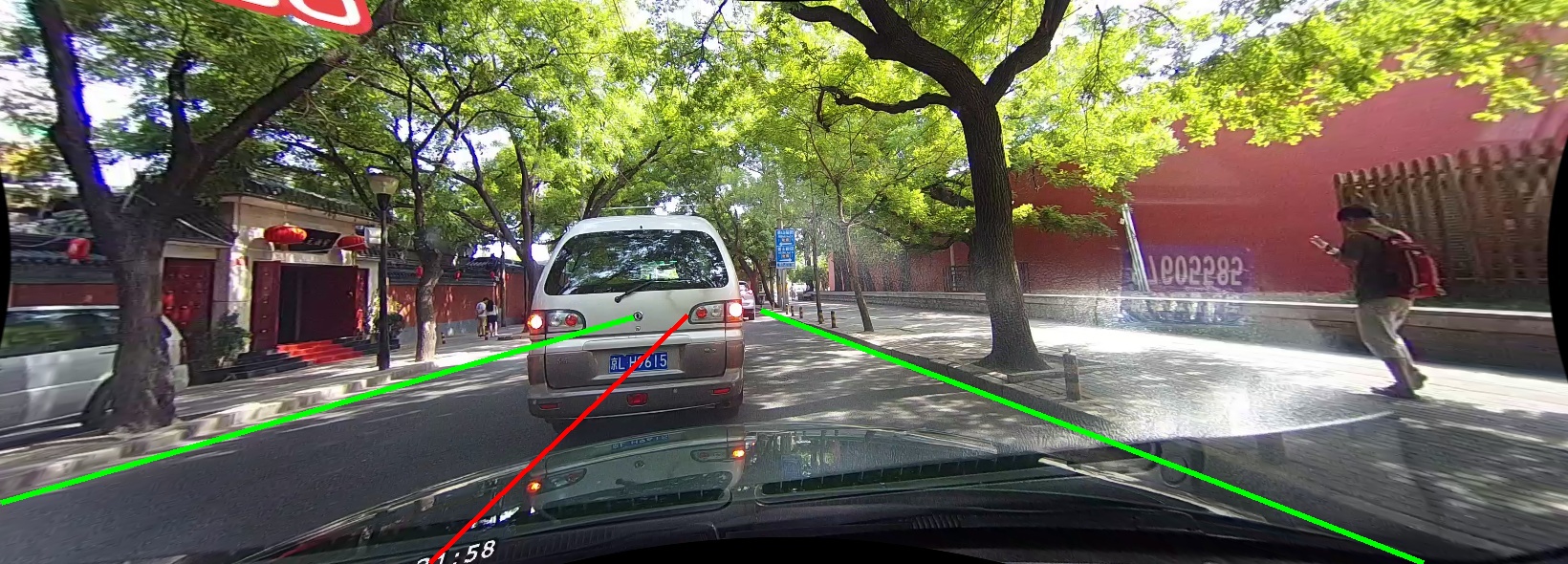}\hfill
\end{minipage}\hfill \\
\begin{minipage}[b]{\linewidth}
  \rotatebox{90}{\parbox[c][0.05\linewidth][c]{0.11\linewidth}{\centering \self F1(0.2, 60)}}
  \includegraphics[width=0.313\linewidth]{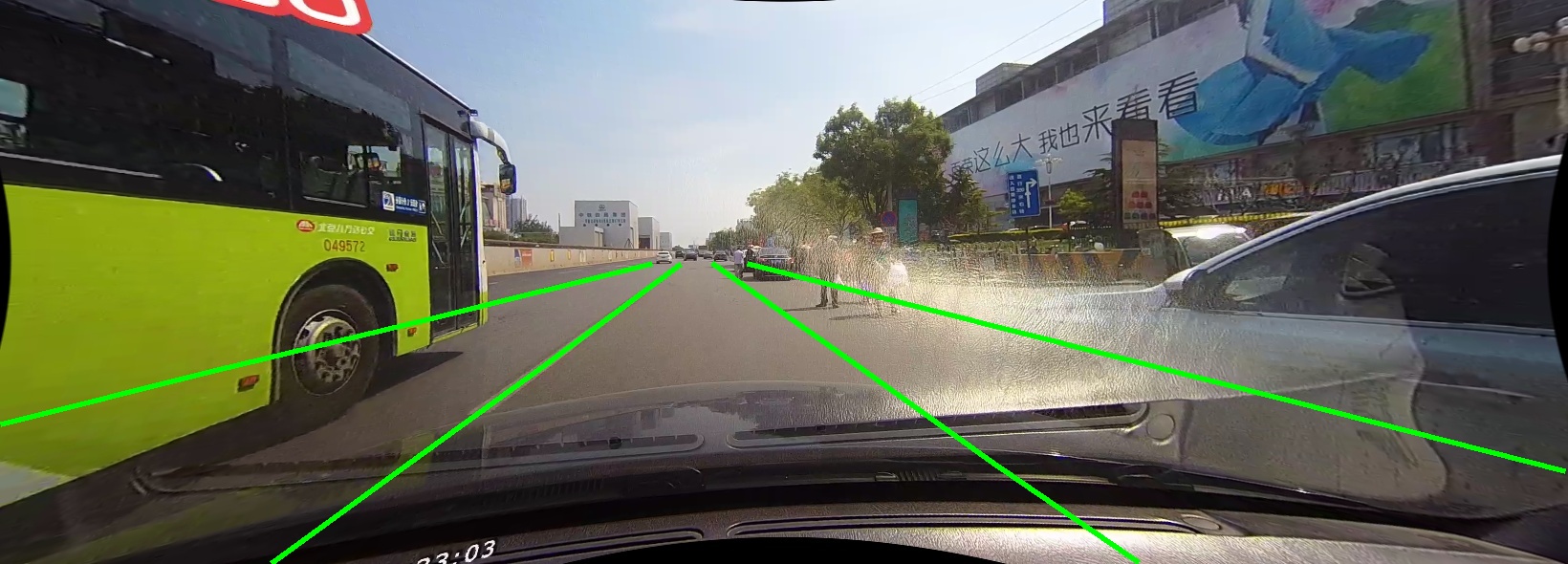}\hfill
  \includegraphics[width=0.313\linewidth]{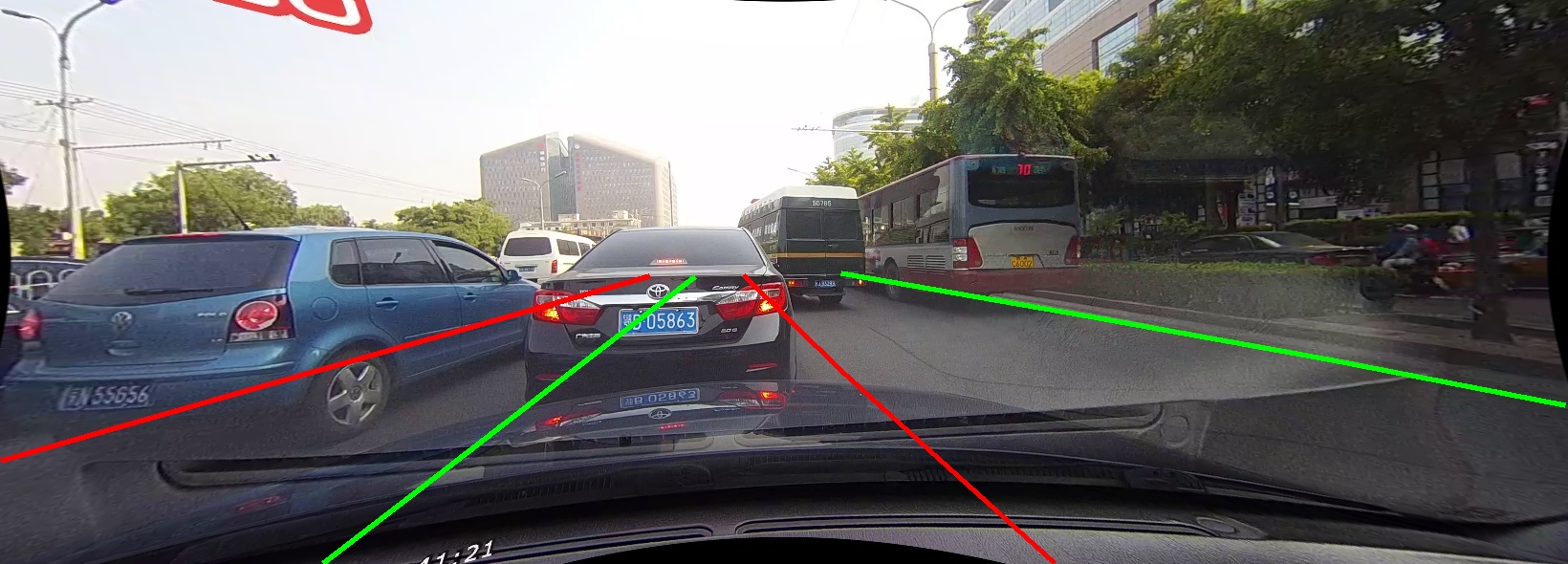}\hfill
  \includegraphics[width=0.313\linewidth]{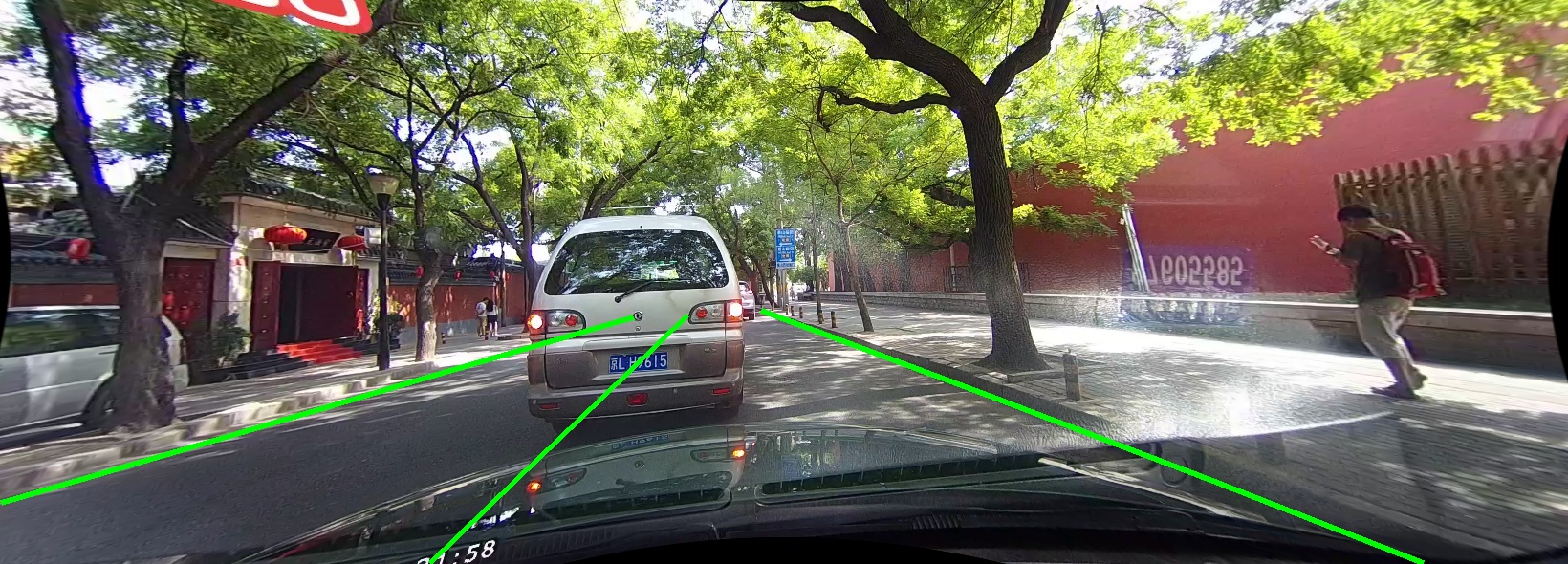}\hfill
\end{minipage}\hfill \\
\begin{minipage}[b]{\linewidth}
  \rotatebox{90}{\parbox[c][0.05\linewidth][c]{0.11\linewidth}{\centering GT}}
  \includegraphics[width=0.313\linewidth]{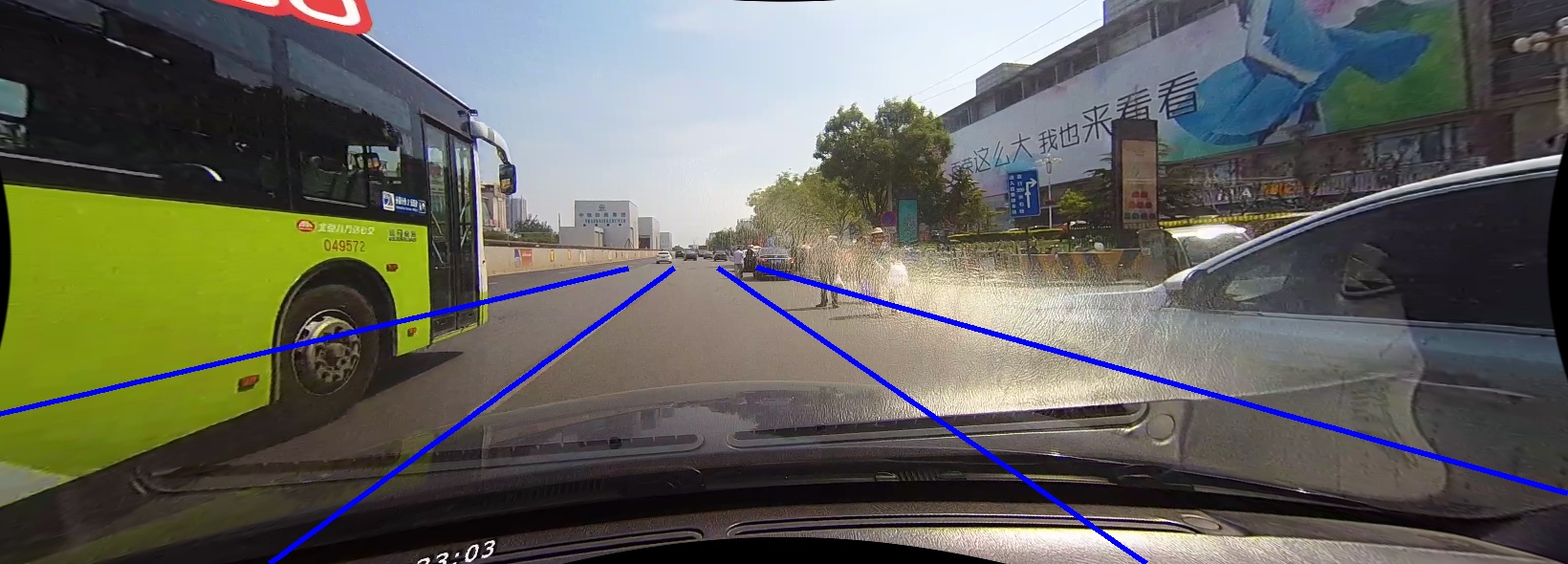}\hfill
  \includegraphics[width=0.313\linewidth]{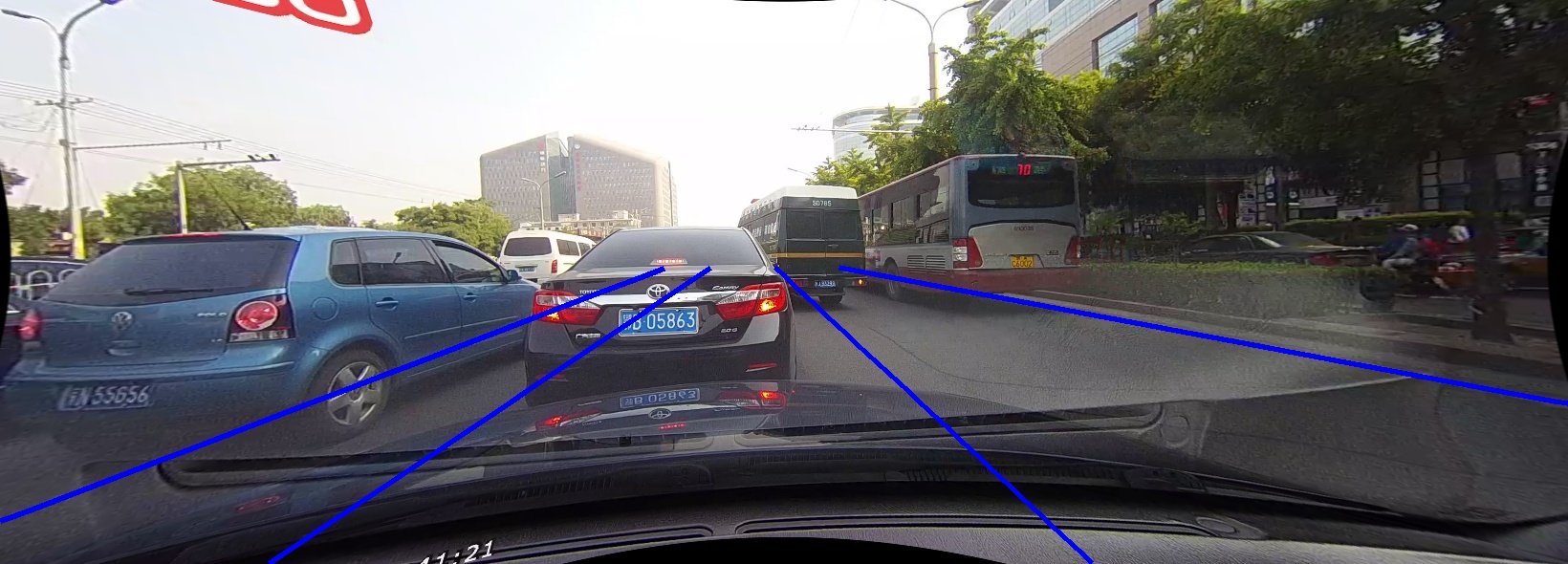}\hfill
  \includegraphics[width=0.313\linewidth]{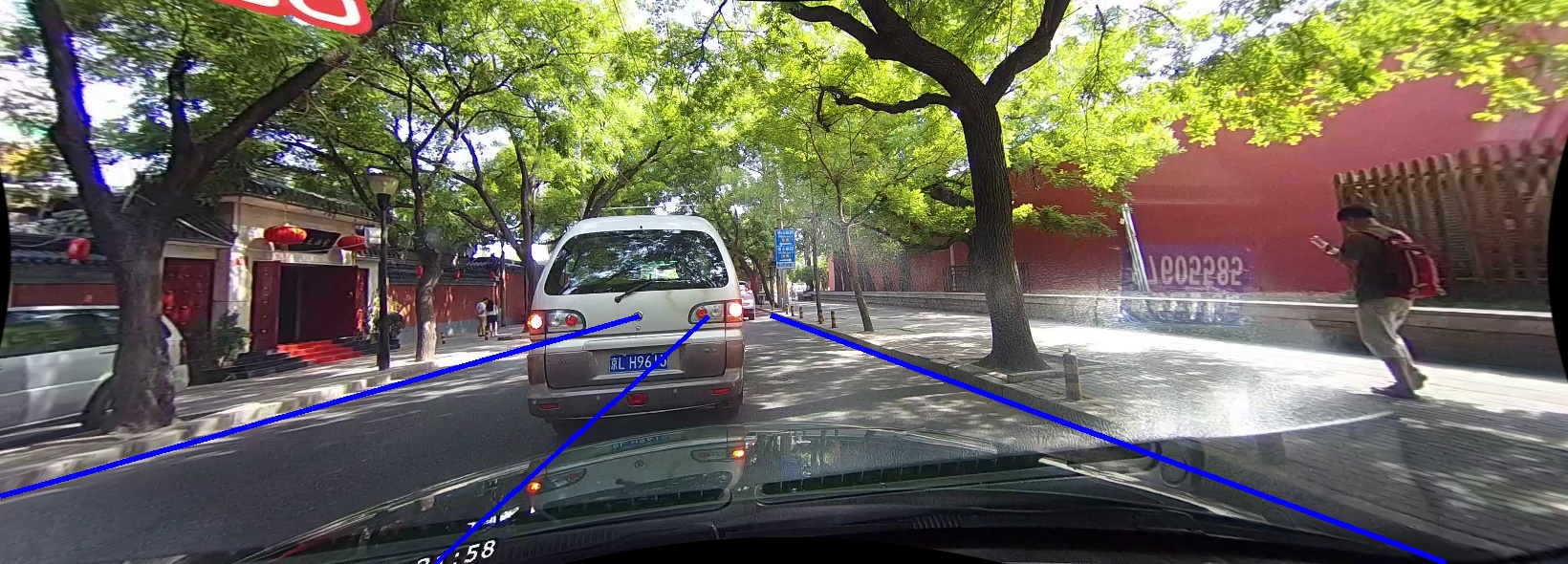}\hfill
\end{minipage}\hfill \\
\begin{minipage}[b]{\linewidth}
  \rotatebox{90}{\parbox[c][0.05\linewidth][c]{0.11\linewidth}{}}
  \rotatebox{0}{\parbox[c][0.05\linewidth][c]{0.313\linewidth}{\centering Frame 1}}
  \rotatebox{0}{\parbox[c][0.05\linewidth][c]{0.313\linewidth}{\centering Frame 2}}
  \rotatebox{0}{\parbox[c][0.05\linewidth][c]{0.313\linewidth}{\centering Frame 3}}
\end{minipage}\hfill \\

\begin{minipage}[b]{\linewidth}
  \rotatebox{90}{\parbox[c][0.05\linewidth][c]{0.11\linewidth}{\centering CLRNet F1(0.5, $+\infty$)}}
  \includegraphics[width=0.313\linewidth]{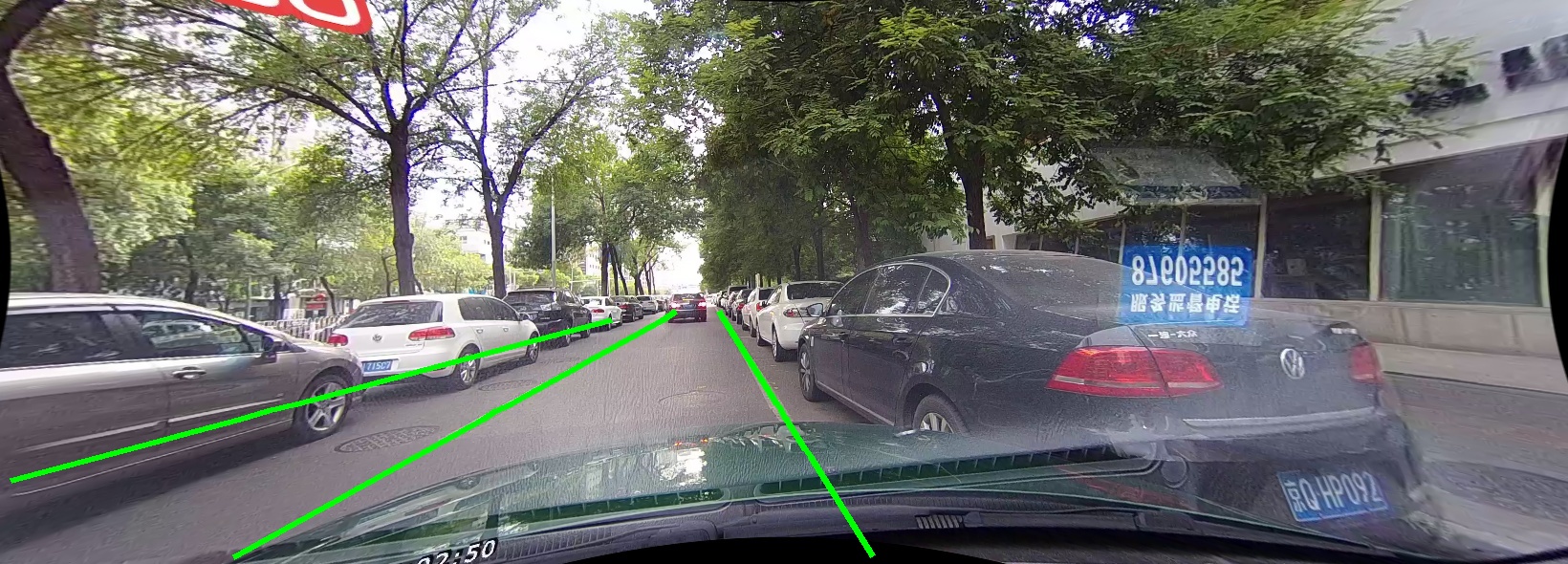}\hfill
  \includegraphics[width=0.313\linewidth]{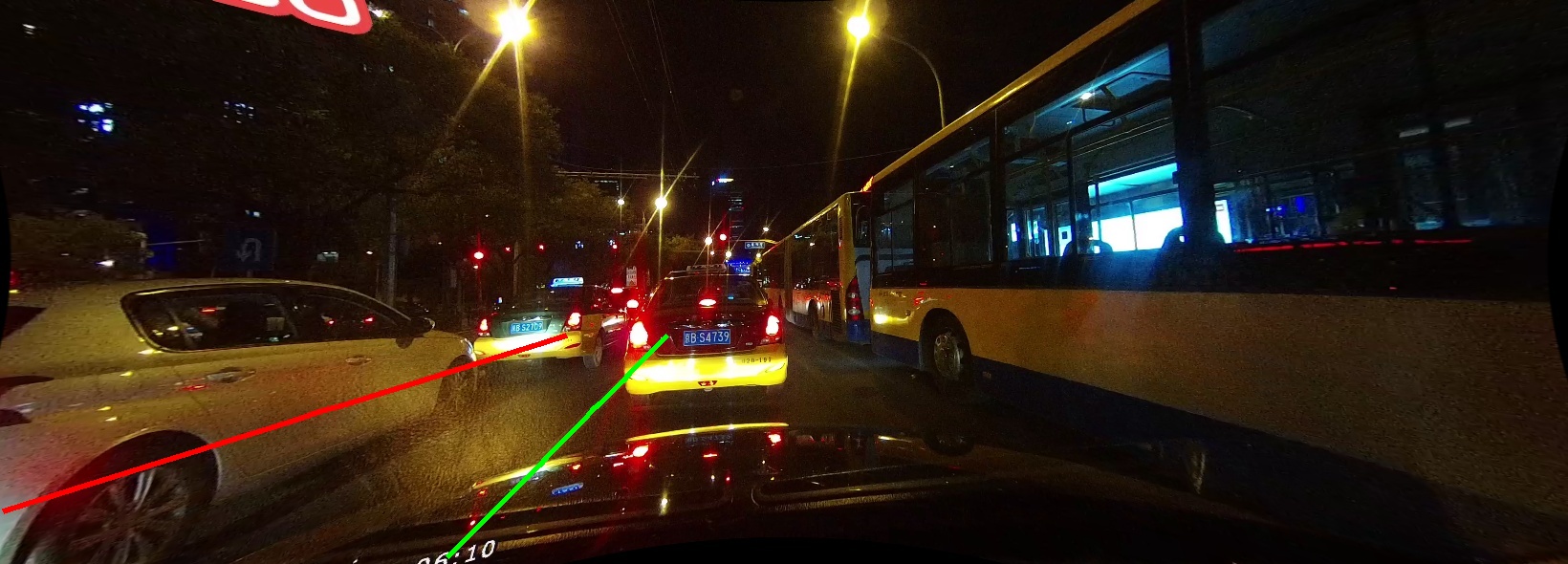}\hfill
  \includegraphics[width=0.313\linewidth]{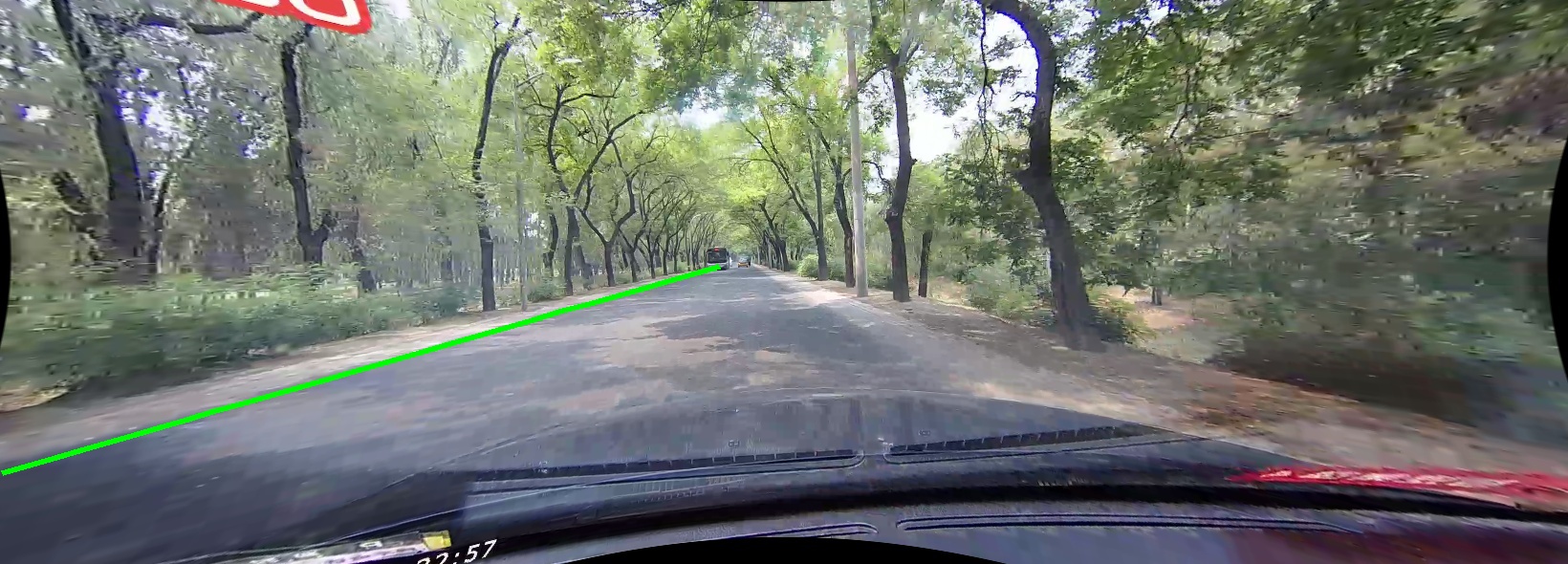}\hfill
\end{minipage}\hfill \\
\begin{minipage}[b]{\linewidth}
  \rotatebox{90}{\parbox[c][0.05\linewidth][c]{0.11\linewidth}{\centering \self F1(0.5, $+\infty$)}}
  \includegraphics[width=0.313\linewidth]{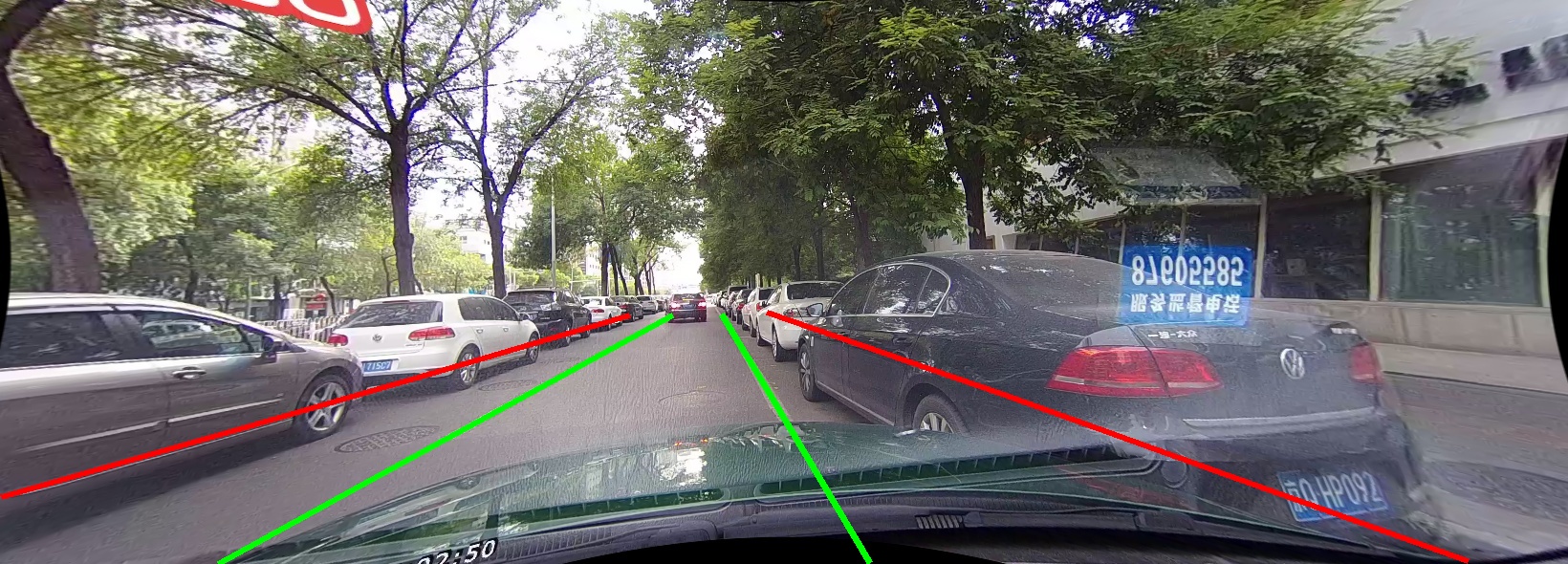}\hfill
  \includegraphics[width=0.313\linewidth]{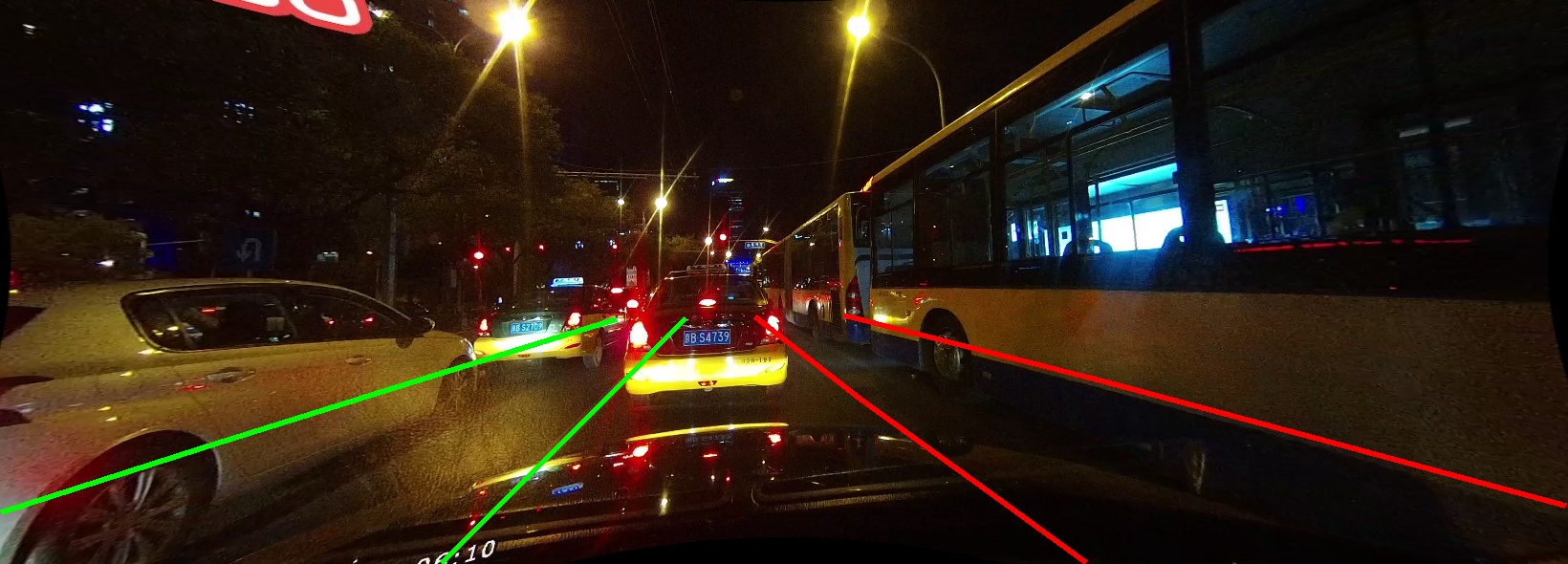}\hfill
  \includegraphics[width=0.313\linewidth]{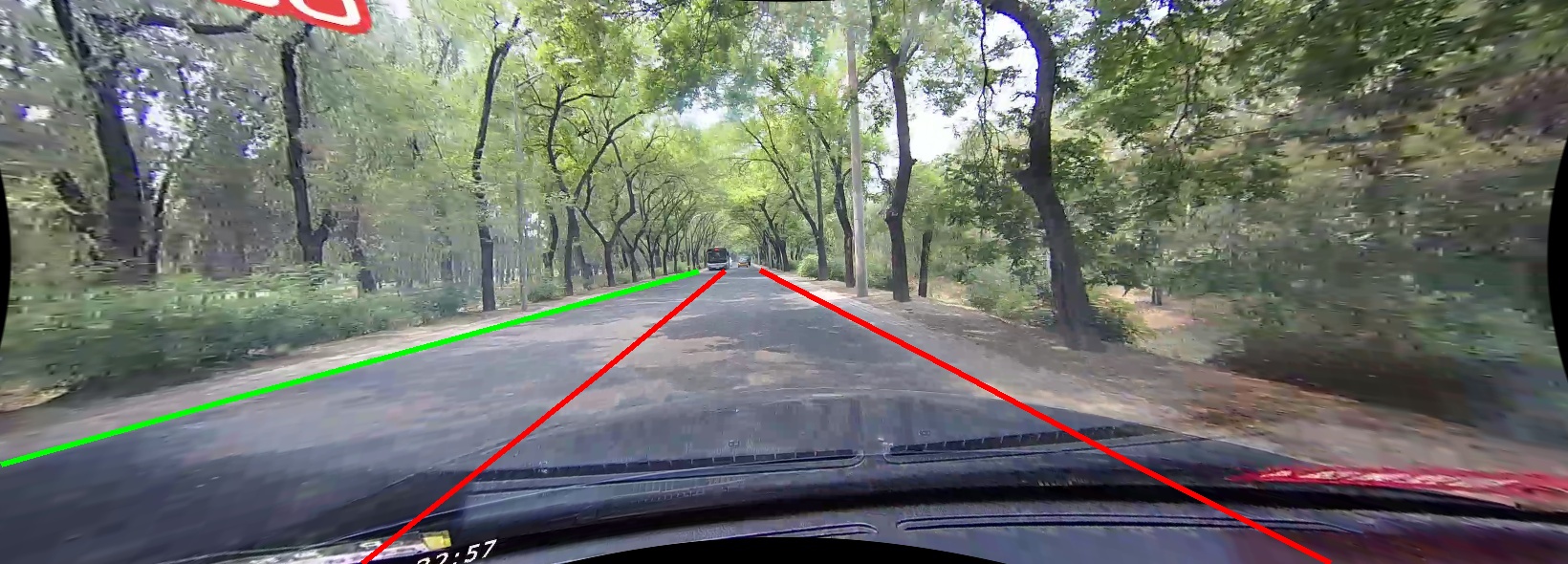}\hfill
\end{minipage}\hfill \\
\begin{minipage}[b]{\linewidth}
  \rotatebox{90}{\parbox[c][0.05\linewidth][c]{0.11\linewidth}{\centering \self F1(0.2, 60)}}
  \includegraphics[width=0.313\linewidth]{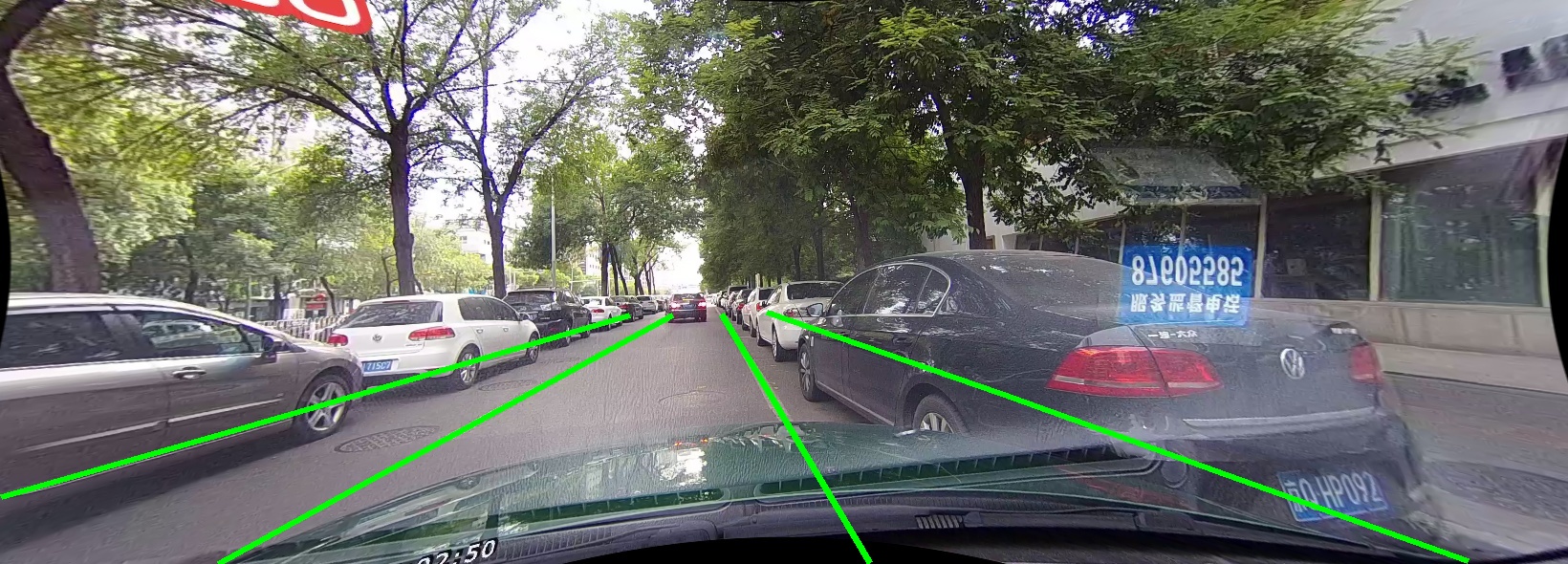}\hfill
  \includegraphics[width=0.313\linewidth]{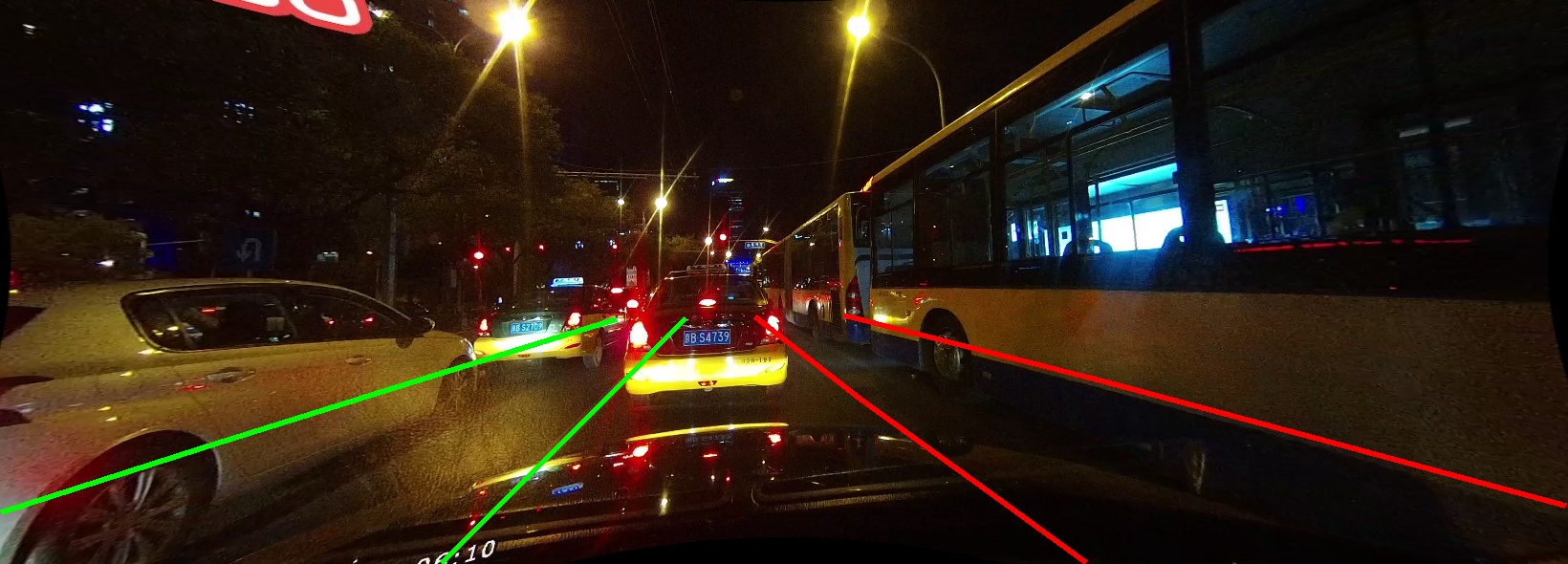}\hfill
  \includegraphics[width=0.313\linewidth]{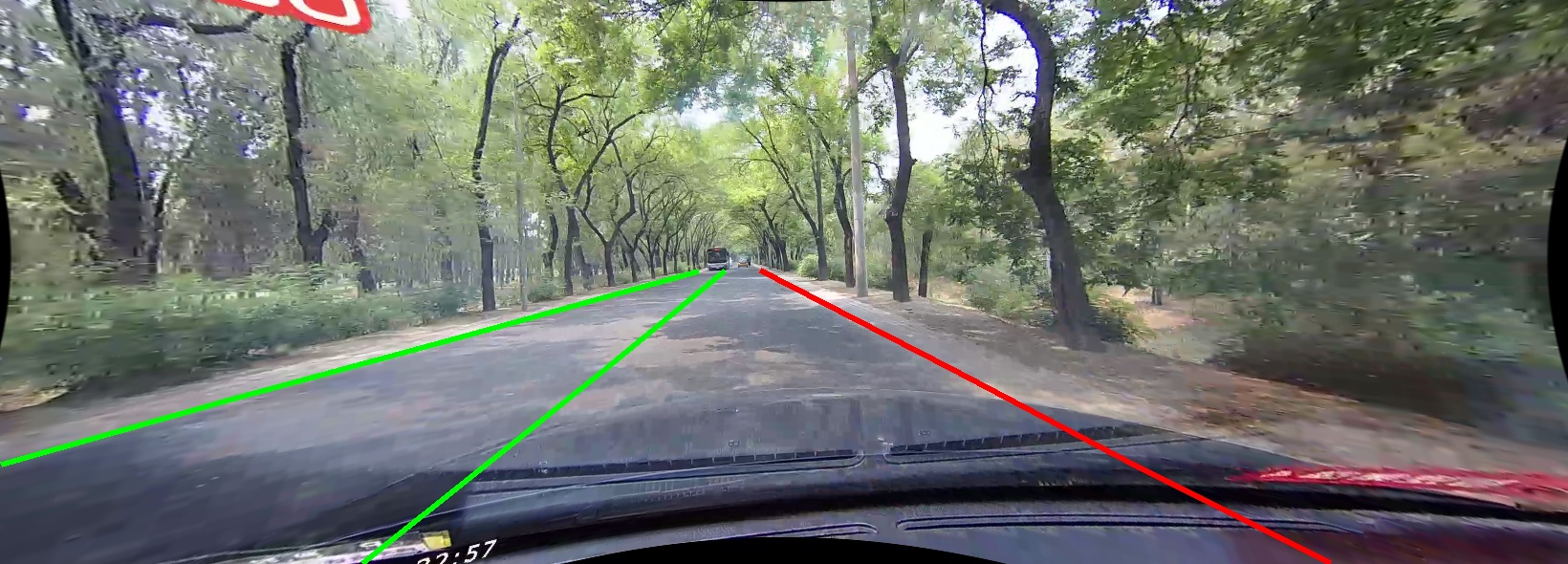}\hfill
\end{minipage}\hfill \\
\begin{minipage}[b]{\linewidth}
  \rotatebox{90}{\parbox[c][0.05\linewidth][c]{0.11\linewidth}{\centering GT}}
  \includegraphics[width=0.313\linewidth]{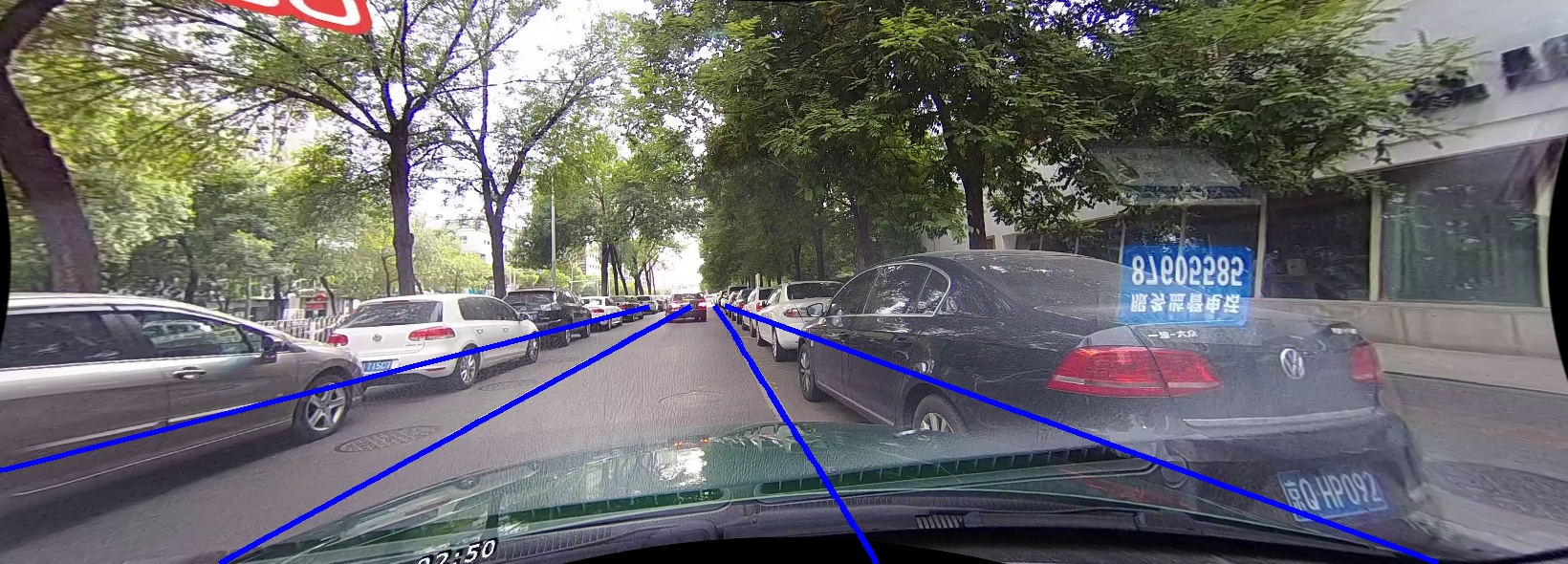}\hfill
  \includegraphics[width=0.313\linewidth]{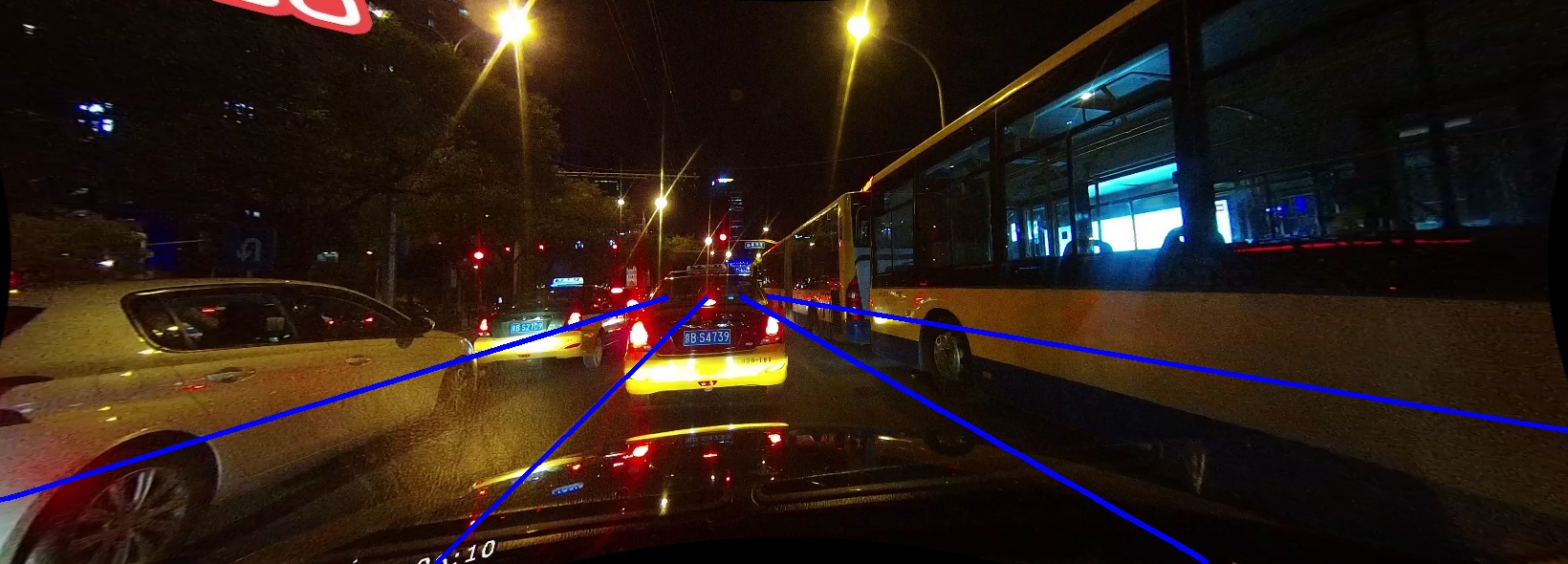}\hfill
  \includegraphics[width=0.313\linewidth]{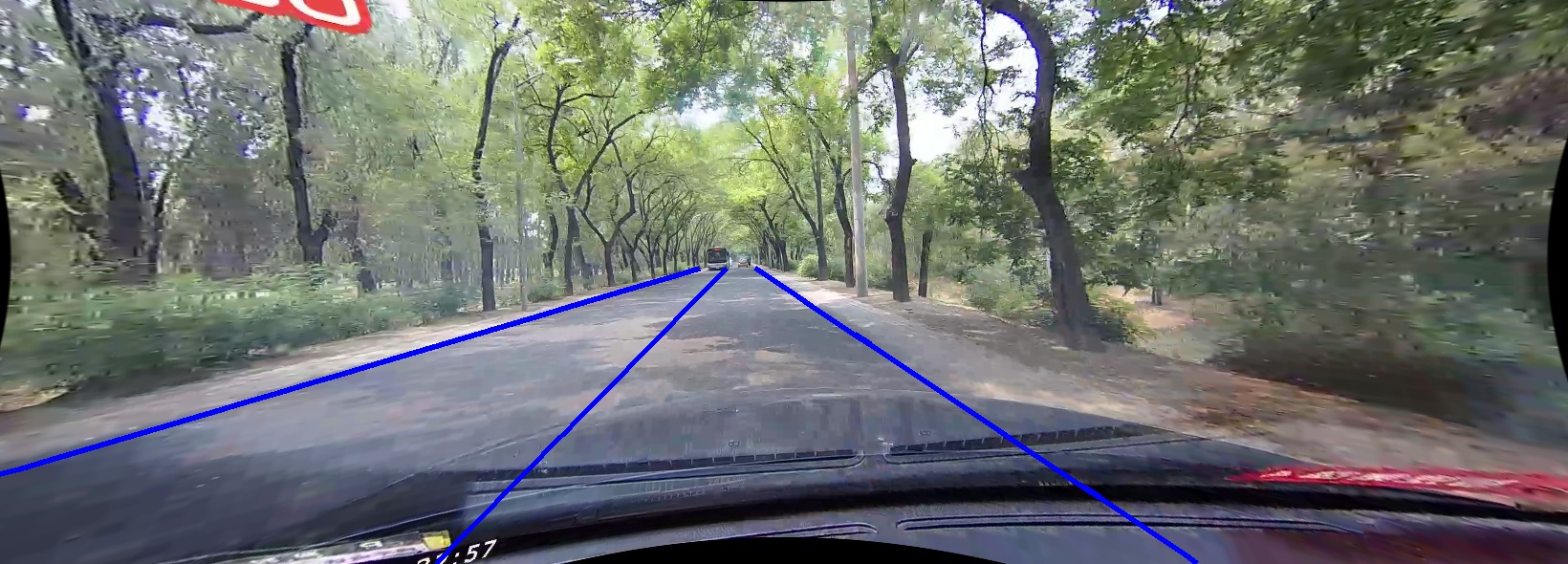}\hfill
\end{minipage}\hfill \\
\begin{minipage}[b]{\linewidth}
  \rotatebox{90}{\parbox[c][0.05\linewidth][c]{0.11\linewidth}{}}
  \rotatebox{0}{\parbox[c][0.05\linewidth][c]{0.313\linewidth}{\centering Frame 4}}
  \rotatebox{0}{\parbox[c][0.05\linewidth][c]{0.313\linewidth}{\centering Frame 5}}
  \rotatebox{0}{\parbox[c][0.05\linewidth][c]{0.313\linewidth}{\centering Frame 6}}
\end{minipage}\hfill \\

\caption{Prediction ability comparison between CLRNet and \self for ``Congested
      Scenarios'' in CULane. The green and red lines represent the true positive (TP) and false positive (FP) of the model's prediction results under a specific evaluation indicator, respectively, while the ground truth (GT) is represented by blue lines.}
\label{fig:visual_compare}
\end{figure*}  % fig:visual_compare

\section{Visual Comparison}
For a more intuitive comparison, we have selected and included several frames in
Fig.~\ref{fig:visual_compare}. These scenes include different lighting
conditions, occlusions and most  have no-visual-clue for lanes. Considering  the ground truth shows that \self can determine nearly all the instances while
CLRNet misses some. However, those predictions recalled by
\self but missed by CLRNet are often mistakenly identified as false positives in the default
evaluation metric (F1(0.5, $+\infty$)), leading to an underestimation of model
performance. The use of the F1(0.2, 60) evaluation metric can partially
alleviate such misjudgment and provide a more objective and comprehensive evaluation of model performance.

\section{Conclusions and Future work}
\label{sec:concl-future-work}
There are still fundamental challenges in lane detection to be addressed:
predicting lanes with little- or no-visual-clue, and describing lanes of any shape.
Aiming at these goals, this paper proposes \self, a transformer-based network,
to leverage the global perceptual ability of transformers to improve the instance
recall capability. The anchor-chain representation enables \self to model lanes
flexibly and precisely. To speed up convergence and reduce computation, a
multi-referenced deformable cross-attention module is proposed to work together
with the anchor-chain. In addition, two line IoU algorithms are presented to
underpin the cost and loss functions, which further enhances \self's
representation capability along with the auxiliary branch. Considering the
differences between lanes and typical objects, the F1-score is extended to accept
Fréchet distance as an additional parameter besides reducing the IoU threshold.
Meanwhile, several synthetic metrics are devised to evaluate \self along with
 classic metrics. Our experimental results show that \self achieves state-of-the-art performance on
two well-known datasets.

In the future, we plan to first improve the inference speed of \self (it is currently the
slowest approach in Table~\ref{tab:curvelanes}). In addition, the temporal information implied in the video can provide valuable insights when there are no-visual-clue for lanes in the following frames. Existing methods, such as RVLD~\cite{RVLD}, take advantage of this. In real-time videos, we will investigate sharing semantic information between frames by utilizing updated queries in the decoder.

\section*{Electronic Supplementary Material}
A demo video is available in the online version of this article, which shows how
\self and CLRNet (the current SOTA model) perform on CULane validation set.

The frame is divided into four parts. The upper and lower background areas of
the frame represent the prediction results of \self and CLRNet respectively,
where green and red lines indicate TP (true positive) and FP (false positive)
respectively measured in the default evaluation metric. The
picture-in-picture area in the upper part of the frame also displays the
prediction results of \self, but using the F1(0.2, 60) evaluation metric. The
picture-in-picture area in the lower part of the frame 
shows the GT (ground truth).

\appendix

\section*{Declarations}

\subsection*{Availability of data and materials}
The datasets analyzed during the current study are available in the following repositories: https://github.com/SoulmateB/CurveLanes and https://xingangpan.github.io/projects/CULane.html.

\subsection*{Competing interests}
The authors have no competing interests to declare that are relevant to the content of this article.

\subsection*{Funding}
This paper was supported by the National Natural Science Foundation of China (No. U23A6007).

\subsection*{Author contributions}
Zhongyu Yang, Chen Shen and Wei Shao conceived of the presented idea, Zhongyu Yang proposed and implemented the prototype system. Zhongyu Yang, Chen Shen, Wei Shao and Tengfei Xing designed and performed the experiments and analyzed the data. Runbo Hu, Pengfei Xu, Hua Chai and Ruini Xue supervised the project. Ruini Xue contributed to the interpretation of the results and took the lead in writing the manuscript. All authors discussed the results and contributed to the final manuscript.

\subsection*{Acknowledgements}
We would like to express our sincere gratitude to Xingxu Yao and Yueming Zhang for their valuable assistance in analyzing the experimental results. We would also like to thank Ge Zhang for her valuable insights and assistance in refining the wording of the paper.

% for bibtex
\bibliographystyle{CVMbib}
\bibliography{refs}

\subsection*{Author biography}

\begin{biography}[yangzhongyu]{Zhongyu Yang} received his M.S. degree from the
  University of Electronic Science and Technology of China (UESTC) , Chengdu, in 2023. He
  is now an algorithm engineer in Didi Chuxing.
\end{biography}

\vspace*{2.6em}
\begin{biography}[xueruini]{Ruini Xue} received a Ph.D. degree from Tsinghua
  University in 2009, and is currently an associate professor with the School of
  Computer Science and Engineering in UESTC. His research interests include
  distributed storage systems, graph computing and AI.
\end{biography}

\end{document}